\documentclass{article}
\usepackage{iclr2027_conference,times}

\usepackage{amsmath,amsfonts,bm}

\def\eqref#1{equation~\ref{#1}}

\def\1{\bm{1}}

\DeclareMathAlphabet{\mathsfit}{\encodingdefault}{\sfdefault}{m}{sl}
\SetMathAlphabet{\mathsfit}{bold}{\encodingdefault}{\sfdefault}{bx}{n}

\newcommand{\E}{\mathbb{E}}

\newcommand{\Var}{\mathrm{Var}}

\newcommand{\Cov}{\mathrm{Cov}}

\usepackage{float}
\usepackage{hyperref}
\usepackage{url}
\usepackage{graphicx}
\usepackage{booktabs,multirow,threeparttable,array}
\usepackage{xcolor}
\usepackage{colortbl}
\usepackage{placeins}
\usepackage{xurl}
\definecolor{MoraAccent}{RGB}{28,77,87}
\definecolor{MoraMuted}{RGB}{92,92,92}
\definecolor{MoraTint}{RGB}{239,246,245}
\definecolor{MoraPositive}{RGB}{31,99,94}
\definecolor{MoraNegative}{RGB}{139,58,70}
\definecolor{MoraRule}{RGB}{193,204,203}
\newcommand{\MoraScore}[2]{\ifnum#2=1\textbf{#1}\else#1\fi}

\newcommand{\MoraQuoted}{\ensuremath{^{\dagger}}}
\newcommand{\MoraImp}[1]{{\fontsize{8}{9.6}\selectfont
  \ifdim#1pt<0pt\color{MoraNegative}\else
    \ifdim#1pt>0pt\color{MoraPositive}\else\color{MoraMuted}\fi
  \fi#1}}
\newcommand{\MoraMetric}[1]{{\fontsize{8}{9.6}\selectfont\color{MoraMuted}#1}}
\newcommand{\MoraEntry}[1]{{\fontsize{8}{9.6}\selectfont\color{MoraMuted}#1}}

\newcommand{\MoraGroupRule}{%
  \arrayrulecolor{MoraRule}%
  \specialrule{0.25pt}{1.1pt}{1.1pt}%
  \arrayrulecolor{black}}

\input{rebuttal_admin/revision_markup}
\graphicspath{{Images/}}

\title{MORA: Correcting Time Series Forecasts with Coarse Future Signals}
\author{%
Hu Li$^{1}$ \quad Kenan Li$^{2}$ \quad Haonan Xu$^{1}$ \quad longlong$^{1}$\\
\textbf{Shuqin Gao$^{1}$ \quad Yucheng Zhang$^{1}$ \quad Feng Dai$^{1}$}\\[5pt]
\textnormal{\small $^{1}$Institute of Computing Technology, Chinese Academy of Sciences}\\
\textnormal{\small $^{2}$Institute of Automation, Chinese Academy of Sciences}%
}
\hypersetup{pdfauthor={Hu Li, Kenan Li, Haonan Xu, longlong, Shuqin Gao, Yucheng Zhang, Feng Dai}}

\iclrfinalcopy
\begin{document}
\maketitle
\lhead{} 
\begin{abstract}
Recent architectural advances have improved time-series forecasting. However, our analysis of forecasters with diverse architectures reveals a common pattern: a substantial share of residual energy is concentrated in a small number of low-frequency components. Simple linear probes further show that the forecasters' historical inputs still contain predictive information about errors in the forecast mean. Building on these findings, we propose \textbf{M}odel-\textbf{O}utput \textbf{R}esidual \textbf{A}daptation (\textbf{MORA}), a model-agnostic residual adapter that refines the outputs of frozen forecasters without requiring additional inputs. MORA estimates segment-wise means of the target sequence through structural extrapolation and learned estimation, rather than directly predicting backbone-specific residuals. It then converts discrepancies between these estimates and the corresponding statistics of the backbone forecast into stagewise corrections, scaling each update according to estimation reliability. Extensive experiments on twelve benchmarks with six forecasting backbones show average MSE reductions of \textbf{7.2\%} for long-term forecasting and \textbf{28.1\%} for short-term forecasting, with lower MSE in \textbf{96.2\%} of the 288 evaluated cases. The benefits also extend to three time-series foundation models evaluated at different context lengths. Together, these results demonstrate that predictable coarse-scale residuals can be effectively exploited to improve existing forecasts. Code repository: \url{https://github.com/hugerli-1/MORA}.
\end{abstract}

\section{introduction}
Time-series forecasting aims to predict how a system will evolve over time based on historical observations and plays an important role in applications such as energy management, weather forecasting, and traffic flow prediction~\citep{zhou2021informer,song2020stsgcn}. Advances in deep learning have driven rapid progress in this area. Early approaches relied mainly on recurrent neural networks and temporal convolutional networks~\citep{lai2018lstnet,bai2018empirical}, followed by a growing variety of Transformer- and MLP-based models~\citep{nie2023patchtst,liu2024itransformer,yi2023frets}. These models capture temporal patterns and dependencies in different ways and have continued to improve forecasting performance.

\begin{figure}[!t]
    \centering
    \includegraphics[width=\linewidth]{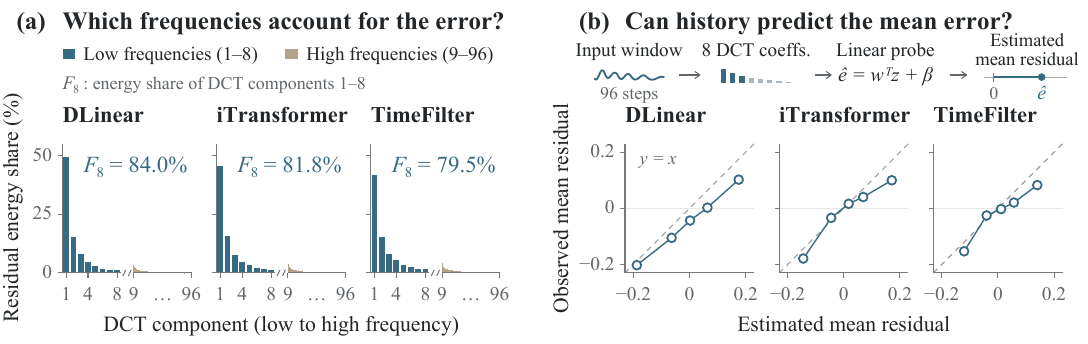}
    \caption{\textbf{Residual analysis of three forecasters on Weather.}
    \textbf{(a)} Residual energy in the first eight DCT components and the remaining components.
    \textbf{(b)} Test samples are grouped by the output of a history-based linear probe. Probe estimates and observed future-mean residuals are shown as group averages. Here, the future-mean residual refers to the true future mean minus the predicted future mean.
}
    \label{fig:motivation}
\end{figure}

However, recent work often pursues further gains through new architectural designs or more complex modules~\citep{hu2025timefilter,liu2026vpnet}, while the improvements are typically incremental~\citep{nematirad2025embeddings,feng2026beyond}. This leads us to ask a different question: rather than continuing to refine the model architecture, do the residuals of existing forecasters exhibit systematic patterns? \morarevised{We analyze the residuals of nine forecasters spanning diverse architectures, including three time-series foundation models, across eight datasets and find that a small number of low-frequency components consistently account for a substantial share of residual energy across model families (Appendix~\ref{app:residual_spectrum_definition}).} As shown in Figure~\ref{fig:motivation}(a), on Weather, the first eight DCT components account for roughly \(80\%\) of the residual energy for each of the three forecasters, despite comprising only \(1/12\) of the frequency dimensions. Since these components capture overall level and slow variation, this result indicates that coarse-scale deviations constitute a substantial part of the remaining forecasting error.

We next examine whether the same observed history contains clues to these coarse-scale errors. Keeping each forecaster fixed, we fit a simple linear probe using the first eight DCT coefficients of its input history to predict the difference between the true and predicted means over the forecast horizon. Figure~\ref{fig:motivation}(b) further shows that, after grouping test samples by the probe output, all three forecasters exhibit the same overall pattern: larger group-average estimates generally correspond to larger observed group-mean residuals. Thus, a simple linear probe using only the history already seen by the forecaster can still distinguish groups that are over- or underpredicted on average. Similar patterns are observed on other datasets, with additional results provided in Appendix~\ref{app:residual_analysis}.

This association suggests scope for history-dependent correction beyond a constant offset. Assuming finite second moments, a nonzero population covariance between the residual along a fixed coarse-scale direction and a scalar function of history implies the existence of an affine correction with strictly lower expected MSE than the best constant correction along that direction (Proposition~1). In practice, however, estimation error can offset this potential gain, motivating corrections that account for the reliability of the estimates.

Building on this, we propose \textbf{M}odel-\textbf{O}utput \textbf{R}esidual \textbf{A}daptation (\textbf{MORA}), a model-agnostic residual adapter that operates directly on the output of a frozen backbone, without modifying the backbone or requiring additional inputs. MORA consists of two components: \textbf{\textit{Coarse Future Estimation}} and \textbf{\textit{Stagewise Residual Adaptation}}. The former represents the future at a coarse temporal resolution using segment-wise means over contiguous intervals, and estimates these statistics from the same input history through structural extrapolation and learned estimation. Because these estimates target coarse-scale statistics of the future itself rather than backbone-specific residuals, they decouple future-state estimation from backbone-specific correction and become residual signals only when compared with the backbone forecast. The latter compares these estimates with the corresponding coarse statistics of the backbone forecast and, guided by estimation reliability, progressively converts their discrepancies into structured trajectory corrections. Rather than relearning the entire future trajectory, MORA uses the backbone forecast as its starting point and corrects only those coarse-scale deviations for which the observed history still contains predictive information.

We evaluate MORA on twelve benchmarks spanning long- and short-term forecasting, using six representative forecasting backbones. Across 288 settings, MORA reduces prediction error in \textbf{96.2\%} of cases, as measured by MSE, with average MSE reductions of \textbf{7.2\%} on the long-term benchmarks and \textbf{28.1\%} on the short-term benchmarks. We further evaluate MORA on three time-series foundation models with frozen parameters, yielding average MSE reductions of \textbf{24.0\%} and \textbf{6.9\%} at context lengths of 96 and 512, respectively. These gains are also accompanied by lower MAE on average. Taken together, these results show that MORA is broadly effective across model architectures, forecast horizons, and context lengths.

In summary, our main contributions are as follows:
\begin{itemize}
    \item By analyzing residuals from forecasters with diverse architectures, we show that the remaining errors exhibit pronounced coarse-scale structure and that the same input history used for forecasting still contains predictive signals about these errors.
    \item We propose \textbf{MORA}, a model-agnostic residual adapter that separates coarse future estimation from backbone-specific correction. MORA refines the outputs of frozen forecasters in stages, scaling each correction according to the reliability of the underlying estimate.
    \item Extensive experiments on twelve forecasting benchmarks show that MORA reduces MSE in \textbf{96.2\%} of 288 long- and short-term forecasting settings across six representative backbones. Results on three time-series foundation models further demonstrate its broad effectiveness.
\end{itemize}


\section{methodology}
\subsection{Problem Setup}
Given a historical window
$\mathbf{X}\in\mathbb{R}^{L\times C}$
and its corresponding future sequence
$\mathbf{Y}\in\mathbb{R}^{H\times C}$,
where $L$, $H$, and $C$ denote the lookback length, forecast horizon, and number of variables, respectively, a pretrained forecaster $f_{\theta}$ produces an $H$-step prediction
\begin{equation}
    \widehat{\mathbf{Y}}^{(0)}
    =
    f_{\theta}(\mathbf{X})
    \in\mathbb{R}^{H\times C}.
\end{equation}

We freeze all parameters of $f_{\theta}$ and learn an output-space residual adapter that refines the backbone prediction through a structured correction,
\begin{equation}
    \widehat{\mathbf{Y}}
    =
    \widehat{\mathbf{Y}}^{(0)}
    +
    \Delta_{\phi}
    \bigl(
        \mathbf{X},
        \widehat{\mathbf{Y}}^{(0)}
    \bigr),
\end{equation}
where
$\Delta_{\phi}(\mathbf{X},\widehat{\mathbf{Y}}^{(0)})
\in\mathbb{R}^{H\times C}$
estimates the predictable component of the residual
$\mathbf{Y}-\widehat{\mathbf{Y}}^{(0)}$.
The adapter is fitted offline with the backbone frozen.
At inference, its parameters remain fixed, and it operates on the standard forecasting query and the backbone prediction without modifying the backbone or extending its input window.

\subsection{Model-Output Residual Adaptation}
\label{sec:mora}
\begin{figure}[!t]
    \centering
    \includegraphics[width=\linewidth]{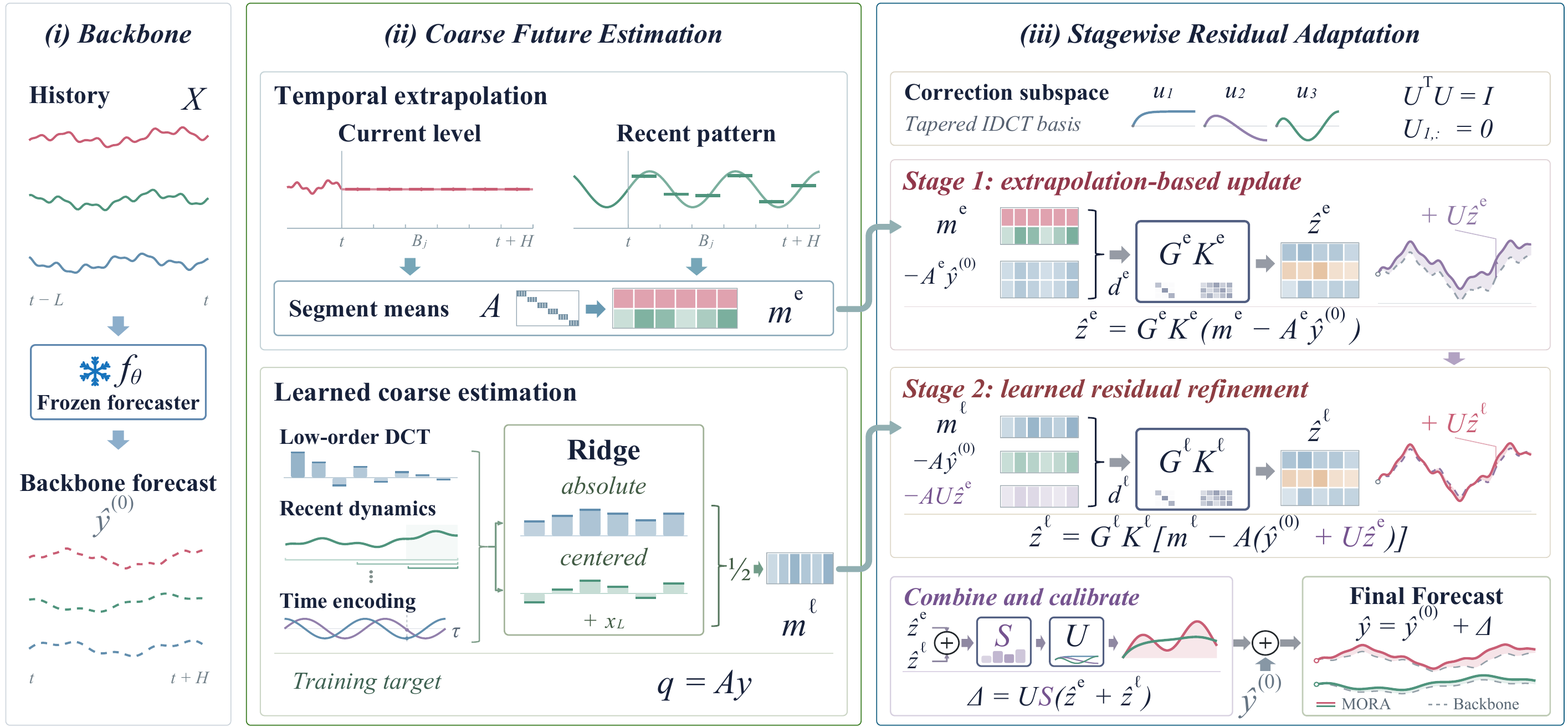}
    \caption{
    \textbf{Overview of MORA.}
    (i) A frozen backbone produces the initial forecast.
    (ii) Temporal extrapolation and learned coarse estimation provide complementary estimates of future segment means.
    (iii) These estimates guide two sequential residual updates, with the second stage addressing the discrepancy remaining after the first-stage correction.
    The two updates are combined, calibrated, and reconstructed into a correction that is added to the initial forecast.
    Curves and matrix entries are schematic; channel indices are omitted for clarity.
    }
    \label{fig:mora}
\end{figure}

We introduce \textbf{Model-Output Residual Adaptation (MORA)},
illustrated in Figure~\ref{fig:mora}.
MORA comprises two components: \textbf{\emph{Coarse Future Estimation}} and \textbf{\emph{Stagewise Residual Adaptation}}. The first estimates a coarse representation of the future rather than backbone-specific residuals. The second compares these estimates with the backbone forecast at the same coarse scale and translates their discrepancies into structured residual updates. Estimates based on the current level and recurring patterns guide an initial update; a learned estimate is then compared with the updated forecast to refine the remaining residual. The resulting correction is calibrated, reconstructed over the forecast horizon, and added to the backbone prediction.

\subsubsection{Coarse Future Estimation}
\label{sec:coarse_future_estimation}
MORA first estimates a coarse representation of the future rather than
a residual tied to a particular backbone.
For channel $c$, let
$\mathbf{x}_c=\mathbf{X}_{:,c}\in\mathbb{R}^{L}$
and
$\mathbf{y}_c=\mathbf{Y}_{:,c}\in\mathbb{R}^{H}$
denote its historical and future trajectories, respectively.
We partition the forecast horizon into $Q$ contiguous, non-overlapping
intervals $\{\mathcal{B}_j\}_{j=1}^{Q}$ of approximately equal length
and define a fixed linear operator
$\mathbf{A}\in\mathbb{R}^{Q\times H}$ such that
\begin{equation}
    \mathbf{q}_c
    =
    \mathbf{A}\mathbf{y}_c,
    \qquad
    q_{j,c}
    =
    \frac{1}{|\mathcal{B}_j|}
    \sum_{t\in\mathcal{B}_j} y_{t,c}.
    \label{eq:coarse_representation}
\end{equation}
Here $\mathbf{q}_c\in\mathbb{R}^{Q}$ captures the evolution of the
future at a coarser temporal scale.
We choose
$Q=\min\{Q_{\max},\lfloor H/b_{\min}\rfloor\}$,
where $Q_{\max}$ caps the representation dimension and $b_{\min}$
sets the minimum temporal span of each interval.
In our experiments, $Q_{\max}=64$ and $b_{\min}=4$.

We estimate the same coarse representation in two ways.
First, we extrapolate simple temporal structure from the observed
history.
Let $\boldsymbol{\tau}$ denote the temporal coordinates of the forecast timestamps, and let
$\mathcal{C}_j(\mathbf{x}_c,\boldsymbol{\tau})\in\mathbb{R}^{H}$
denote the $j$-th extrapolated trajectory.
These extrapolations use the current level, recent-pattern
continuation, and, when enabled, a periodic template fitted on the
training data.
Each extrapolation yields
\begin{equation}
    \mathbf{m}_{j,c}^{e}
    =
    \mathbf{A}\mathcal{C}_j(\mathbf{x}_c,\boldsymbol{\tau})
    \in\mathbb{R}^{Q},
    \qquad j=1,\ldots,J.
    \label{eq:temporal_extrapolation}
\end{equation}
We retain the $J$ estimates separately and stack them as
$\mathbf{m}_c^{e}\in\mathbb{R}^{JQ}$.
The exact extrapolation rules and periodic-template construction are
given in Appendix~\ref{app:coarse_estimates}.

These temporal extrapolations provide low-capacity coarse future
estimates under explicit temporal assumptions.
We further introduce a learned predictor for the same coarse future target
$\mathbf{q}_c$.
For target channel $c$, we construct a deterministic feature vector $\boldsymbol{\phi}_c(\mathbf{X},\boldsymbol{\tau})\in\mathbb{R}^{D}$ 
from the multivariate history $\mathbf{X}$ and known temporal context $\boldsymbol{\tau}$, where $D$ denotes the feature dimension. The feature vector summarizes low-order DCT components~\citep{ahmed1974discrete} of the history, whole-window and block-level statistics, multi-scale recent dynamics, and periodic time encodings. For $L=96$, the compact representation has $D=78$ dimensions. We fit ridge regressors~\citep{hoerl1970ridge} separately for each target channel to map these historical features to the $Q$-dimensional coarse future quantity. To capture both the absolute level and changes relative to the current state, we use two coordinate systems: one retains the original coordinates, while the other is centered at the latest observation.
With $x_{L,c}$ denoting the latest observation of channel $c$, the
centered history and target are
\begin{equation}
    \widetilde{\mathbf{x}}_c
    =
    \mathbf{x}_c-x_{L,c}\mathbf{1}_{L},
    \qquad
    \widetilde{\mathbf{q}}_c
    =
    \mathbf{q}_c-x_{L,c}\mathbf{1}_{Q},
    \label{eq:last_centering}
\end{equation}
where $\mathbf{1}_n$ denotes the $n$-dimensional all-ones vector. Let
$\mathbf{m}_c^{\mathrm{abs}}\in\mathbb{R}^{Q}$
denote the prediction in the original coordinates and
$\widetilde{\mathbf{m}}_c^{\mathrm{rel}}\in\mathbb{R}^{Q}$
the prediction in the centered coordinates.
After restoring the latter to the original level, the learned coarse
estimate is
\begin{equation}
    \mathbf{m}_c^{\ell}
    =
    \frac{1}{2}\mathbf{m}_c^{\mathrm{abs}}
    +
    \frac{1}{2}
    \left(
        \widetilde{\mathbf{m}}_c^{\mathrm{rel}}
        +
        x_{L,c}\mathbf{1}_{Q}
    \right).
    \label{eq:learned_coarse_estimate}
\end{equation}
Thus, the learned predictor estimates the future coarse representation $\mathbf{q}_c$, rather than a backbone-specific residual. The exact feature construction, feature and target standardization, and ridge-fitting details are provided in
Appendix~\ref{app:coarse_estimates}.

For series with many variables, fitting a separate learned estimator for every channel makes the parameter count and fitting cost scale linearly with $C$.
We therefore replace the channel-wise estimator with a factorized
coarse estimator based on a shared low-dimensional channel-factor
model. Specifically, we first estimate a channel-wise periodic baseline from the training data by averaging each channel at every phase of a configured period.
After subtracting this baseline and standardizing the remaining
multivariate deviations, we project them into the shared factor space.
A shared ridge model predicts the coarse future states of these factors,
which are then mapped back to the original channels with the
channel-wise mean, scale, and periodic baseline restored.
The resulting channel-level estimates retain the form
$\mathbf{m}_c^{\ell}\in\mathbb{R}^{Q}$, leaving the subsequent residual adaptation unchanged. Details are provided in
Appendix~\ref{app:factorized_estimator}.
\paragraph{Backbone-relative residual signal.}
The coarse future estimates become backbone-relative by comparing them
with the backbone forecast at the same temporal scale.
For any individual coarse estimate
$\mathbf{m}_c\in\mathbb{R}^{Q}$ of $\mathbf{q}_c$,
\begin{equation}
    \mathbf{d}_c
    =
    \mathbf{m}_c
    -
    \mathbf{A}\widehat{\mathbf{y}}_c^{(0)}
    =
    \mathbf{A}
    \left(
        \mathbf{y}_c-\widehat{\mathbf{y}}_c^{(0)}
    \right)
    +
    \left(
        \mathbf{m}_c-\mathbf{A}\mathbf{y}_c
    \right).
    \label{eq:backbone_relative_signal}
\end{equation}
The first term is the backbone residual at the same coarse scale,
while the second is the error of the coarse future estimate.
Thus, a coarse future estimate can be converted into a
backbone-relative residual signal by subtracting the corresponding
coarse prediction of the backbone.

\subsubsection{Stagewise Residual Adaptation}
\label{sec:ordered_residual_adaptation}
\paragraph{Correction subspace.}
The coarse representation does not uniquely determine a pointwise
correction. We therefore restrict the correction to a predefined low-dimensional subspace over the forecast horizon, reducing its degrees of freedom while retaining the backbone forecast as the reference prediction.
Let $\mathbf{U}\in\mathbb{R}^{H\times k}$ contain an orthonormal basis for this correction subspace,
where $k$ is the correction rank and
$\mathbf{U}^{\top}\mathbf{U}=\mathbf{I}_k$.
For the backbone residual
$\mathbf{r}_c=\mathbf{y}_c-\widehat{\mathbf{y}}_c^{(0)}$, we write
\begin{equation}
    \mathbf{r}_c
    =
    \mathbf{U}\mathbf{z}_c
    +
    \mathbf{r}_{\perp,c},
    \qquad
    \mathbf{z}_c
    =
    \mathbf{U}^{\top}\mathbf{r}_c,
    \label{eq:residual_coordinates}
\end{equation}
where $\mathbf{z}_c\in\mathbb{R}^{k}$ contains the coordinates of the residual within the correction subspace, and $\mathbf{r}_{\perp,c}\in\mathbb{R}^{H}$ is the orthogonal remainder.

MORA estimates these residual coordinates as $\widehat{\mathbf{z}}_c$ and reconstructs the correction within the span of $\mathbf{U}$.
We construct $\mathbf{U}$ by tapering low-order inverse discrete cosine
transform (IDCT) modes near the forecast boundary and orthonormalizing
the resulting columns.
The first row of $\mathbf{U}$ is zero, so the correction leaves the
backbone's first predicted value unchanged.
For long-term forecasting, we use $k=64$; construction details are provided in
Appendix~\ref{app:correction_space}.

For any individual coarse estimate
$\mathbf{m}_c\in\mathbb{R}^{Q}$, the backbone-relative discrepancy
satisfies
\begin{equation}
    \mathbf{d}_c
    =
    \mathbf{A}\mathbf{U}\mathbf{z}_c
    +
    \boldsymbol{\eta}_c,
    \qquad
    \boldsymbol{\eta}_c
    =
    \left(
        \mathbf{m}_c-\mathbf{q}_c
    \right)
    +
    \mathbf{A}\mathbf{r}_{\perp,c}.
    \label{eq:coarse_residual_model}
\end{equation}
Here $\boldsymbol{\eta}_c\in\mathbb{R}^{Q}$ contains the coarse-estimation
error and the contribution of residuals outside the correction space.
Thus, the discrepancy $\mathbf{d}_c$ provides a noisy signal of the
residual coordinates $\mathbf{z}_c$.
\paragraph{Residual refinement.}
Rather than combining the extrapolation-based and learned estimates in a single update, MORA performs residual adaptation in two stages. The extrapolation-based estimates first produce an initial correction, after which the learned estimate is evaluated against the updated forecast to refine the residual discrepancy.

Let $\mathbf{A}^{e}\in\mathbb{R}^{JQ\times H}$ denote $J$ vertically
stacked copies of $\mathbf{A}$.
The channel-specific matrices
$\mathbf{K}_c^{e}\in\mathbb{R}^{k\times JQ}$ and
$\mathbf{K}_c^{\ell}\in\mathbb{R}^{k\times Q}$
map coarse discrepancies to residual coordinates.
The diagonal matrices
$\mathbf{G}^{e},\mathbf{G}^{\ell}\in\mathbb{R}^{k\times k}$
provide coordinate-wise reliability scaling shared across channels.
The ordered updates are
\begin{equation}
    \begin{aligned}
        \mathbf{d}_c^{e}
        &=
        \mathbf{m}_c^{e}
        -
        \mathbf{A}^{e}\widehat{\mathbf{y}}_c^{(0)},
        &
        \widehat{\mathbf{z}}_c^{e}
        &=
        \mathbf{G}^{e}\mathbf{K}_c^{e}\mathbf{d}_c^{e},
        \\[1mm]
        \mathbf{d}_c^{\ell}
        &=
        \mathbf{m}_c^{\ell}
        -
        \mathbf{A}
        \left(
            \widehat{\mathbf{y}}_c^{(0)}
            +
            \mathbf{U}\widehat{\mathbf{z}}_c^{e}
        \right),
        &
        \widehat{\mathbf{z}}_c^{\ell}
        &=
        \mathbf{G}^{\ell}\mathbf{K}_c^{\ell}\mathbf{d}_c^{\ell}.
    \end{aligned}
    \label{eq:ordered_updates}
\end{equation}
Here, $\widehat{\mathbf{z}}_c^{e}$ and
$\widehat{\mathbf{z}}_c^{\ell}$ denote the coefficient-space residual
updates from the first and second corrections, respectively. The learned discrepancy is then computed against the updated forecast
$\widehat{\mathbf{y}}_c^{(0)}+\mathbf{U}\widehat{\mathbf{z}}_c^{e}$,
so the second update targets the residual remaining after the
extrapolation-based correction.

\paragraph{Final calibration and prediction.}
The two ordered updates produce a combined correction
$\widehat{\mathbf{z}}_c^{e}+\widehat{\mathbf{z}}_c^{\ell}$ in the $k$-dimensional coefficient space.
Before reconstructing this correction over the forecast horizon, we calibrate its magnitude to account for differences across channels and basis-coordinate ranges. For channel $c$, the $k$ basis coordinates are partitioned into four consecutive groups, with one validation-fitted scale shared within each group. We collect these scales in a diagonal matrix $\mathbf{S}_c\in\mathbb{R}^{k\times k}$. Thus, although $\mathbf{S}_c$ is written as a $k\times k$ matrix, it contains only four free scalar parameters.

The final forecast is
\begin{equation}
    \boxed{
        \widehat{\mathbf{y}}_c
        =
        \widehat{\mathbf{y}}_c^{(0)}
        +
        \mathbf{U}\mathbf{S}_c
        \left(
            \widehat{\mathbf{z}}_c^{e}
            +
            \widehat{\mathbf{z}}_c^{\ell}
        \right)
    }.
    \label{eq:final_prediction}
\end{equation}
The matrix $\mathbf{S}_c$ calibrates the combined residual coefficients
before $\mathbf{U}$ maps them back to the forecast horizon.
This calibration is applied only after both residual updates and does
not affect the intermediate forecast used by the learned stage.
Stacking the channel-wise forecasts in
Eq.~\eqref{eq:final_prediction} yields
$\widehat{\mathbf{Y}}$, while the corresponding corrections form the
residual adapter
$\Delta_{\phi}(\mathbf{X},\widehat{\mathbf{Y}}^{(0)})$
introduced in Eq.~(2).

\subsubsection{Fitting the Residual Adapter}
\label{sec:adapter_training}
For the coarse estimates from temporal extrapolation, the residual
mapping is estimated using pairs
$(\mathbf{d}_c^{e}, \mathbf{z}_c)$.
For the learned coarse estimate, we define the remaining target as
$\mathbf{z}_c^{\ell}
=
\mathbf{z}_c-\widehat{\mathbf{z}}_c^{e}$.
The corresponding discrepancy
$\mathbf{d}_c^{\ell}$
is computed against the forecast corrected by
$\widehat{\mathbf{z}}_c^{e}$,
following Equation~\eqref{eq:ordered_updates}.
The learned residual mapping is then estimated using pairs
$(\mathbf{d}_c^{\ell}, \mathbf{z}_c^{\ell})$.

The channel-specific matrices
$\mathbf{K}_c^{e}$ and $\mathbf{K}_c^{\ell}$
are computed from regularized second-order statistics of the
corresponding discrepancy--target pairs.
The gains
$\mathbf{G}^{e}$ and $\mathbf{G}^{\ell}$
are estimated separately using a coordinate-wise reliability rule
shared across channels.
The complete estimators, regularization, and reliability-scaling rules
are provided in Appendix~\ref{app:update_estimation}.

The learned coarse predictor is fitted on the training split.
For estimating the residual mappings, we use blocked out-of-fold (OOF)
predictions on the same split to construct the required statistics and
the corresponding $\widehat{\mathbf{z}}_c^{e}$ used to form
$\mathbf{z}_c^{\ell}$.
Once $\mathbf{K}_c^{e}$, $\mathbf{K}_c^{\ell}$,
$\mathbf{G}^{e}$, and $\mathbf{G}^{\ell}$ are fixed, the final
calibration $\mathbf{S}_c$ is fitted on validation data using the
calibration objective defined above.
The complete fitting protocol is deferred to
Appendix~\ref{app:fitting_protocol}.

\section{Experiments}
\label{sec:experiments}
\newcommand{\MoraTaskMSE}{7.188}
\newcommand{\MoraTaskMAE}{3.458}
\newcommand{\MoraTaskMSEGroups}{46}
\newcommand{\MoraTaskMAEGroups}{45}
\newcommand{\MoraTaskGroups}{48}
\newcommand{\MoraTaskSettings}{192}
\newcommand{\MoraTaskMSESettings}{187}
\newcommand{\MoraTaskMAESettings}{179}
\newcommand{\MoraTaskCGTFraMSE}{4.340}
\newcommand{\MoraTaskCGTFraMAE}{1.013}
\newcommand{\MoraTaskDLinearMSE}{14.935}
\newcommand{\MoraTaskDLinearMAE}{10.601}
\newcommand{\MoraTaskPatchTSTMSE}{9.293}
\newcommand{\MoraTaskPatchTSTMAE}{4.224}
\newcommand{\MoraTaskTimeFilterMSE}{4.438}
\newcommand{\MoraTaskTimeFilterMAE}{1.596}
\newcommand{\MoraTaskVPNetMSE}{2.757}
\newcommand{\MoraTaskVPNetMAE}{0.548}
\newcommand{\MoraTaskiTransformerMSE}{7.366}
\newcommand{\MoraTaskiTransformerMAE}{2.766}
\newcommand{\MoraFoundationShortMSE}{23.973}
\newcommand{\MoraFoundationShortMAE}{10.685}
\newcommand{\MoraFoundationShortMSEGroups}{24}
\newcommand{\MoraFoundationShortMAEGroups}{24}
\newcommand{\MoraFoundationShortGroups}{24}
\newcommand{\MoraFoundationShortSettings}{96}
\newcommand{\MoraFoundationShortMSESettings}{94}
\newcommand{\MoraFoundationShortMAESettings}{90}
\newcommand{\MoraFoundationShortChronosTwoMSE}{22.266}
\newcommand{\MoraFoundationShortChronosTwoMAE}{6.676}
\newcommand{\MoraFoundationShortSundialMSE}{20.808}
\newcommand{\MoraFoundationShortSundialMAE}{10.920}
\newcommand{\MoraFoundationShortTimesFMTwoFiveMSE}{28.846}
\newcommand{\MoraFoundationShortTimesFMTwoFiveMAE}{14.458}
\newcommand{\MoraFoundationLongMSE}{6.932}
\newcommand{\MoraFoundationLongMAE}{1.276}
\newcommand{\MoraFoundationLongMSEGroups}{24}
\newcommand{\MoraFoundationLongMAEGroups}{19}
\newcommand{\MoraFoundationLongGroups}{24}
\newcommand{\MoraFoundationLongSettings}{96}
\newcommand{\MoraFoundationLongMSESettings}{93}
\newcommand{\MoraFoundationLongMAESettings}{71}
\newcommand{\MoraFoundationLongChronosTwoMSE}{6.302}
\newcommand{\MoraFoundationLongChronosTwoMAE}{0.926}
\newcommand{\MoraFoundationLongSundialMSE}{5.985}
\newcommand{\MoraFoundationLongSundialMAE}{1.891}
\newcommand{\MoraFoundationLongTimesFMTwoFiveMSE}{8.508}
\newcommand{\MoraFoundationLongTimesFMTwoFiveMAE}{1.011}
\newcommand{\MoraPemsMSE}{28.095}
\newcommand{\MoraPemsMAE}{18.474}
\newcommand{\MoraPemsMSEGroups}{24}
\newcommand{\MoraPemsMAEGroups}{24}
\newcommand{\MoraPemsGroups}{24}
\newcommand{\MoraPemsSettings}{96}
\newcommand{\MoraPemsMSESettings}{90}
\newcommand{\MoraPemsMAESettings}{94}
\newcommand{\MoraPemsTimeFilterMSE}{12.105}
\newcommand{\MoraPemsTimeFilterMAE}{7.437}
\newcommand{\MoraPemsiTransformerMSE}{11.570}
\newcommand{\MoraPemsiTransformerMAE}{9.267}
\newcommand{\MoraPemsCGTFraMSE}{18.589}
\newcommand{\MoraPemsCGTFraMAE}{10.235}
\newcommand{\MoraPemsPatchTSTMSE}{52.888}
\newcommand{\MoraPemsPatchTSTMAE}{32.643}
\newcommand{\MoraPemsDLinearMSE}{54.984}
\newcommand{\MoraPemsDLinearMAE}{39.109}
\newcommand{\MoraPemsVPNetMSE}{18.435}
\newcommand{\MoraPemsVPNetMAE}{12.149}

\subsection{Experimental Setup}
\label{sec:experimental_setup}

\paragraph{Datasets and backbones.}
We evaluate MORA on eight long-horizon forecasting datasets,
ETTh1, ETTh2, ETTm1, ETTm2, Weather, ECL, Traffic, and Solar,
and four short-horizon traffic datasets, PEMS03, PEMS04, PEMS07,
and PEMS08. All datasets are split chronologically into training,
validation, and test sets. ETT follows the standard 12/4/4-month
split; the other long-horizon datasets and PEMS use ratios of
7:1:2 and 6:2:2, respectively.
We select six representative backbones spanning classical architectures
and recent state-of-the-art methods: DLinear~\citep{zeng2023transformers},
PatchTST~\citep{nie2023patchtst},
iTransformer~\citep{liu2024itransformer},
TimeFilter~\citep{hu2025timefilter},
CGTFra~\citep{Yu2026CGTFra}, and VPNet~\citep{liu2026vpnet}.
For long-horizon forecasting, we additionally evaluate three pretrained
time-series foundation models: Chronos-2~\citep{ansari2025chronos2},
Sundial~\citep{liu2025sundial}, and
TimesFM~2.5~\citep{das2024timesfm,google2025timesfm25}.
More details on the datasets and backbones are provided in
Appendices~\ref{app:experimental_data} and~\ref{app:experimental_backbones}.

\paragraph{Implementation details.}
Following common forecasting protocols~\citep{liu2024itransformer}, we use a lookback length of
$L=96$ for task-specific models, with $H\in\{96,192,336,720\}$ for
long-horizon forecasting and $H\in\{12,24,48,96\}$ for short-horizon
forecasting. To examine adaptation with longer histories, foundation
models are evaluated at both $L=96$ and $L=512$, with an adapter
fitted separately for each context length. All backbone parameters
remain frozen during adaptation. MORA's coarse estimators and residual
maps are fitted on training data, followed by final calibration on
validation forecasts; neither the backbone nor the adapter is updated
at test time. For ECL, Traffic, and PEMS, we use the factorized coarse
estimator, while the remaining datasets use the channel-wise estimator.
We report MSE and MAE, and use Imp.\ to denote the percentage error
reduction relative to the corresponding baseline.
More details on hyperparameters, fitting, and computing resources are
given in Appendix~\ref{app:experimental_fitting}; results for individual
horizons and seeds are provided in Appendix~\ref{app:complete_results}.

\subsection{Forecasting Results}
\label{sec:forecasting_results}
\label{sec:task_specific_results}
\begin{table}[!tp]
\centering
\begingroup
\caption{Baseline and MORA results for six backbones on eight
long-term forecasting datasets. MSE and MAE are averaged over
$H\in\{96,192,336,720\}$ at $L=96$.
In Tables~\ref{tab:mora_backbones_main}--\ref{tab:mora_foundation_main},
bold marks the lower displayed error within each baseline--MORA
comparison.}
\label{tab:mora_backbones_main}
\fontsize{9}{10.2}\selectfont
\setlength{\tabcolsep}{1.5pt}
\renewcommand{\arraystretch}{0.90}
\setlength{\aboverulesep}{1.2pt}
\setlength{\belowrulesep}{1.5pt}
\resizebox{0.90\linewidth}{!}{%
\begin{tabular*}{\linewidth}{@{}l@{\hspace{4pt}}l@{\extracolsep{\fill}\hspace{5.5pt}}r@{\extracolsep{0pt}\hspace{3pt}}r@{\extracolsep{\fill}\hspace{5.5pt}}r@{\extracolsep{0pt}\hspace{3pt}}r@{\extracolsep{\fill}\hspace{5.5pt}}r@{\extracolsep{0pt}\hspace{3pt}}r@{\extracolsep{\fill}\hspace{5.5pt}}r@{\extracolsep{0pt}\hspace{3pt}}r@{\extracolsep{\fill}\hspace{5.5pt}}r@{\extracolsep{0pt}\hspace{3pt}}r@{\extracolsep{\fill}\hspace{5.5pt}}r@{\extracolsep{0pt}\hspace{3pt}}r@{}}
\toprule[0.8pt]
\multicolumn{2}{l}{\textbf{Dataset}} & \multicolumn{2}{c}{\textbf{TimeFilter}} & \multicolumn{2}{c}{\textbf{iTransformer}} & \multicolumn{2}{c}{\textbf{CGTFra}} & \multicolumn{2}{c}{\textbf{PatchTST}} & \multicolumn{2}{c}{\textbf{DLinear}} & \multicolumn{2}{c}{\textbf{VPNet}} \\
\cmidrule(lr){3-4}\cmidrule(lr){5-6}\cmidrule(lr){7-8}\cmidrule(lr){9-10}\cmidrule(lr){11-12}\cmidrule(lr){13-14}
 & & \MoraMetric{MSE} & \MoraMetric{MAE} & \MoraMetric{MSE} & \MoraMetric{MAE} & \MoraMetric{MSE} & \MoraMetric{MAE} & \MoraMetric{MSE} & \MoraMetric{MAE} & \MoraMetric{MSE} & \MoraMetric{MAE} & \MoraMetric{MSE} & \MoraMetric{MAE} \\
\midrule
 & \MoraEntry{Base} & \MoraScore{0.418}{0} & \MoraScore{0.427}{0} & \MoraScore{0.455}{0} & \MoraScore{0.447}{0} & \MoraScore{0.435}{1} & \MoraScore{0.428}{1} & \MoraScore{0.448}{0} & \MoraScore{0.449}{0} & \MoraScore{0.462}{0} & \MoraScore{0.459}{0} & \MoraScore{0.436}{0} & \MoraScore{0.429}{0} \\
 & \textcolor{MoraAccent}{\textbf{+MORA}} & \MoraScore{0.410}{1} & \MoraScore{0.423}{1} & \MoraScore{0.448}{1} & \MoraScore{0.442}{1} & \MoraScore{0.438}{0} & \MoraScore{0.430}{0} & \MoraScore{0.428}{1} & \MoraScore{0.436}{1} & \MoraScore{0.439}{1} & \MoraScore{0.445}{1} & \MoraScore{0.433}{1} & \MoraScore{0.427}{1} \\
\rowcolor{MoraTint}
\multirow{-3}{*}{\textbf{ETTh1}} & \MoraEntry{Imp.\ (\%)} & \MoraImp{1.985} & \MoraImp{0.978} & \MoraImp{1.510} & \MoraImp{1.196} & \MoraImp{-0.546} & \MoraImp{-0.334} & \MoraImp{4.673} & \MoraImp{2.761} & \MoraImp{5.048} & \MoraImp{2.926} & \MoraImp{0.737} & \MoraImp{0.314} \\
\MoraGroupRule
 & \MoraEntry{Base} & \MoraScore{0.372}{0} & \MoraScore{0.403}{0} & \MoraScore{0.383}{0} & \MoraScore{0.407}{0} & \MoraScore{0.376}{0} & \MoraScore{0.397}{0} & \MoraScore{0.390}{0} & \MoraScore{0.413}{0} & \MoraScore{0.562}{0} & \MoraScore{0.519}{0} & \MoraScore{0.359}{1} & \MoraScore{0.385}{1} \\
 & \textcolor{MoraAccent}{\textbf{+MORA}} & \MoraScore{0.370}{1} & \MoraScore{0.399}{1} & \MoraScore{0.372}{1} & \MoraScore{0.398}{1} & \MoraScore{0.369}{1} & \MoraScore{0.393}{1} & \MoraScore{0.367}{1} & \MoraScore{0.397}{1} & \MoraScore{0.487}{1} & \MoraScore{0.466}{1} & \MoraScore{0.363}{0} & \MoraScore{0.389}{0} \\
\rowcolor{MoraTint}
\multirow{-3}{*}{\textbf{ETTh2}} & \MoraEntry{Imp.\ (\%)} & \MoraImp{0.508} & \MoraImp{0.925} & \MoraImp{2.895} & \MoraImp{2.078} & \MoraImp{1.734} & \MoraImp{1.069} & \MoraImp{5.752} & \MoraImp{3.997} & \MoraImp{13.410} & \MoraImp{10.098} & \MoraImp{-1.044} & \MoraImp{-1.106} \\
\MoraGroupRule
 & \MoraEntry{Base} & \MoraScore{0.377}{0} & \MoraScore{0.394}{0} & \MoraScore{0.407}{0} & \MoraScore{0.411}{0} & \MoraScore{0.393}{0} & \MoraScore{0.390}{0} & \MoraScore{0.387}{0} & \MoraScore{0.402}{0} & \MoraScore{0.403}{0} & \MoraScore{0.408}{0} & \MoraScore{0.376}{0} & \MoraScore{0.382}{0} \\
 & \textcolor{MoraAccent}{\textbf{+MORA}} & \MoraScore{0.357}{1} & \MoraScore{0.385}{1} & \MoraScore{0.369}{1} & \MoraScore{0.394}{1} & \MoraScore{0.360}{1} & \MoraScore{0.380}{1} & \MoraScore{0.362}{1} & \MoraScore{0.390}{1} & \MoraScore{0.362}{1} & \MoraScore{0.386}{1} & \MoraScore{0.355}{1} & \MoraScore{0.378}{1} \\
\rowcolor{MoraTint}
\multirow{-3}{*}{\textbf{ETTm1}} & \MoraEntry{Imp.\ (\%)} & \MoraImp{5.184} & \MoraImp{2.088} & \MoraImp{9.317} & \MoraImp{4.279} & \MoraImp{8.295} & \MoraImp{2.544} & \MoraImp{6.616} & \MoraImp{2.936} & \MoraImp{10.261} & \MoraImp{5.209} & \MoraImp{5.695} & \MoraImp{1.111} \\
\MoraGroupRule
 & \MoraEntry{Base} & \MoraScore{0.271}{0} & \MoraScore{0.321}{0} & \MoraScore{0.290}{0} & \MoraScore{0.335}{0} & \MoraScore{0.278}{0} & \MoraScore{0.318}{0} & \MoraScore{0.293}{0} & \MoraScore{0.336}{0} & \MoraScore{0.352}{0} & \MoraScore{0.400}{0} & \MoraScore{0.274}{0} & \MoraScore{0.315}{0} \\
 & \textcolor{MoraAccent}{\textbf{+MORA}} & \MoraScore{0.264}{1} & \MoraScore{0.317}{1} & \MoraScore{0.270}{1} & \MoraScore{0.322}{1} & \MoraScore{0.265}{1} & \MoraScore{0.313}{1} & \MoraScore{0.275}{1} & \MoraScore{0.327}{1} & \MoraScore{0.303}{1} & \MoraScore{0.354}{1} & \MoraScore{0.265}{1} & \MoraScore{0.311}{1} \\
\rowcolor{MoraTint}
\multirow{-3}{*}{\textbf{ETTm2}} & \MoraEntry{Imp.\ (\%)} & \MoraImp{2.637} & \MoraImp{1.059} & \MoraImp{7.073} & \MoraImp{3.853} & \MoraImp{4.724} & \MoraImp{1.703} & \MoraImp{6.265} & \MoraImp{2.642} & \MoraImp{13.825} & \MoraImp{11.501} & \MoraImp{3.423} & \MoraImp{1.205} \\
\MoraGroupRule
 & \MoraEntry{Base} & \MoraScore{0.240}{0} & \MoraScore{0.270}{0} & \MoraScore{0.259}{0} & \MoraScore{0.279}{0} & \MoraScore{0.240}{0} & \MoraScore{0.262}{0} & \MoraScore{0.258}{0} & \MoraScore{0.280}{0} & \MoraScore{0.267}{0} & \MoraScore{0.319}{0} & \MoraScore{0.238}{0} & \MoraScore{0.262}{0} \\
 & \textcolor{MoraAccent}{\textbf{+MORA}} & \MoraScore{0.227}{1} & \MoraScore{0.265}{1} & \MoraScore{0.231}{1} & \MoraScore{0.275}{1} & \MoraScore{0.222}{1} & \MoraScore{0.257}{1} & \MoraScore{0.231}{1} & \MoraScore{0.277}{1} & \MoraScore{0.236}{1} & \MoraScore{0.294}{1} & \MoraScore{0.228}{1} & \MoraScore{0.259}{1} \\
\rowcolor{MoraTint}
\multirow{-3}{*}{\textbf{Weather}} & \MoraEntry{Imp.\ (\%)} & \MoraImp{5.522} & \MoraImp{1.683} & \MoraImp{10.970} & \MoraImp{1.611} & \MoraImp{7.356} & \MoraImp{1.758} & \MoraImp{10.624} & \MoraImp{1.176} & \MoraImp{11.648} & \MoraImp{7.568} & \MoraImp{4.021} & \MoraImp{1.188} \\
\MoraGroupRule
 & \MoraEntry{Base} & \MoraScore{0.164}{0} & \MoraScore{0.261}{0} & \MoraScore{0.175}{0} & \MoraScore{0.267}{0} & \MoraScore{0.167}{0} & \MoraScore{0.255}{0} & \MoraScore{0.204}{0} & \MoraScore{0.294}{0} & \MoraScore{0.209}{0} & \MoraScore{0.295}{0} & \MoraScore{0.164}{0} & \MoraScore{0.252}{0} \\
 & \textcolor{MoraAccent}{\textbf{+MORA}} & \MoraScore{0.156}{1} & \MoraScore{0.256}{1} & \MoraScore{0.160}{1} & \MoraScore{0.259}{1} & \MoraScore{0.159}{1} & \MoraScore{0.252}{1} & \MoraScore{0.174}{1} & \MoraScore{0.275}{1} & \MoraScore{0.177}{1} & \MoraScore{0.277}{1} & \MoraScore{0.156}{1} & \MoraScore{0.250}{1} \\
\rowcolor{MoraTint}
\multirow{-3}{*}{\textbf{ECL}} & \MoraEntry{Imp.\ (\%)} & \MoraImp{4.827} & \MoraImp{1.943} & \MoraImp{8.571} & \MoraImp{2.991} & \MoraImp{4.639} & \MoraImp{1.026} & \MoraImp{14.626} & \MoraImp{6.630} & \MoraImp{15.337} & \MoraImp{6.132} & \MoraImp{4.784} & \MoraImp{1.158} \\
\MoraGroupRule
 & \MoraEntry{Base} & \MoraScore{0.408}{0} & \MoraScore{0.269}{0} & \MoraScore{0.423}{0} & \MoraScore{0.282}{0} & \MoraScore{0.430}{0} & \MoraScore{0.257}{0} & \MoraScore{0.481}{0} & \MoraScore{0.308}{0} & \MoraScore{0.625}{0} & \MoraScore{0.385}{0} & \MoraScore{0.421}{0} & \MoraScore{0.274}{0} \\
 & \textcolor{MoraAccent}{\textbf{+MORA}} & \MoraScore{0.400}{1} & \MoraScore{0.266}{1} & \MoraScore{0.411}{1} & \MoraScore{0.277}{1} & \MoraScore{0.424}{1} & \MoraScore{0.256}{1} & \MoraScore{0.457}{1} & \MoraScore{0.297}{1} & \MoraScore{0.537}{1} & \MoraScore{0.343}{1} & \MoraScore{0.413}{1} & \MoraScore{0.271}{1} \\
\rowcolor{MoraTint}
\multirow{-3}{*}{\textbf{Traffic}} & \MoraEntry{Imp.\ (\%)} & \MoraImp{1.931} & \MoraImp{1.051} & \MoraImp{2.767} & \MoraImp{1.837} & \MoraImp{1.599} & \MoraImp{0.085} & \MoraImp{5.146} & \MoraImp{3.604} & \MoraImp{14.123} & \MoraImp{10.696} & \MoraImp{1.886} & \MoraImp{0.883} \\
\MoraGroupRule
 & \MoraEntry{Base} & \MoraScore{0.227}{0} & \MoraScore{0.258}{0} & \MoraScore{0.235}{0} & \MoraScore{0.262}{0} & \MoraScore{0.226}{0} & \MoraScore{0.231}{0} & \MoraScore{0.259}{0} & \MoraScore{0.300}{0} & \MoraScore{0.330}{0} & \MoraScore{0.401}{0} & \MoraScore{0.204}{0} & \MoraScore{0.211}{1} \\
 & \textcolor{MoraAccent}{\textbf{+MORA}} & \MoraScore{0.198}{1} & \MoraScore{0.250}{1} & \MoraScore{0.198}{1} & \MoraScore{0.251}{1} & \MoraScore{0.210}{1} & \MoraScore{0.230}{1} & \MoraScore{0.205}{1} & \MoraScore{0.270}{1} & \MoraScore{0.212}{1} & \MoraScore{0.278}{1} & \MoraScore{0.199}{1} & \MoraScore{0.212}{0} \\
\rowcolor{MoraTint}
\multirow{-3}{*}{\textbf{Solar}} & \MoraEntry{Imp.\ (\%)} & \MoraImp{12.910} & \MoraImp{3.043} & \MoraImp{15.829} & \MoraImp{4.282} & \MoraImp{6.917} & \MoraImp{0.253} & \MoraImp{20.638} & \MoraImp{10.044} & \MoraImp{35.827} & \MoraImp{30.677} & \MoraImp{2.557} & \MoraImp{-0.369} \\
\bottomrule[0.8pt]
\end{tabular*}%
}
\endgroup
\end{table}

\begin{table}[!tp]
\centering
\begingroup
\caption{Baseline and MORA results for six backbones on the four
PEMS datasets. MSE and MAE are averaged over
$H\in\{12,24,48,96\}$ at $L=96$.}
\label{tab:mora_pems_main}
\fontsize{9}{10.2}\selectfont
\setlength{\tabcolsep}{1.5pt}
\renewcommand{\arraystretch}{0.90}
\setlength{\aboverulesep}{1.2pt}
\setlength{\belowrulesep}{1.5pt}
\resizebox{0.90\linewidth}{!}{%
\begin{tabular*}{\linewidth}{@{}l@{\hspace{4pt}}l@{\extracolsep{\fill}\hspace{5.5pt}}r@{\extracolsep{0pt}\hspace{3pt}}r@{\extracolsep{\fill}\hspace{5.5pt}}r@{\extracolsep{0pt}\hspace{3pt}}r@{\extracolsep{\fill}\hspace{5.5pt}}r@{\extracolsep{0pt}\hspace{3pt}}r@{\extracolsep{\fill}\hspace{5.5pt}}r@{\extracolsep{0pt}\hspace{3pt}}r@{\extracolsep{\fill}\hspace{5.5pt}}r@{\extracolsep{0pt}\hspace{3pt}}r@{\extracolsep{\fill}\hspace{5.5pt}}r@{\extracolsep{0pt}\hspace{3pt}}r@{}}
\toprule[0.8pt]
\multicolumn{2}{l}{\textbf{Dataset}} & \multicolumn{2}{c}{\textbf{TimeFilter}} & \multicolumn{2}{c}{\textbf{iTransformer}} & \multicolumn{2}{c}{\textbf{CGTFra}} & \multicolumn{2}{c}{\textbf{PatchTST}} & \multicolumn{2}{c}{\textbf{DLinear}} & \multicolumn{2}{c}{\textbf{VPNet}} \\
\cmidrule(lr){3-4}\cmidrule(lr){5-6}\cmidrule(lr){7-8}\cmidrule(lr){9-10}\cmidrule(lr){11-12}\cmidrule(lr){13-14}
 & & \MoraMetric{MSE} & \MoraMetric{MAE} & \MoraMetric{MSE} & \MoraMetric{MAE} & \MoraMetric{MSE} & \MoraMetric{MAE} & \MoraMetric{MSE} & \MoraMetric{MAE} & \MoraMetric{MSE} & \MoraMetric{MAE} & \MoraMetric{MSE} & \MoraMetric{MAE} \\
\midrule
 & \MoraEntry{Base} & \MoraScore{0.101}{0} & \MoraScore{0.210}{0} & \MoraScore{0.115}{0} & \MoraScore{0.224}{0} & \MoraScore{0.109}{0} & \MoraScore{0.213}{0} & \MoraScore{0.206}{0} & \MoraScore{0.295}{0} & \MoraScore{0.245}{0} & \MoraScore{0.341}{0} & \MoraScore{0.117}{0} & \MoraScore{0.227}{0} \\
 & \textcolor{MoraAccent}{\textbf{+MORA}} & \MoraScore{0.098}{1} & \MoraScore{0.206}{1} & \MoraScore{0.114}{1} & \MoraScore{0.216}{1} & \MoraScore{0.099}{1} & \MoraScore{0.203}{1} & \MoraScore{0.127}{1} & \MoraScore{0.232}{1} & \MoraScore{0.137}{1} & \MoraScore{0.241}{1} & \MoraScore{0.107}{1} & \MoraScore{0.215}{1} \\
\rowcolor{MoraTint}
\multirow{-3}{*}{\textbf{PEMS03}} & \MoraEntry{Imp.\ (\%)} & \MoraImp{3.019} & \MoraImp{2.008} & \MoraImp{1.178} & \MoraImp{3.494} & \MoraImp{9.246} & \MoraImp{4.928} & \MoraImp{38.291} & \MoraImp{21.592} & \MoraImp{44.103} & \MoraImp{29.397} & \MoraImp{8.201} & \MoraImp{5.479} \\
\MoraGroupRule
 & \MoraEntry{Base} & \MoraScore{0.095}{0} & \MoraScore{0.201}{0} & \MoraScore{0.113}{0} & \MoraScore{0.223}{0} & \MoraScore{0.107}{0} & \MoraScore{0.210}{0} & \MoraScore{0.256}{0} & \MoraScore{0.333}{0} & \MoraScore{0.246}{0} & \MoraScore{0.350}{0} & \MoraScore{0.111}{0} & \MoraScore{0.224}{0} \\
 & \textcolor{MoraAccent}{\textbf{+MORA}} & \MoraScore{0.079}{1} & \MoraScore{0.180}{1} & \MoraScore{0.090}{1} & \MoraScore{0.193}{1} & \MoraScore{0.082}{1} & \MoraScore{0.183}{1} & \MoraScore{0.097}{1} & \MoraScore{0.196}{1} & \MoraScore{0.105}{1} & \MoraScore{0.201}{1} & \MoraScore{0.084}{1} & \MoraScore{0.187}{1} \\
\rowcolor{MoraTint}
\multirow{-3}{*}{\textbf{PEMS04}} & \MoraEntry{Imp.\ (\%)} & \MoraImp{16.706} & \MoraImp{10.467} & \MoraImp{20.277} & \MoraImp{13.700} & \MoraImp{23.066} & \MoraImp{12.876} & \MoraImp{62.321} & \MoraImp{41.041} & \MoraImp{57.482} & \MoraImp{42.593} & \MoraImp{24.279} & \MoraImp{16.433} \\
\MoraGroupRule
 & \MoraEntry{Base} & \MoraScore{0.084}{0} & \MoraScore{0.184}{0} & \MoraScore{0.101}{0} & \MoraScore{0.204}{0} & \MoraScore{0.096}{0} & \MoraScore{0.189}{0} & \MoraScore{0.220}{0} & \MoraScore{0.301}{0} & \MoraScore{0.288}{0} & \MoraScore{0.381}{0} & \MoraScore{0.101}{0} & \MoraScore{0.211}{0} \\
 & \textcolor{MoraAccent}{\textbf{+MORA}} & \MoraScore{0.071}{1} & \MoraScore{0.167}{1} & \MoraScore{0.091}{1} & \MoraScore{0.178}{1} & \MoraScore{0.078}{1} & \MoraScore{0.166}{1} & \MoraScore{0.086}{1} & \MoraScore{0.185}{1} & \MoraScore{0.098}{1} & \MoraScore{0.197}{1} & \MoraScore{0.076}{1} & \MoraScore{0.176}{1} \\
\rowcolor{MoraTint}
\multirow{-3}{*}{\textbf{PEMS07}} & \MoraEntry{Imp.\ (\%)} & \MoraImp{15.459} & \MoraImp{9.520} & \MoraImp{10.030} & \MoraImp{12.593} & \MoraImp{18.250} & \MoraImp{12.044} & \MoraImp{60.714} & \MoraImp{38.640} & \MoraImp{65.851} & \MoraImp{48.317} & \MoraImp{24.449} & \MoraImp{16.480} \\
\MoraGroupRule
 & \MoraEntry{Base} & \MoraScore{0.098}{0} & \MoraScore{0.203}{0} & \MoraScore{0.152}{0} & \MoraScore{0.227}{0} & \MoraScore{0.145}{0} & \MoraScore{0.226}{0} & \MoraScore{0.210}{0} & \MoraScore{0.296}{0} & \MoraScore{0.239}{0} & \MoraScore{0.346}{0} & \MoraScore{0.106}{0} & \MoraScore{0.215}{0} \\
 & \textcolor{MoraAccent}{\textbf{+MORA}} & \MoraScore{0.085}{1} & \MoraScore{0.187}{1} & \MoraScore{0.129}{1} & \MoraScore{0.210}{1} & \MoraScore{0.110}{1} & \MoraScore{0.201}{1} & \MoraScore{0.105}{1} & \MoraScore{0.209}{1} & \MoraScore{0.114}{1} & \MoraScore{0.221}{1} & \MoraScore{0.088}{1} & \MoraScore{0.193}{1} \\
\rowcolor{MoraTint}
\multirow{-3}{*}{\textbf{PEMS08}} & \MoraEntry{Imp.\ (\%)} & \MoraImp{13.234} & \MoraImp{7.754} & \MoraImp{14.797} & \MoraImp{7.281} & \MoraImp{23.796} & \MoraImp{11.092} & \MoraImp{50.224} & \MoraImp{29.300} & \MoraImp{52.499} & \MoraImp{36.130} & \MoraImp{16.811} & \MoraImp{10.205} \\
\bottomrule[0.8pt]
\end{tabular*}%
}
\endgroup
\end{table}

\begin{table}[!tp]
\centering
\begingroup
\caption{Baseline and MORA results for three frozen time-series
foundation models at $L=96$ and $L=512$. MSE and MAE are averaged
over $H\in\{96,192,336,720\}$ separately for each context length.}
\label{tab:mora_foundation_main}
\fontsize{9}{10.2}\selectfont
\setlength{\tabcolsep}{1.5pt}
\renewcommand{\arraystretch}{0.90}
\setlength{\aboverulesep}{1.2pt}
\setlength{\belowrulesep}{1.5pt}
\resizebox{0.90\linewidth}{!}{%
\begin{tabular*}{\linewidth}{@{}l@{\hspace{4pt}}l@{\extracolsep{\fill}\hspace{5.5pt}}r@{\extracolsep{0pt}\hspace{3pt}}r@{\extracolsep{0pt}\hspace{5pt}}r@{\extracolsep{0pt}\hspace{3pt}}r@{\extracolsep{\fill}\hspace{5.5pt}}r@{\extracolsep{0pt}\hspace{3pt}}r@{\extracolsep{0pt}\hspace{5pt}}r@{\extracolsep{0pt}\hspace{3pt}}r@{\extracolsep{\fill}\hspace{5.5pt}}r@{\extracolsep{0pt}\hspace{3pt}}r@{\extracolsep{0pt}\hspace{5pt}}r@{\extracolsep{0pt}\hspace{3pt}}r@{}}
\toprule[0.8pt]
\multicolumn{2}{l}{\textbf{Dataset}} & \multicolumn{4}{c}{\textbf{Chronos-2}} & \multicolumn{4}{c}{\textbf{Sundial}} & \multicolumn{4}{c}{\textbf{TimesFM 2.5}} \\
\cmidrule(lr){3-6}\cmidrule(lr){7-10}\cmidrule(lr){11-14}
 & & \multicolumn{2}{c}{\MoraMetric{$L=96$}} & \multicolumn{2}{c}{\MoraMetric{$L=512$}} & \multicolumn{2}{c}{\MoraMetric{$L=96$}} & \multicolumn{2}{c}{\MoraMetric{$L=512$}} & \multicolumn{2}{c}{\MoraMetric{$L=96$}} & \multicolumn{2}{c}{\MoraMetric{$L=512$}} \\
\cmidrule(lr){3-4}\cmidrule(lr){5-6}\cmidrule(lr){7-8}\cmidrule(lr){9-10}\cmidrule(lr){11-12}\cmidrule(lr){13-14}
 & & \MoraMetric{MSE} & \MoraMetric{MAE} & \MoraMetric{MSE} & \MoraMetric{MAE} & \MoraMetric{MSE} & \MoraMetric{MAE} & \MoraMetric{MSE} & \MoraMetric{MAE} & \MoraMetric{MSE} & \MoraMetric{MAE} & \MoraMetric{MSE} & \MoraMetric{MAE} \\
\midrule
 & \MoraEntry{Base} & \MoraScore{0.492}{0} & \MoraScore{0.439}{0} & \MoraScore{0.418}{0} & \MoraScore{0.412}{1} & \MoraScore{0.473}{0} & \MoraScore{0.435}{0} & \MoraScore{0.397}{0} & \MoraScore{0.412}{0} & \MoraScore{0.484}{0} & \MoraScore{0.435}{0} & \MoraScore{0.421}{0} & \MoraScore{0.409}{1} \\
 & \textcolor{MoraAccent}{\textbf{+MORA}} & \MoraScore{0.456}{1} & \MoraScore{0.431}{1} & \MoraScore{0.402}{1} & \MoraScore{0.415}{0} & \MoraScore{0.436}{1} & \MoraScore{0.428}{1} & \MoraScore{0.387}{1} & \MoraScore{0.409}{1} & \MoraScore{0.435}{1} & \MoraScore{0.429}{1} & \MoraScore{0.403}{1} & \MoraScore{0.412}{0} \\
\rowcolor{MoraTint}
\multirow{-3}{*}{\textbf{ETTh1}} & \MoraEntry{Imp.\ (\%)} & \MoraImp{7.370} & \MoraImp{1.880} & \MoraImp{3.900} & \MoraImp{-0.753} & \MoraImp{7.739} & \MoraImp{1.610} & \MoraImp{2.463} & \MoraImp{0.682} & \MoraImp{10.157} & \MoraImp{1.439} & \MoraImp{4.437} & \MoraImp{-0.866} \\
\MoraGroupRule
 & \MoraEntry{Base} & \MoraScore{0.406}{0} & \MoraScore{0.405}{0} & \MoraScore{0.352}{0} & \MoraScore{0.377}{0} & \MoraScore{0.399}{0} & \MoraScore{0.410}{0} & \MoraScore{0.344}{0} & \MoraScore{0.383}{0} & \MoraScore{0.413}{0} & \MoraScore{0.406}{0} & \MoraScore{0.372}{0} & \MoraScore{0.388}{0} \\
 & \textcolor{MoraAccent}{\textbf{+MORA}} & \MoraScore{0.384}{1} & \MoraScore{0.393}{1} & \MoraScore{0.345}{1} & \MoraScore{0.377}{0} & \MoraScore{0.389}{1} & \MoraScore{0.405}{1} & \MoraScore{0.340}{1} & \MoraScore{0.378}{1} & \MoraScore{0.397}{1} & \MoraScore{0.404}{1} & \MoraScore{0.365}{1} & \MoraScore{0.384}{1} \\
\rowcolor{MoraTint}
\multirow{-3}{*}{\textbf{ETTh2}} & \MoraEntry{Imp.\ (\%)} & \MoraImp{5.413} & \MoraImp{3.038} & \MoraImp{2.031} & \MoraImp{0.005} & \MoraImp{2.598} & \MoraImp{1.068} & \MoraImp{1.001} & \MoraImp{1.170} & \MoraImp{3.895} & \MoraImp{0.512} & \MoraImp{1.898} & \MoraImp{1.184} \\
\MoraGroupRule
 & \MoraEntry{Base} & \MoraScore{0.616}{0} & \MoraScore{0.477}{0} & \MoraScore{0.387}{0} & \MoraScore{0.372}{0} & \MoraScore{0.537}{0} & \MoraScore{0.457}{0} & \MoraScore{0.366}{0} & \MoraScore{0.370}{0} & \MoraScore{0.875}{0} & \MoraScore{0.584}{0} & \MoraScore{0.381}{0} & \MoraScore{0.372}{0} \\
 & \textcolor{MoraAccent}{\textbf{+MORA}} & \MoraScore{0.366}{1} & \MoraScore{0.391}{1} & \MoraScore{0.338}{1} & \MoraScore{0.360}{1} & \MoraScore{0.369}{1} & \MoraScore{0.390}{1} & \MoraScore{0.338}{1} & \MoraScore{0.365}{1} & \MoraScore{0.370}{1} & \MoraScore{0.390}{1} & \MoraScore{0.341}{1} & \MoraScore{0.366}{1} \\
\rowcolor{MoraTint}
\multirow{-3}{*}{\textbf{ETTm1}} & \MoraEntry{Imp.\ (\%)} & \MoraImp{40.496} & \MoraImp{18.127} & \MoraImp{12.787} & \MoraImp{3.163} & \MoraImp{31.329} & \MoraImp{14.632} & \MoraImp{7.521} & \MoraImp{1.389} & \MoraImp{57.682} & \MoraImp{33.247} & \MoraImp{10.354} & \MoraImp{1.629} \\
\MoraGroupRule
 & \MoraEntry{Base} & \MoraScore{0.318}{0} & \MoraScore{0.338}{0} & \MoraScore{0.280}{0} & \MoraScore{0.311}{0} & \MoraScore{0.304}{0} & \MoraScore{0.340}{0} & \MoraScore{0.275}{0} & \MoraScore{0.320}{0} & \MoraScore{0.372}{0} & \MoraScore{0.378}{0} & \MoraScore{0.291}{0} & \MoraScore{0.318}{0} \\
 & \textcolor{MoraAccent}{\textbf{+MORA}} & \MoraScore{0.275}{1} & \MoraScore{0.315}{1} & \MoraScore{0.257}{1} & \MoraScore{0.299}{1} & \MoraScore{0.271}{1} & \MoraScore{0.320}{1} & \MoraScore{0.252}{1} & \MoraScore{0.308}{1} & \MoraScore{0.277}{1} & \MoraScore{0.321}{1} & \MoraScore{0.273}{1} & \MoraScore{0.309}{1} \\
\rowcolor{MoraTint}
\multirow{-3}{*}{\textbf{ETTm2}} & \MoraEntry{Imp.\ (\%)} & \MoraImp{13.376} & \MoraImp{6.946} & \MoraImp{8.182} & \MoraImp{3.574} & \MoraImp{10.916} & \MoraImp{5.858} & \MoraImp{8.162} & \MoraImp{3.754} & \MoraImp{25.500} & \MoraImp{14.896} & \MoraImp{6.079} & \MoraImp{2.624} \\
\MoraGroupRule
 & \MoraEntry{Base} & \MoraScore{0.307}{0} & \MoraScore{0.291}{0} & \MoraScore{0.258}{0} & \MoraScore{0.252}{0} & \MoraScore{0.285}{0} & \MoraScore{0.302}{0} & \MoraScore{0.244}{0} & \MoraScore{0.275}{0} & \MoraScore{0.314}{0} & \MoraScore{0.299}{0} & \MoraScore{0.263}{0} & \MoraScore{0.268}{0} \\
 & \textcolor{MoraAccent}{\textbf{+MORA}} & \MoraScore{0.243}{1} & \MoraScore{0.290}{1} & \MoraScore{0.213}{1} & \MoraScore{0.250}{1} & \MoraScore{0.233}{1} & \MoraScore{0.291}{1} & \MoraScore{0.213}{1} & \MoraScore{0.262}{1} & \MoraScore{0.230}{1} & \MoraScore{0.283}{1} & \MoraScore{0.215}{1} & \MoraScore{0.249}{1} \\
\rowcolor{MoraTint}
\multirow{-3}{*}{\textbf{Weather}} & \MoraEntry{Imp.\ (\%)} & \MoraImp{20.807} & \MoraImp{0.330} & \MoraImp{17.467} & \MoraImp{0.761} & \MoraImp{18.407} & \MoraImp{3.803} & \MoraImp{12.890} & \MoraImp{4.853} & \MoraImp{26.704} & \MoraImp{5.584} & \MoraImp{18.340} & \MoraImp{6.779} \\
\MoraGroupRule
 & \MoraEntry{Base} & \MoraScore{0.234}{0} & \MoraScore{0.294}{0} & \MoraScore{0.163}{0} & \MoraScore{0.244}{0} & \MoraScore{0.204}{0} & \MoraScore{0.284}{0} & \MoraScore{0.148}{0} & \MoraScore{0.243}{0} & \MoraScore{0.239}{0} & \MoraScore{0.308}{0} & \MoraScore{0.164}{0} & \MoraScore{0.249}{0} \\
 & \textcolor{MoraAccent}{\textbf{+MORA}} & \MoraScore{0.184}{1} & \MoraScore{0.274}{1} & \MoraScore{0.154}{1} & \MoraScore{0.241}{1} & \MoraScore{0.173}{1} & \MoraScore{0.270}{1} & \MoraScore{0.143}{1} & \MoraScore{0.240}{1} & \MoraScore{0.187}{1} & \MoraScore{0.283}{1} & \MoraScore{0.156}{1} & \MoraScore{0.246}{1} \\
\rowcolor{MoraTint}
\multirow{-3}{*}{\textbf{ECL}} & \MoraEntry{Imp.\ (\%)} & \MoraImp{21.589} & \MoraImp{6.881} & \MoraImp{5.026} & \MoraImp{1.022} & \MoraImp{14.910} & \MoraImp{4.971} & \MoraImp{2.899} & \MoraImp{0.979} & \MoraImp{22.001} & \MoraImp{8.226} & \MoraImp{4.987} & \MoraImp{1.161} \\
\MoraGroupRule
 & \MoraEntry{Base} & \MoraScore{0.574}{0} & \MoraScore{0.313}{0} & \MoraScore{0.392}{0} & \MoraScore{0.238}{1} & \MoraScore{0.604}{0} & \MoraScore{0.355}{0} & \MoraScore{0.458}{0} & \MoraScore{0.290}{0} & \MoraScore{0.599}{0} & \MoraScore{0.334}{0} & \MoraScore{0.382}{0} & \MoraScore{0.244}{0} \\
 & \textcolor{MoraAccent}{\textbf{+MORA}} & \MoraScore{0.518}{1} & \MoraScore{0.301}{1} & \MoraScore{0.389}{1} & \MoraScore{0.239}{0} & \MoraScore{0.531}{1} & \MoraScore{0.327}{1} & \MoraScore{0.441}{1} & \MoraScore{0.283}{1} & \MoraScore{0.544}{1} & \MoraScore{0.316}{1} & \MoraScore{0.379}{1} & \MoraScore{0.244}{0} \\
\rowcolor{MoraTint}
\multirow{-3}{*}{\textbf{Traffic}} & \MoraEntry{Imp.\ (\%)} & \MoraImp{9.796} & \MoraImp{3.687} & \MoraImp{0.846} & \MoraImp{-0.171} & \MoraImp{12.143} & \MoraImp{7.929} & \MoraImp{3.734} & \MoraImp{2.131} & \MoraImp{9.119} & \MoraImp{5.503} & \MoraImp{0.748} & \MoraImp{0.082} \\
\MoraGroupRule
 & \MoraEntry{Base} & \MoraScore{0.528}{0} & \MoraScore{0.327}{0} & \MoraScore{0.187}{0} & \MoraScore{0.193}{1} & \MoraScore{0.666}{0} & \MoraScore{0.530}{0} & \MoraScore{0.229}{0} & \MoraScore{0.257}{0} & \MoraScore{0.913}{0} & \MoraScore{0.520}{0} & \MoraScore{0.281}{0} & \MoraScore{0.250}{1} \\
 & \textcolor{MoraAccent}{\textbf{+MORA}} & \MoraScore{0.215}{1} & \MoraScore{0.286}{1} & \MoraScore{0.187}{0} & \MoraScore{0.194}{0} & \MoraScore{0.210}{1} & \MoraScore{0.278}{1} & \MoraScore{0.208}{1} & \MoraScore{0.256}{1} & \MoraScore{0.222}{1} & \MoraScore{0.279}{1} & \MoraScore{0.222}{1} & \MoraScore{0.261}{0} \\
\rowcolor{MoraTint}
\multirow{-3}{*}{\textbf{Solar}} & \MoraEntry{Imp.\ (\%)} & \MoraImp{59.281} & \MoraImp{12.521} & \MoraImp{0.174} & \MoraImp{-0.196} & \MoraImp{68.421} & \MoraImp{47.491} & \MoraImp{9.212} & \MoraImp{0.170} & \MoraImp{75.707} & \MoraImp{46.260} & \MoraImp{21.222} & \MoraImp{-4.503} \\
\bottomrule[0.8pt]
\end{tabular*}%
}
\endgroup
\end{table}

\paragraph{Main Results.}
Tables~\ref{tab:mora_backbones_main} and~\ref{tab:mora_pems_main}
report horizon-averaged results on long- and short-term forecasting,
respectively. Across the 192 dataset--backbone--horizon settings on
the long-term benchmarks, MORA reduces MSE in 187 cases (97.4\%).
Averaged across the eight datasets, MORA lowers the MSE of DLinear
and PatchTST by 14.9\% and 9.3\%, respectively. The gains remain
evident for more competitive recent forecasters, with average MSE
reductions of 4.4\% for TimeFilter and 4.3\% for CGTFra. Overall,
across all six backbones, MORA reduces MSE by 7.2\% on average.

\label{sec:pems_results}
Similar improvements are observed on the four short-term PEMS
benchmarks, where MORA improves both MSE and MAE for all 24
dataset--backbone combinations in the horizon-averaged results.
Across the six backbones, average MSE reductions range from 11.6\%
to 55.0\%. Notably, the benefit persists for TimeFilter, a strong
recent backbone: MORA further reduces its MSE by 3.0\%--16.7\%
across the four datasets, averaging 12.1\% relative to the same
frozen checkpoints. Together, these results show that MORA
complements diverse forecasting architectures in both long- and
short-term settings, including strong backbones, without updating
their weights.

\paragraph{Time-Series Foundation Models.}
\label{sec:foundation_results}
Table~\ref{tab:mora_foundation_main} reports the results of MORA on
three time-series foundation models using their officially released
pretrained checkpoints. Since all three models support substantially
longer context windows at inference, we evaluate $L=512$ in addition
to $L=96$ to examine whether MORA remains effective when the backbone
is provided with more historical information. Each adapted forecast
is compared against its unadapted counterpart using the same context
length. At $L=96$, MORA reduces MSE by 22.3\%, 20.8\%, and 28.8\%
for Chronos-2, Sundial, and TimesFM~2.5, respectively, averaged
equally across the eight datasets. Even at $L=512$, where the
foundation models can draw on substantially more historical context,
MORA yields further average MSE reductions of 6.3\%, 6.0\%, and
8.5\% for the same three models. MORA also reduces MAE on average at
both context lengths. Together with the results on task-specific
forecasters, these findings show that MORA provides a broadly
applicable output-space refinement mechanism whose benefits persist
across model families and context lengths.
\FloatBarrier

\subsection{Ablation Study}
\label{sec:mora_ablations}

We first conduct component ablations to examine the roles of MORA's
coarse future estimates and stagewise residual updates. We then compare
alternative adaptation strategies to assess the benefit of deriving
residual corrections from coarse future estimates.

\begin{table}[!htbp]
\centering
\begin{minipage}[t]{0.485\linewidth}
\vspace{0pt}
\caption{Component ablations on TimeFilter.}
\label{tab:mora_ablation_components}
\begingroup
\fontsize{8}{9.5}\selectfont
\renewcommand{\arraystretch}{0.95}
\setlength{\aboverulesep}{0.7pt}
\setlength{\belowrulesep}{0.7pt}
\setlength{\cmidrulekern}{2pt}
\ifdefined\MoraAblWidthBox\else\newsavebox{\MoraAblWidthBox}\fi
\def\MoraAblBody{%
\begin{tabular}{@{}lrrrrrr@{}}
\toprule[0.65pt]
 & \multicolumn{2}{c}{\textbf{ETTm1}} & \multicolumn{2}{c}{\textbf{ETTm2}} & \multicolumn{2}{c}{\textbf{ECL}} \\
\cmidrule(lr){2-3}\cmidrule(lr){4-5}\cmidrule(lr){6-7}
Variant & \textcolor{MoraMuted}{MSE} & \textcolor{MoraMuted}{MAE} & \textcolor{MoraMuted}{MSE} & \textcolor{MoraMuted}{MAE} & \textcolor{MoraMuted}{MSE} & \textcolor{MoraMuted}{MAE} \\
\midrule[0.35pt]
TimeFilter & 0.452 & 0.437 & 0.390 & 0.393 & 0.206 & 0.302 \\
\midrule[0.35pt]
w/o learned & 0.427 & 0.431 & 0.389 & 0.390 & 0.194 & 0.291 \\
w/o extrapolation & 0.436 & 0.427 & 0.380 & 0.391 & 0.195 & 0.294 \\
\midrule[0.35pt]
Independent & 0.427 & 0.429 & 0.379 & 0.389 & 0.191 & 0.290 \\
\midrule[0.35pt]
\rowcolor{MoraTint}
\textbf{MORA} & \textbf{0.422} & \textbf{0.426} & \textbf{0.376} & \textbf{0.388} & \textbf{0.190} & \textbf{0.288} \\
\bottomrule[0.65pt]
\end{tabular}
}
\sbox{\MoraAblWidthBox}{\setlength{\tabcolsep}{0pt}\MoraAblBody}
\setlength{\tabcolsep}{\dimexpr(\linewidth-\wd\MoraAblWidthBox)/12\relax}
\ifdim\tabcolsep<1.2pt
  \PackageError{MORAablation}{Ablation table is too crowded}{Increase panel width or shorten labels rather than scaling text.}
\fi
\MoraAblBody\par
\endgroup

\end{minipage}\hfill
\begin{minipage}[t]{0.485\linewidth}
\vspace{0pt}
\caption{Alternative strategies on TimeFilter.}
\label{tab:mora_ablation_posthoc}
\begingroup
\fontsize{8}{9.5}\selectfont
\renewcommand{\arraystretch}{0.95}
\setlength{\aboverulesep}{0.7pt}
\setlength{\belowrulesep}{0.7pt}
\setlength{\cmidrulekern}{2pt}
\ifdefined\MoraAblWidthBox\else\newsavebox{\MoraAblWidthBox}\fi
\def\MoraAblBody{%
\begin{tabular}{@{}lrrrrrr@{}}
\toprule[0.65pt]
 & \multicolumn{2}{c}{\textbf{ETTm1}} & \multicolumn{2}{c}{\textbf{Weather}} & \multicolumn{2}{c}{\textbf{Solar}} \\
\cmidrule(lr){2-3}\cmidrule(lr){4-5}\cmidrule(lr){6-7}
Strategy & \textcolor{MoraMuted}{MSE} & \textcolor{MoraMuted}{MAE} & \textcolor{MoraMuted}{MSE} & \textcolor{MoraMuted}{MAE} & \textcolor{MoraMuted}{MSE} & \textcolor{MoraMuted}{MAE} \\
\midrule[0.35pt]
TimeFilter & 0.313 & 0.354 & 0.153 & 0.199 & 0.199 & 0.239 \\
\midrule[0.35pt]
Affine calibration & 0.314 & 0.357 & 0.153 & 0.228 & 0.200 & 0.239 \\
Direct residual & \textbf{0.300} & 0.357 & 0.150 & 0.211 & 0.181 & 0.239 \\
$\delta$-Adapter (O) & 0.370 & 0.415 & 0.156 & 0.224 & 0.192 & 0.243 \\
$\delta$-Adapter (I+O) & 0.312 & 0.356 & 0.153 & 0.201 & \textbf{0.171} & 0.234 \\
\midrule[0.35pt]
\rowcolor{MoraTint}
\textbf{MORA} & 0.304 & \textbf{0.350} & \textbf{0.148} & \textbf{0.198} & 0.173 & \textbf{0.231} \\
\bottomrule[0.65pt]
\end{tabular}
}
\sbox{\MoraAblWidthBox}{\setlength{\tabcolsep}{0pt}\MoraAblBody}
\setlength{\tabcolsep}{\dimexpr(\linewidth-\wd\MoraAblWidthBox)/12\relax}
\ifdim\tabcolsep<1.2pt
  \PackageError{MORAablation}{Ablation table is too crowded}{Increase panel width or shorten labels rather than scaling text.}
\fi
\MoraAblBody\par
\vspace{2pt}
{\scriptsize O: output-only; I+O: input--output.\par}
\endgroup

\end{minipage}
\end{table}

\paragraph{Component Ablations.}
Table~\ref{tab:mora_ablation_components} examines the contributions
of the two sources of coarse future estimates and the design of the
second-stage correction. Removing either \emph{structural extrapolation} or
the \emph{learned estimator} increases both MSE and MAE on all three
datasets. Although both sources target the same coarse future
quantities, neither alone matches the performance of the full adapter,
indicating that they provide complementary correction signals.
With both sources retained, we compare the stagewise design against
an \emph{independent-update variant} in which both corrections are fitted
to the original backbone residual. The stagewise design achieves
slightly lower MSE and MAE on all three datasets. This result suggests
that the learned correction benefits from accounting for the
first-stage correction, allowing it to target the remaining residual
rather than fitting another correction to the original backbone
residual.

\paragraph{Comparison with Alternative Strategies.}
Table~\ref{tab:mora_ablation_posthoc} compares MORA with \emph{affine
calibration}, \emph{direct residual regression}, and two variants of
$\delta$-Adapter~\citep{liang2026the}. The direct regressor uses the
same historical features and correction basis as MORA, with a
comparable parameter count, but predicts the backbone residual
directly. Affine calibration improves neither MSE nor MAE. The
learned alternatives reduce MSE in several settings, but their gains
do not consistently extend to MAE. Direct residual regression
achieves the lowest MSE on ETTm1, while $\delta$-Adapter (I+O) does
so on Solar. MORA achieves the lowest MAE on all three datasets while
reducing MSE relative to the unadapted backbone in every case. These
results suggest that deriving corrections from coarse future
estimates provides an effective adaptation strategy, yielding
consistent improvements in both metrics across the evaluated
settings. Additional ablations and experimental details are provided in
Appendix~\ref{app:mora_ablation}.
\FloatBarrier

\subsection{Low-Frequency Residual Analysis}
\label{sec:mora_residual_visualization}

To examine whether MORA effectively corrects low-frequency
forecasting errors, we visualize the low-frequency residuals of
DLinear and TimeFilter on Solar before and after adaptation.
As shown in Figure~\ref{fig:mora_residual_spectrum}, MORA reduces
low-frequency reconstruction MSE by 41.0\% and 13.9\%, respectively.
The corresponding overall forecasting MSE decreases by 38.9\%
and 12.8\%. The reduction in low-frequency error accounts for
nearly all of the overall MSE improvement in these settings,
supporting the intended role of MORA's coarse future estimates
in correcting coarse-scale deviations.

\begingroup
\setlength{\intextsep}{4pt}
\begin{figure}[!htbp]
    \centering
    \includegraphics[width=\linewidth]{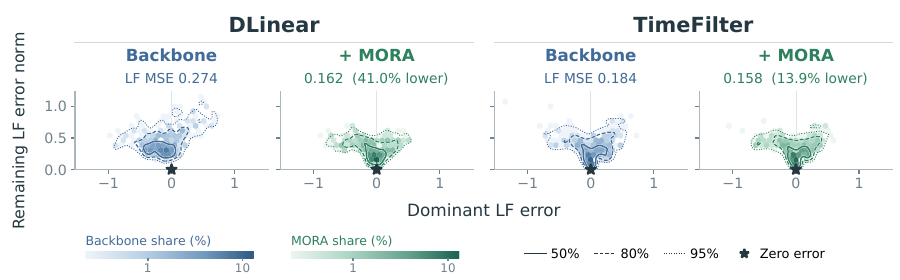}
    \caption{\textbf{Low-frequency residuals before and after applying MORA on Solar.}
    DCT-8 residuals are represented by their signed projection onto a shared principal direction fitted on the validation set, together with the norm of the orthogonal residual. The squared distance from the origin corresponds to the low-frequency MSE.}
    \label{fig:mora_residual_spectrum}
\end{figure}

\FloatBarrier
\endgroup

\begin{figure}[!htbp]
    \centering
    \includegraphics[width=\linewidth]{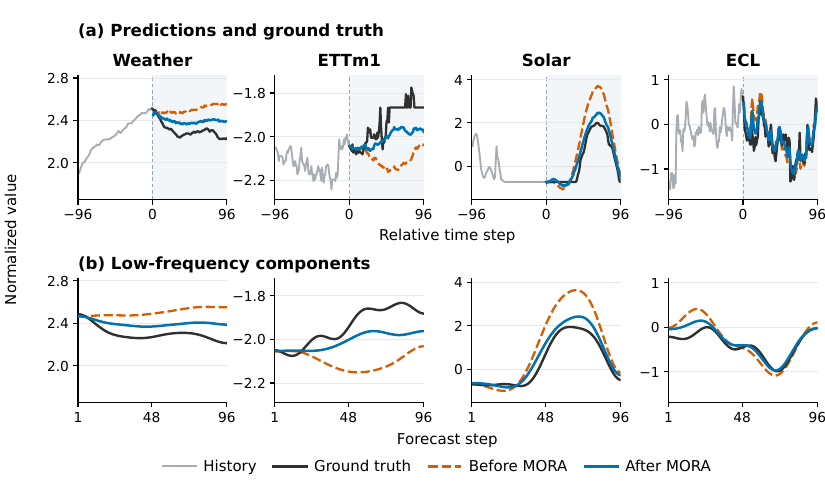}
    \caption{\morarevised{\textbf{TimeFilter predictions before and after applying MORA.}
    \textbf{(a)} The historical input is shown in gray.
    In the shaded forecast interval, we compare the ground truth
    with TimeFilter's predictions before and after applying MORA.
    \textbf{(b)} The corresponding low-frequency components,
    reconstructed from the first eight DCT components of each
    future sequence, show how adaptation changes the predicted
    level and slow variation.}}
    \label{fig:mora_timefilter_examples}
\end{figure}

\morarevised{We compare TimeFilter's predictions before and after
adaptation on four datasets in Figure~\ref{fig:mora_timefilter_examples}.
On Weather, the backbone overestimates the future level;
on Solar, it overestimates the peak magnitude.
MORA reduces these deviations, with the corresponding
low-frequency components following the ground truth more closely.
Additional comparisons across four forecasters are provided
in Appendix~\ref{app:prediction_examples}.}

\FloatBarrier

\section{Conclusion}
We presented MORA, a model-agnostic residual adapter for frozen
time-series forecasters. Motivated by the coarse-scale structure
of forecasting errors, MORA estimates segment-wise future means
and uses their discrepancies with the corresponding means of
backbone forecasts to construct stagewise corrections, scaled by
estimation reliability. Experiments on long- and short-term
forecasting benchmarks demonstrate broad improvements across
diverse architectures, including time-series foundation models
at different context lengths. These findings highlight coarse
future estimation as an effective complement to existing
forecasters.

\clearpage
\section*{AI Use Statement}
We used generative AI tools to assist with dataset preparation and
reformatting, manuscript organization, drafting, translation, language
editing, and \LaTeX{} typesetting. We carefully reviewed all AI-assisted
content and take full responsibility for the final manuscript.

\section*{Reproducibility Statement}
We describe MORA's coarse future estimation and stagewise residual
adaptation in the main paper. The estimator definitions, correction
subspace, and fitting and calibration procedures are detailed in
Appendix~\ref{app:method_details}. Dataset splits, preprocessing,
backbone configurations, hyperparameters, and computing resources are
documented in Appendix~\ref{app:experimental_details}. Additional
ablations and horizon-wise results are provided in
Appendices~\ref{app:mora_ablation} and~\ref{app:complete_results},
respectively. The anonymous code repository is linked in the abstract.

\bibliography{iclr2027_conference,experiments/references}
\bibliographystyle{iclr2027_conference}

\clearpage
\appendix
\section{Related Work}
\label{sec:related_work}
\label{app:related_work}

\paragraph{Time-Series Forecasting.}
Early neural forecasting models used recurrent and convolutional networks
to capture temporal patterns, as exemplified by LSTNet~\citep{lai2018lstnet}.
Subsequent Transformer-based approaches extended this line of research to
long-horizon prediction: Informer~\citep{zhou2021informer} improved attention
efficiency, while Autoformer~\citep{wu2021autoformer} incorporated progressive
trend--seasonal decomposition.
Spectral structure became another useful modeling tool, with FEDformer
combining decomposition and frequency-domain representations and FiLM using
Fourier filtering alongside Legendre memory~\citep{zhou2022film,zhou2022fedformer}.
Later work explored simpler predictors and alternative representations.
DLinear~\citep{zeng2023transformers} revisited decomposition-based linear
forecasting, and PatchTST~\citep{nie2023patchtst} combined temporal patching
with channel-independent modeling.
FreTS~\citep{yi2023frets} and FITS~\citep{xu2024fits} further developed spectral
forecasting through frequency-domain MLPs and complex-valued interpolation,
respectively, while iTransformer~\citep{liu2024itransformer} used variate
tokens to model cross-variate dependencies.
More recent approaches have refined multivariate dependency modeling
through patch-specific graph filtration in TimeFilter~\citep{hu2025timefilter},
local cross-variate interactions in VPNet~\citep{liu2026vpnet}, and
information-theoretic alignment in CGTFra~\citep{Yu2026CGTFra}.
Across these designs, temporal and cross-variate structure is incorporated
into the predictor through its architecture or representation.

\paragraph{Time-Series Foundation Models.}
Large-scale pretraining has broadened forecasting beyond dataset-specific
model development, allowing a single model to learn from heterogeneous
time-series collections and transfer to new tasks.
Chronos~\citep{ansari2024chronos} formulates forecasting as language modeling
over quantized time-series values, whereas TimesFM~\citep{das2024timesfm}
adopts a patched decoder-only architecture.
Moirai~\citep{woo2024moirai} further supports heterogeneous sampling
frequencies and variable dimensions within a unified forecasting model.
Subsequent developments have expanded both the modeling approach and the
range of supported tasks: Sundial~\citep{liu2025sundial} uses flow matching
for continuous-valued generative forecasting, while
Chronos-2~\citep{ansari2025chronos2} supports univariate, multivariate, and
covariate-informed forecasting through group attention.
These models provide reusable forecasting backbones that can be applied
without task-specific fine-tuning, creating a natural setting for studying
how their predictions can be refined without modifying pretrained parameters.

\paragraph{Forecast Adaptation and Post-Processing.}
Beyond backbone design and pretraining, another line of research improves
forecasts by addressing non-stationarity or adapting existing predictions.
RevIN~\citep{kim2022revin} removes and restores instance-level statistics,
while Non-stationary Transformers~\citep{liu2022nonstationary} combine series
stationarization with de-stationary attention to preserve informative
non-stationary structure.
These mechanisms are incorporated into the forecasting pipeline rather
than fitted as standalone residual corrections.
Test-time adaptation instead adjusts a deployed forecasting system as new
observations become available.
TAFAS~\citep{kim2025tafas} uses partially observed targets for online
calibration, and COSA~\citep{im2026cosa} updates a gated output-space residual
adapter using recently observed ground truth.
Closer to our setting, $\delta$-Adapter~\citep{liang2026the} learns lightweight
input perturbations and output residual corrections around a frozen forecaster.
MORA also keeps the backbone frozen, but separates coarse future
estimation from backbone-specific residual adaptation.
Its coarse estimators predict segment-wise means of the future from
the same historical window.
Backbone-specific residual mappings then translate discrepancies between
these estimates and the corresponding coarse statistics of the current
forecast into reliability-scaled, stagewise corrections.
Unlike TAFAS and COSA, MORA is fitted offline, and both the backbone
and adapter remain fixed throughout testing.

\clearpage

\section{History-Dependent Residual Correction}
\label{app:residual_correction_theory}

This appendix proves Proposition~1 from the introduction.
The result connects a history-dependent residual signal to a reduction
in expected MSE beyond the best constant correction along the same
coarse-scale direction. We first give the population least-squares
argument, then relate it to the forecast-mean probe and the residual
coordinates used by MORA.

\subsection{Improvement Beyond a Constant Correction}
\label{app:population_correction}

Following the problem setup, keep the pretrained forecaster
$f_{\theta}$ frozen and consider channel $c$, with backbone residual
$\mathbf{r}_c=\mathbf{y}_c-\widehat{\mathbf{y}}_c^{(0)}\in\mathbb{R}^{H}$.
Let $\mathbf{u}\in\mathbb{R}^{H}$ be a fixed unit vector along a
coarse-scale direction, and let $p=g(\mathbf{X})$ be a scalar signal
from the same historical window available to the backbone.
The function $g$ need not be linear; one example is the fixed linear
probe used in the introduction. All expectations below refer to the
joint distribution of $(\mathbf{X},\mathbf{Y})$, with $f_{\theta}$ and
$g$ held fixed. Define
\begin{equation}
    z=\mathbf{u}^{\top}\mathbf{r}_c,
    \qquad
    \mu_z=\E[z],
    \qquad
    \mu_p=\E[p],
    \qquad
    \sigma_p^2=\Var(p),
    \qquad
    \gamma=\Cov(z,p).
    \label{eq:prop_moments}
\end{equation}
Here $z$ is a scalar projection of the backbone residual; its relation
to the vector $\mathbf{z}_c=\mathbf{U}^{\top}\mathbf{r}_c$ is given in
Appendix~\ref{app:correction_estimation_scope}.
Assume $\E\|\mathbf{r}_c\|_2^2<\infty$ and $\E[p^2]<\infty$.
For coefficients $a,b\in\mathbb{R}$, the corrected forecast is
$\widehat{\mathbf{y}}_c^{(0)}+\mathbf{u}(a+bp)$, with expected MSE
\begin{equation}
    \mathcal{R}_c(a,b)
    =\frac{1}{H}\E\left[
        \left\|\mathbf{r}_c-\mathbf{u}(a+bp)\right\|_2^2
    \right].
    \label{eq:prop_risk}
\end{equation}
The factor $1/H$ averages the error over the forecast horizon.
A constant correction has $b=0$: its coefficient is the same for
every query, while its temporal shape is fixed by $\mathbf{u}$.

\newtheorem{moraproposition}{Proposition}
\begin{moraproposition}[History-dependent correction beyond a constant offset]
\label{prop:history_correction}
Under the assumptions above, suppose $\sigma_p^2>0$.
The best constant correction along $\mathbf{u}$ has coefficient
$\mu_z$. The optimal affine correction has
\begin{equation}
    b^{\star}=\frac{\gamma}{\sigma_p^2},
    \qquad
    a^{\star}=\mu_z-b^{\star}\mu_p,
    \qquad
    a^{\star}+b^{\star}p
    =\mu_z+\frac{\gamma}{\sigma_p^2}(p-\mu_p).
    \label{eq:prop_optimal_correction}
\end{equation}
Its improvement over the best constant correction is exactly
\begin{equation}
    \underbrace{\min_{a}\mathcal{R}_c(a,0)}_{\mathcal{R}_{\mathrm{const}}}
    -\underbrace{\min_{a,b}\mathcal{R}_c(a,b)}_{\mathcal{R}_{\mathrm{aff}}}
    =\frac{\gamma^2}{H\sigma_p^2}.
    \label{eq:prop_strict_gain}
\end{equation}
Consequently, $\Cov(z,p)\ne 0$ implies a strict reduction in expected
MSE beyond the best constant correction along that direction.
\end{moraproposition}

\noindent\textit{Proof.}
Since $\|\mathbf{u}\|_2=1$ and
$\mathbf{u}^{\top}(\mathbf{r}_c-\mathbf{u}z)=0$, the squared error
decomposes pointwise as
\begin{equation}
    \left\|\mathbf{r}_c-\mathbf{u}(a+bp)\right\|_2^2
    =\|\mathbf{r}_c-\mathbf{u}z\|_2^2+(z-a-bp)^2.
    \label{eq:prop_orthogonal_error}
\end{equation}
The first term is unaffected by either coefficient.
Centering $z$ and $p$ gives
\begin{equation}
    \E[(z-a-bp)^2]
    =\Var(z)-2b\gamma+b^2\sigma_p^2
      +(a+b\mu_p-\mu_z)^2.
    \label{eq:prop_affine_expansion}
\end{equation}
For any fixed $b$, the last term is minimized at
$a=\mu_z-b\mu_p$. Completing the square in the remaining terms yields
\begin{equation}
    \Var(z)-2b\gamma+b^2\sigma_p^2
    =\Var(z)-\frac{\gamma^2}{\sigma_p^2}
      +\sigma_p^2\left(b-\frac{\gamma}{\sigma_p^2}\right)^2.
    \label{eq:prop_complete_square}
\end{equation}
Because $\sigma_p^2>0$, the minimizing slope is
$b^{\star}=\gamma/\sigma_p^2$. Restricting instead to $b=0$ gives
the best constant coefficient $a=\mu_z$. Their minimum scalar squared
errors differ by $\gamma^2/\sigma_p^2$. The common orthogonal term
cancels, and division by $H$ proves Eq.~(\ref{eq:prop_strict_gain}).
\hfill\rule{0.5em}{0.5em}

If $\sigma_p^2=0$, the affine and constant correction classes coincide;
if $\sigma_p^2>0$ but $\gamma=0$, the chosen signal yields no additional
improvement over the best constant correction within the affine class.

\subsection{Connection to the Forecast-Mean Probe}
\label{app:mean_probe_connection}

For the forecast-mean probe, take
$\mathbf{u}=\mathbf{1}_H/\sqrt{H}$ and define the future-mean residual
$\bar r_c=H^{-1}\mathbf{1}_H^{\top}\mathbf{r}_c$, namely the true
future mean minus the predicted future mean.
Then $z=\sqrt{H}\bar r_c$, and Proposition~\ref{prop:history_correction}
gives
\begin{equation}
    \mathcal{R}_{\mathrm{const}}-\mathcal{R}_{\mathrm{aff}}
    =\frac{\Cov(\bar r_c,p)^2}{\Var(p)}.
    \label{eq:prop_mean_gain}
\end{equation}
The corresponding optimal correction is
\begin{equation}
    \mathbf{u}(a^{\star}+b^{\star}p)
    =\mathbf{1}_H\left[
        \E[\bar r_c]+\frac{\Cov(\bar r_c,p)}{\Var(p)}(p-\E[p])
    \right].
    \label{eq:prop_mean_correction}
\end{equation}
This history-dependent level correction is added to every forecast
step. It can therefore reduce the MSE of the full future sequence
without changing the deviations around the predicted future mean.

Because $p=g(\mathbf{X})$, the law of total expectation also gives
\begin{equation}
    \Cov(z,p)=\Cov\!\left(\E[z\mid\mathbf{X}],p\right).
    \label{eq:prop_conditional_mean}
\end{equation}
Nonzero covariance rules out a conditional mean
$\E[z\mid\mathbf{X}]$ that is constant almost surely.
Hence, the historical window contains predictive information about
the residual beyond a constant offset. The diagnostic probe differs
from MORA's learned coarse estimator, which predicts the future
quantity $\mathbf{q}_c=\mathbf{A}\mathbf{y}_c$ rather than a
backbone-specific residual.

\subsection{Connection to Stagewise Residual Adaptation}
\label{app:correction_estimation_scope}

The correction subspace in
Section~\ref{sec:ordered_residual_adaptation} makes the connection to
the residual coordinates explicit. For any fixed
$\mathbf{v}\in\mathbb{R}^{k}$ with $\|\mathbf{v}\|_2=1$,
orthonormality of $\mathbf{U}$ gives
\begin{equation}
    \mathbf{u}=\mathbf{U}\mathbf{v},
    \qquad
    \|\mathbf{u}\|_2=1,
    \qquad
    z=\mathbf{v}^{\top}\mathbf{z}_c.
    \label{eq:prop_subspace_coordinates}
\end{equation}
Thus Proposition~\ref{prop:history_correction} applies to a direction
within MORA's correction subspace whenever its covariance condition
holds. The forecast-mean direction illustrates the motivation,
whereas the tapered basis used by MORA leaves the first predicted
value unchanged. A mean-probe association alone does not establish
nonzero covariance for every direction in that subspace.

The remaining issue is estimation reliability.
For $\sigma_p^2>0$ and any fixed fitted coefficients
$(\widehat a,\widehat b)$ in the affine family above, the difference
from the best constant correction is
\begin{equation}
\begin{aligned}
    \mathcal{R}_c(\widehat a,\widehat b)
       -\mathcal{R}_{\mathrm{const}}
    =\frac{1}{H}\Bigl[
       &(\widehat a+\widehat b\mu_p-\mu_z)^2\\
       &+\sigma_p^2(\widehat b-b^{\star})^2
        -\frac{\gamma^2}{\sigma_p^2}\Bigr].
\end{aligned}
\label{eq:prop_estimation_penalty}
\end{equation}
The two nonnegative terms quantify coefficient-estimation error,
which can offset the population improvement.
This motivates accounting for estimation reliability, as in MORA's
coordinate-wise reliability scaling $\mathbf{G}^{e}$ and
$\mathbf{G}^{\ell}$. These gains are estimated from training data;
the final calibration $\mathbf{S}_c$ is fitted separately on validation
data, following Section~\ref{sec:adapter_training}.

Proposition~\ref{prop:history_correction} establishes the existence of
an improving correction under the evaluation distribution. It does
not guarantee that finite-sample estimates, or coefficients fitted
on a different distribution, achieve that improvement. The grouped
probe results provide empirical evidence relevant to the covariance
condition, rather than a proof of it. Window overlap leaves the risk
identity unchanged but must be considered when estimating uncertainty.

\clearpage

\section{Additional Method Details}
\label{app:method_details}

\subsection{Coarse Future Estimation}
\label{app:coarse_estimates}

\paragraph{Temporal extrapolation.}
The current-level estimate extends the latest observation over the
forecast horizon. Recent-pattern continuation instead repeats the
last $P_{\mathrm{eff}}=\min\{P,L\}$ observations, where $P$ is the
configured base period in sampling steps. For $h=1,\ldots,H$,
\begin{equation}
\begin{aligned}
\bigl[\mathcal C_1(\mathbf x_c,\boldsymbol\tau)\bigr]_h
    &=x_{L,c},\\
\bigl[\mathcal C_2(\mathbf x_c,\boldsymbol\tau)\bigr]_h
    &=x_{L-P_{\mathrm{eff}}+1+((h-1)\bmod P_{\mathrm{eff}}),c}.
\end{aligned}
\end{equation}
Both estimates use only the supplied historical window.
When $P>L$, recent-pattern continuation repeats the available
window rather than accessing observations outside it.

\paragraph{Training-fitted periodic template.}
When enabled, the periodic estimate uses a channel-wise harmonic
template fitted on the training data,
\begin{equation}
T_c(s)=a_{0,c}+\sum_{p\in\mathcal P}
\left[a_{p,c}\cos(2\pi s/p)+b_{p,c}\sin(2\pi s/p)\right].
\end{equation}
Here $s$ denotes the absolute sampling index. The period dictionary
$\mathcal P$ combines configured daily and weekly periods with
separated spectral peaks identified from detrended training series.
The harmonic coefficients are estimated by regularized least squares.
For a query ending at index $s_0$, we estimate a local level offset
$b_c^{\mathrm{loc}}=\sum_{\ell=1}^{L}w_\ell
[x_{\ell,c}-T_c(s_0-L+\ell)]$, where the normalized exponential
weights $w_\ell$ give greater weight to recent observations.
The extrapolated future is
$[\mathcal C_3(\mathbf x_c,\boldsymbol\tau)]_h
=T_c(s_0+h)+b_c^{\mathrm{loc}}$.
Each trajectory is mapped through $\mathbf A$; the resulting
estimates are retained separately in $\mathbf m_c^e$.

The dictionary contains at most eight periods, including $P$ and
$7P$, with the remaining periods selected from training-only spectral
peaks. Harmonic fitting uses ridge regularization of $10^{-3}$ under
an average squared-loss convention. The local level offset uses
normalized exponential weights with a half-life of 16 sampling steps.

\paragraph{Historical features.}
The compact feature vector contains 78 entries at $L=96$:
16 orthonormal DCT coefficients, eight whole-window statistics,
12 block means, 36 recent-dynamics features, and six time encodings.
The whole-window statistics comprise the mean, standard deviation,
first and last values, endpoint difference, linear slope, and the
mean and standard deviation of first differences.
Recent dynamics comprise 12 block means of the last 24 observations,
12 block means of their first differences, and the mean, standard
deviation, slope, and endpoint difference over the last 8, 16, and
24 observations. Time encodings are sine--cosine pairs evaluated
at the forecast origin for periods $P$, $P/2$, and $7P$.

The channel-wise estimator also uses extended feature representations.
At $L=96$, the full representation has 252 dimensions, additionally
including raw history, block standard deviations, recent increments,
and cross-channel summaries. Adding interactions between historical
state features and temporal encodings gives 348 dimensions.
These representations use the same historical window and coarse
future target, without changing the stagewise residual updates.

\paragraph{Channel-wise ridge estimation.}
We fit the absolute and last-centered views separately.
For view $v\in\{\mathrm{abs},\mathrm{rel}\}$, let
$\boldsymbol\Phi_c^v$ and $\mathbf Q_c^v$ collect the standardized
historical features and coarse targets over the fitting samples.
The channel-wise coefficient matrix is estimated as
\begin{equation}
\mathbf W_c^v=\arg\min_{\mathbf W}
\|\mathbf Q_c^v-\boldsymbol\Phi_c^v\mathbf W\|_F^2
+\lambda_v\|\mathbf W\|_F^2.
\end{equation}
Feature and target statistics are computed on the fitting samples,
with standard-deviation floors of $10^{-5}$ and $10^{-4}$, respectively.
Predictions are restored to the target scale, and the centered
prediction is additionally shifted by $x_{L,c}\mathbf 1_Q$.
The two estimates are then averaged as in
Equation~(\ref{eq:learned_coarse_estimate}).
The fitting and regularization-selection protocol is specified in
Appendix~\ref{app:fitting_protocol}.

\subsection{Factorized Coarse Estimation}
\label{app:factorized_estimator}

\paragraph{Periodic baseline.}
Let $\mathbf x_t\in\mathbb R^C$ denote the multivariate observation
at absolute sampling index $t$, and let $P_{\mathrm f}$ be the
configured period. For training indices $\mathcal T$, define
$\mathcal T_t=\{s\in\mathcal T:s\bmod P_{\mathrm f}
=t\bmod P_{\mathrm f}\}$.
The channel-wise periodic baseline is
\begin{equation}
\mathbf b(t)=\frac{1}{|\mathcal T_t|}
\sum_{s\in\mathcal T_t}\mathbf x_s.
\end{equation}
Unlike the harmonic template used for temporal extrapolation,
this baseline directly averages observations sharing the same phase.

\paragraph{Shared factor space.}
We subtract the periodic baseline and standardize the remaining
deviations using their training means $\boldsymbol\mu$ and
standard deviations $\boldsymbol\sigma$:
\begin{equation}
\boldsymbol\xi_t
=(\mathbf x_t-\mathbf b(t)-\boldsymbol\mu)
 \oslash\boldsymbol\sigma,
\qquad
\mathbf s_t=\mathbf V^\top\boldsymbol\xi_t.
\end{equation}
Here $\oslash$ denotes elementwise division, with standard deviations
floored at $10^{-5}$. The columns of
$\mathbf V\in\mathbb R^{C\times d_{\mathrm f}}$ are the leading
$d_{\mathrm f}$ eigenvectors of the training covariance of
$\boldsymbol\xi_t$, satisfying $\mathbf V^\top\mathbf V=\mathbf I$.
The resulting factor states summarize variation shared across channels.

\paragraph{Coarse factor prediction.}
For a query ending at $s_0$, the factor-space target for interval
$\mathcal B_j$ is
\begin{equation}
\mathbf q_j^{\mathrm f}
=\frac{1}{|\mathcal B_j|}
\sum_{h\in\mathcal B_j}\mathbf s_{s_0+h}.
\end{equation}
The predictor uses the latest factor state; means over the last
6, 12, 24, 48, and 96 steps; differences from the states 6, 12,
and 24 steps earlier; slopes over the last 24 steps; and the six
time encodings described in Appendix~\ref{app:coarse_estimates}.
At $L=96$, these features contain $10d_{\mathrm f}+6$ entries.
A single multi-output ridge regressor is fitted to the concatenated
absolute targets $\mathbf q_j^{\mathrm f}$ and last-relative targets
$\mathbf q_j^{\mathrm f}-\mathbf s_{s_0}$ for all factors and intervals.
The relative predictions are restored to absolute factor coordinates
by adding $\mathbf s_{s_0}$.
Training-internal rolling validation selects the ridge strength
and one fixed prediction rule: the absolute prediction, the restored
relative prediction, or their equal-weight average.
The selected rule remains fixed at inference.
Feature and target standardization follows the channel-wise ridge
estimator. Denote the selected prediction by
$\widehat{\mathbf q}_j^{\mathrm f}$.

\paragraph{Reconstruction in channel space.}
The factor predictions are mapped back to the original channels,
with the channel-wise mean, scale, and periodic baseline restored:
\begin{equation}
[\mathbf m_c^\ell]_j
=\frac{1}{|\mathcal B_j|}
 \sum_{h\in\mathcal B_j}b_c(s_0+h)
+\mu_c+\sigma_c\mathbf V_{c,:}\widehat{\mathbf q}_j^{\mathrm f}.
\end{equation}
We use $d_{\mathrm f}=32$ and $P_{\mathrm f}=7P$,
where $P=24$ for ECL/Traffic and $P=288$ for PEMS.
All train-dependent quantities follow the fitting protocol in
Appendix~\ref{app:fitting_protocol}.
The resulting $\mathbf m_c^\ell\in\mathbb R^Q$ is used in the
stagewise updates without changing their formulation.

\subsection{Correction Subspace}
\label{app:correction_space}

\paragraph{Cosine modes and boundary taper.}
For a horizon $H$ and correction rank $k<H$, let
$\mathbf B\in\mathbb R^{H\times k}$ contain the first $k$
orthonormal IDCT modes. For $h=1,\ldots,H$ and $j=0,\ldots,k-1$,
\begin{equation}
B_{h,j+1}
=a_j\cos\!\left(\frac{\pi j(h-\tfrac12)}{H}\right),
\qquad
a_0=H^{-1/2},\quad
a_j=\sqrt{2/H}\ \text{for }j\geq1.
\end{equation}
We apply a taper over the first $n_{\mathrm b}$ forecast steps,
where $2\leq n_{\mathrm b}\leq H$:
\begin{equation}
w_h^{\mathrm{tap}}=
\begin{cases}
\displaystyle
\sin^2\!\left(\frac{\pi}{2}
             \frac{h-1}{n_{\mathrm b}-1}\right),
&1\leq h\leq n_{\mathrm b},\\[1mm]
1,&n_{\mathrm b}<h\leq H.
\end{cases}
\end{equation}
The taper starts at zero and reaches one at $n_{\mathrm b}$.
The tapered modes and their reduced QR factorization are
\begin{equation}
\mathbf B_{\mathrm{tap}}
=\operatorname{diag}
 (w_1^{\mathrm{tap}},\ldots,w_H^{\mathrm{tap}})\mathbf B
=\mathbf U\mathbf R,
\end{equation}
where $\mathbf U\in\mathbb R^{H\times k}$ has orthonormal columns.
This orthonormalization preserves the span of the tapered modes.

\paragraph{Boundary preservation.}
The first row of $\mathbf B_{\mathrm{tap}}$ is zero.
Since $\mathbf R$ is nonsingular, the first row of $\mathbf U$
is also zero. Consequently, the final correction satisfies
\begin{equation}
\left[
\mathbf U\mathbf S_c
\left(\widehat{\mathbf z}_c^e+
      \widehat{\mathbf z}_c^\ell\right)
\right]_1=0,
\qquad
\widehat y_{1,c}=\widehat y_{1,c}^{(0)}.
\end{equation}
The constraint therefore preserves the backbone's first predicted
value; it does not force that value to equal the last observation.
The basis is computed once for each configuration and shared
across channels. Because tapering modifies the cosine modes,
its span is not identical to an untapered DCT truncation.

\paragraph{Dimensions.}
For long-term forecasting, we use $k=64$ and $n_{\mathrm b}=13$.
The fixed short-horizon settings for PEMS are listed in
Table~\ref{tab:adapter_short_horizons}.
These settings retain the same construction with $k<H$.

\subsection{Estimating the Residual Updates}
\label{app:update_estimation}

\paragraph{Stage-specific signals.}
For the extrapolation-based stage, we use
$(\mathbf d_c^e,\mathbf z_c)$ and
$\mathbf M^e=\mathbf A^e\mathbf U$.
For the learned stage, we use
$(\mathbf d_c^\ell,\mathbf z_c^\ell)$ and
$\mathbf M^\ell=\mathbf A\mathbf U$.
We describe the estimator for either stage below, omitting the
stage superscript. Let $p$ be the dimension of $\mathbf d_c$ and
define $\boldsymbol\eta_{i,c}=\mathbf d_{i,c}-\mathbf M\mathbf z_{i,c}$.

\paragraph{Residual mapping.}
From $N$ training pairs, we form the uncentered second-order moments
\begin{equation}
\mathbf C_{ab,c}=\frac{1}{N}\sum_{i=1}^{N}
\omega_{i,c}\mathbf a_{i,c}\mathbf b_{i,c}^{\top},
\qquad a,b\in\{z,d,\eta\}.
\end{equation}
No sample mean is subtracted. The weights are
$\omega_{i,c}=s_{i,c}^{-2\gamma}$, where
$s_{i,c}=\max\{\operatorname{RMS}(\mathbf d_{i,c}),\underline s_c\}$.
The floor $\underline s_c$ is the larger of $10^{-5}$ and the
training tenth percentile of the discrepancy RMS.
Setting $\gamma=0$ gives uniform weights.
These weights affect moment estimation; the deployed mapping
remains linear in $\mathbf d_c$.

We apply diagonal shrinkage
$\mathcal D_\beta(\mathbf C)=(1-\beta)\mathbf C+
\beta\operatorname{Diag}(\mathbf C)$, where
$\operatorname{Diag}(\mathbf C)$ retains only the diagonal entries.
Writing $\overline{\mathbf C}_{zz,c}=\mathcal D_\beta(\mathbf C_{zz,c})$
and $\overline{\mathbf C}_{\eta\eta,c}=\mathcal D_\beta(\mathbf C_{\eta\eta,c})$,
the regularized mapping is
\begin{equation}
\begin{aligned}
\boldsymbol\Sigma_c^d
&=(1-\rho)\left(
\mathbf M\overline{\mathbf C}_{zz,c}\mathbf M^\top
+\nu\overline{\mathbf C}_{\eta\eta,c}\right)
+\rho\mathbf C_{dd,c},\\
\mathbf K_c
&=\left[(1-\rho)\overline{\mathbf C}_{zz,c}\mathbf M^\top
+\rho\mathbf C_{zd,c}\right]
(\boldsymbol\Sigma_c^d+\epsilon_c\mathbf I_p)^{-1}.
\end{aligned}
\end{equation}
Here $\beta,\rho\in[0,1]$, $\nu>0$, and
$\epsilon_c=10^{-6}\max\{\operatorname{tr}(\boldsymbol\Sigma_c^d)/p,10^{-12}\}$.
The parameter $\rho$ interpolates between structured and empirical
second-order estimates. The mapping is computed by solving the
corresponding linear system, rather than explicitly forming its inverse.

\paragraph{Reliability scaling.}
Let $\widetilde z_{i,c,r}$ be the unscaled prediction for coordinate
$r$ from a mapping fitted without the temporal block containing
sample $i$. For a set of training samples $\mathcal I$, define
\begin{equation}
\alpha_r(\mathcal I)=
\frac{\sum_{i\in\mathcal I}\sum_c
\widetilde z_{i,c,r}z_{i,c,r}}
{\sum_{i\in\mathcal I}\sum_c\widetilde z_{i,c,r}^{\,2}}.
\end{equation}
The ratio is set to zero when its denominator does not exceed $10^{-12}$.
We use the pooled rule for $\mathbf G^e$ and the blockwise rule for
$\mathbf G^\ell$:
\begin{equation}
\begin{aligned}
g_r^{\mathrm{pool}}
&=\operatorname{clip}_{[0,g_{\max}]}\bigl(\alpha_r(\mathcal I)\bigr),\\
g_r^{\mathrm{block}}
&=\max\!\left\{0,\min\!\left[
g_{\max},\alpha_r(\mathcal I),\min_b\alpha_r(\mathcal I_b)
\right]\right\},
\end{aligned}
\end{equation}
where $\mathcal I_b$ is temporal block $b$ and $g_{\max}=1.5$.
Each stage uses its own targets and cross-fitted predictions.
Both rules pool channels, producing one gain per residual coordinate.
The gains form the diagonal matrices $\mathbf G^e$ and $\mathbf G^\ell$
and remain fixed at inference. Parameter selection and block construction
are described in Appendix~\ref{app:fitting_protocol}.

\subsection{Fitting and Calibration}
\label{app:fitting_protocol}

\paragraph{Training signals.}
We construct blocked out-of-fold (OOF) forecasts on the training split.
The latter half of the training-window origins is partitioned into
four contiguous blocks, and held queries are sampled at stride six.
For each block, task-specific backbone replicas exclude its raw-time
span and any training windows whose histories or targets overlap that
span. Train-dependent coarse estimators follow the same exclusion.
The replicas are used only to construct training signals; deployment
uses the original frozen backbone. For foundation models, pretrained
weights remain fixed throughout, while the local coarse estimators
are fitted using the corresponding training partitions.
Coarse-predictor ridge strengths are selected from
$\{1,10,100,1000,10000\}$ using training-internal rolling validation
with an $L+H-1$ separation between fitting and held window origins.

\paragraph{Stagewise fitting.}
The first-stage mapping is cross-fitted over the training blocks.
Its held-block predictions are used to estimate $\mathbf G^e$
according to Appendix~\ref{app:update_estimation}.
After applying this gain, the resulting first-stage coefficients
define the remaining target $\mathbf z_c^\ell$ and the adjusted
discrepancy $\mathbf d_c^\ell$ used to fit the second stage.
The second-stage mapping is cross-fitted over the same blocks,
and its held-block predictions determine $\mathbf G^\ell$.
Each mapping is then refitted on all of its corresponding training
pairs. The learned coarse predictor continues to target the future
coarse quantities, not the remaining backbone residual.
Train-dependent coarse estimators are fitted on the full training
split for validation and test inference.

\paragraph{Regularization settings.}
The residual mappings use parameter values drawn from
$\beta\in\{0.5,1\}$, $\nu\in\{0.5,1,2\}$,
$\rho\in\{0,0.25,0.5,1\}$, and $\gamma\in\{0,0.5\}$.
Candidates are ranked by their minimum relative reduction in
residual-coordinate squared error across training blocks, using the
blockwise gain rule. Mean block improvement breaks ties.
The factorized realization uses the fixed setting
$(\beta,\nu,\rho,\gamma)=(1,1,0,0)$ in both stages.

\paragraph{Validation calibration.}
Once both update mappings and their gains are fixed, we fit
$\mathbf S_c$ using validation forecasts. The $k$ coordinates are
partitioned into four consecutive groups, with one scale
$s_{c,g}\in[0,1.5]$ per channel and group.
We divide the validation queries into four contiguous blocks and
minimize the equally weighted normalized objective
\begin{equation}
\mathcal L_{\mathrm{cal}}
=\frac14\sum_{b=1}^{4}
\left[
0.1\frac{\mathrm{MSE}_b}{\mathrm{MSE}_b^{(0)}}
+0.9\frac{\mathrm{MAE}_{b,\epsilon}}{\mathrm{MAE}_b^{(0)}}
\right].
\end{equation}
Each numerator is evaluated from the calibrated forecasts and averaged
over queries, forecast steps, and channels within its block.
Superscript $(0)$ denotes the corresponding unadapted backbone error;
both denominators are floored at $10^{-12}$.
The smoothed absolute error replaces $|e|$ by
$\sqrt{e^2+\epsilon^2}$, with
$\epsilon=\max\{10^{-3}\frac14\sum_b\mathrm{MAE}_b^{(0)},10^{-7}\}$.
Starting from unit scales, we perform at most eight iteratively
reweighted least-squares updates, each solving a four-dimensional
box-constrained quadratic problem per channel.
Iteration stops when the largest scale change is below $10^{-5}$.

The resulting calibration is applied only to the sum of the two
updates; it does not alter the intermediate forecast used by the
learned stage. At inference, the backbone, coarse estimators,
residual mappings, reliability gains, and calibration remain fixed.

\clearpage

\section{Additional Experimental Details}
\label{app:experimental_details}

\subsection{Datasets and Forecasting Windows}
\label{app:experimental_data}

\paragraph{Datasets.}
Table~\ref{tab:experimental_datasets} summarizes the twelve benchmarks,
including their dimensions, sampling intervals, forecast horizons, and
split sizes. \textbf{ETT}~\citep{zhou2021informer} records oil temperature
and six load measurements from two electricity transformers, with
hourly and 15-minute versions. \textbf{Weather}~\citep{mpi_weather}
contains meteorological measurements from the Max Planck Institute
for Biogeochemistry in Jena. \textbf{ECL}, \textbf{Traffic}, and
\textbf{Solar} use the LSTNet releases~\citep{lai2018lstnet} and record
electricity consumption, road occupancy, and photovoltaic generation,
respectively. \textbf{PEMS03/04/07/08} use the STSGCN traffic-flow
releases~\citep{song2020stsgcn}; we forecast the first feature of each
sensor in the supplied arrays.
The data sources are the
\href{https://github.com/zhouhaoyi/ETDataset}{ETT repository},
\href{https://www.bgc-jena.mpg.de/wetter/}{MPI-BGC weather archive},
\href{https://github.com/laiguokun/multivariate-time-series-data}{LSTNet data repository},
and \href{https://github.com/Davidham3/STSGCN}{STSGCN repository}.

\paragraph{Chronological splits.}
ETT uses the first 20 benchmark months, divided into 12/4/4 months
for training/validation/test; each month contains 30 days in the
released loader. The remaining long-horizon datasets use 7:1:2
splits, and PEMS uses 6:2:2. Boundaries follow the corresponding
loaders. Table~\ref{tab:experimental_datasets} reports raw observations,
not overlapping forecasting windows.

\begin{table}[!htbp]
\centering
\caption{Forecasting benchmarks. $C$ is the number of variables and
$H$ is the forecast horizon in time steps. Split sizes count raw
observations before windowing.}
\label{tab:experimental_datasets}
\begingroup
\fontsize{9}{10.5}\selectfont
\setlength{\tabcolsep}{2pt}
\renewcommand{\arraystretch}{1.06}
\setlength{\aboverulesep}{1.5pt}
\setlength{\belowrulesep}{1.5pt}
\begin{tabular*}{\linewidth}{@{\extracolsep{\fill}}lrccll@{}}
\toprule[0.7pt]
\textbf{Dataset} & $C$ & \textbf{Horizons $H$}
 & \shortstack{\textbf{Split sizes}\\Train / Val / Test}
 & \textbf{Sampling} & \textbf{Domain}\\
\midrule[0.45pt]
ETTh1, ETTh2 & 7 & $\{96,192,336,720\}$ & (8640, 2880, 2880) & Hourly & Electricity\\
ETTm1, ETTm2 & 7 & $\{96,192,336,720\}$ & (34560, 11520, 11520) & 15 min & Electricity\\
Weather & 21 & $\{96,192,336,720\}$ & (36887, 5270, 10539) & 10 min & Weather\\
ECL & 321 & $\{96,192,336,720\}$ & (18412, 2632, 5260) & Hourly & Electricity\\
Traffic & 862 & $\{96,192,336,720\}$ & (12280, 1756, 3508) & Hourly & Transportation\\
Solar & 137 & $\{96,192,336,720\}$ & (36792, 5256, 10512) & 10 min & Energy\\
\midrule[0.5pt]
PEMS03 & 358 & $\{12,24,48,96\}$ & (15724, 5242, 5242) & 5 min & Transportation\\
PEMS04 & 307 & $\{12,24,48,96\}$ & (10195, 3398, 3399) & 5 min & Transportation\\
PEMS07 & 883 & $\{12,24,48,96\}$ & (16934, 5645, 5645) & 5 min & Transportation\\
PEMS08 & 170 & $\{12,24,48,96\}$ & (10713, 3571, 3572) & 5 min & Transportation\\
\bottomrule[0.7pt]
\end{tabular*}
\endgroup
\end{table}

\paragraph{Long-horizon preprocessing.}
Each channel is standardized using its training mean and population
standard deviation, with unit scale for constant channels. These
statistics are reused on validation and test data. Forecast origins
advance by one step; the future stays within its evaluation split,
while the initial history may extend into the preceding split.
Each baseline--MORA pair uses identical forecast origins, as do the
foundation-model comparisons between $L=96$ and $L=512$.
Incomplete evaluation batches are retained.

\paragraph{PEMS preprocessing.}
For a record of length $T$, split boundaries are
$\lfloor0.6T\rfloor$ and $\lfloor0.8T\rfloor$; windows stay within
their split without borrowing preceding-split history.
iTransformer and CGTFra use training-fitted standardization and test
stride one. TimeFilter uses split-specific standardization and test
stride 12, also adopted by the PatchTST, DLinear, and VPNet transfer
experiments. The latter convention uses test-split normalization
statistics, but does not fit MORA parameters on test labels.
Every local baseline/MORA pair shares preprocessing and window
enumeration, including the endpoint convention. Absolute normalized
errors are not directly comparable across the two preprocessing groups.

\subsection{Models and Inference}
\label{app:experimental_backbones}

\paragraph{Task-specific backbones.}
We use the released architectures and training objectives, with
model- and dataset-specific configurations at $L=96$.
Optimizers, batch sizes, and training budgets follow these configurations.
Checkpoints are selected on validation data before adaptation and then
frozen. Each local baseline--MORA pair uses the same checkpoint;
backbone hyperparameters are not retuned during adaptation.

\paragraph{PEMS baselines.}
TimeFilter, iTransformer, and CGTFra reproductions are checked against
the corresponding paper versions and horizon-specific references.
TimeFilter at $H=96$ is an extension without an original-paper score.
PatchTST, DLinear, and VPNet have no matching PEMS results in the
selected source papers; their transfer recipes retain the original
training principles while adapting to $L=96$, the PEMS channel count,
and TimeFilter's preprocessing. VPNet uses its Traffic recipe with
patch length 12 and stride six.

\morarevised{Across the 96 PEMS settings, Table~\ref{tab:complete_pems}
reports 90 baseline results from our local reproductions and uses
author-reported scores ($\dagger$) for the remaining six settings,
where our locally reproduced MSE exceeded the corresponding
author-reported value by more than 5\%: five from Table~9 of
iTransformer~\citep{liu2024itransformer} and one from
CGTFra~\citep{Yu2026CGTFra} for PEMS03 at $H=48$.
All \textsc{+MORA} forecasts are obtained by adapting our local
checkpoints; using the lower reference scores therefore yields
smaller displayed improvements than the paired local comparisons.
Table~\ref{tab:pems_quoted_paired} presents the author-reported
baselines, paired local baselines, and local \textsc{+MORA} results
side by side.}

\paragraph{Foundation models.}
Table~\ref{tab:foundation_checkpoints} lists the frozen checkpoints.
Chronos-2 uses its multivariate interface and released median forecast;
TimesFM~2.5 and Sundial use their univariate interfaces, returning the
released point forecast and the mean of 20 sampled trajectories,
respectively. These choices are fixed within each baseline/MORA pair.

\begin{table}[!ht]
\centering
\begin{threeparttable}
\caption{Pretrained checkpoints used at both context lengths.
All foundation-model weights remain frozen.}
\label{tab:foundation_checkpoints}
\fontsize{9}{10.5}\selectfont
\setlength{\tabcolsep}{4pt}
\renewcommand{\arraystretch}{1.05}
\begin{tabular*}{\linewidth}{@{\extracolsep{\fill}}ll@{}}
\toprule
\textbf{Model} & \textbf{Checkpoint identifier}\\
\midrule
Chronos-2 & \path{amazon/chronos-2}\\
Sundial & \path{thuml/sundial-base-128m}\\
TimesFM 2.5 & \path{google/timesfm-2.5-200m-pytorch}\\
\bottomrule
\end{tabular*}
\end{threeparttable}
\end{table}

\subsection{Adapter Configuration and Fitting}
\label{app:experimental_fitting}

Tables~\ref{tab:adapter_hyperparameters}
and~\ref{tab:adapter_short_horizons} collect the shared configuration
and fixed short-horizon dimensions. The correction basis, residual
estimators, and fitting procedure are specified in
Appendices~\ref{app:correction_space}, \ref{app:update_estimation},
and~\ref{app:fitting_protocol}. The feature and periodic-template
configurations are kept fixed within each dataset--backbone group
across horizons, without horizon-wise test selection or per-query fallback.

\begin{table}[!ht]
\centering
\caption{Shared MORA configuration. PEMS-specific dimensions are
given in Table~\ref{tab:adapter_short_horizons}.}
\label{tab:adapter_hyperparameters}
\fontsize{9}{10.5}\selectfont
\setlength{\tabcolsep}{4pt}
\renewcommand{\arraystretch}{1.02}
\begin{tabular*}{\linewidth}{@{\extracolsep{\fill}}
>{\raggedright\arraybackslash}p{0.40\linewidth}
>{\raggedright\arraybackslash}p{0.57\linewidth}@{}}
\toprule
\textbf{Setting} & \textbf{Value or selection rule}\\
\midrule
Long-horizon correction rank & $k=64$\\
Coarse intervals & $Q=\min\{64,\lfloor H/4\rfloor\}$\\
Coarse-predictor ridge candidates & $\{1,10,100,1000,10000\}$\\
\addlinespace[2pt]
Channel-factor dimension & 32, for the factorized estimator\\
Factor-predictor training stride & One time step\\
\addlinespace[2pt]
Residual-fitting OOF blocks & Four; held origins in the later half of training\\
Residual-fitting origin stride & Six time steps\\
\addlinespace[2pt]
Final calibration & Four coefficient groups per channel\\
Calibration scale range & $[0,1.5]$; fitted on validation data\\
Test-time parameter updates & None\\
\bottomrule
\end{tabular*}
\end{table}

\begin{table}[!ht]
\centering
\caption{PEMS dimensions, fixed across datasets and backbones.
Boundary taper is measured in forecast steps.}
\label{tab:adapter_short_horizons}
\fontsize{9}{10.5}\selectfont
\renewcommand{\arraystretch}{1.02}
\begin{tabular*}{\linewidth}{@{\extracolsep{\fill}}lrrrr@{}}
\toprule
\textbf{Forecast horizon $H$} & 12 & 24 & 48 & 96\\
\midrule
Coarse intervals $Q$ & 3 & 6 & 12 & 24\\
Correction rank $k$ & 8 & 16 & 32 & 64\\
Boundary taper & 2 & 3 & 6 & 13\\
\bottomrule
\end{tabular*}
\end{table}

\paragraph{Dataset-specific fitting.}
ECL, Traffic, and PEMS use the factorized estimator described in
Appendix~\ref{app:factorized_estimator}; other datasets use the
channel-wise estimator. This dataset-level choice was fixed during
development on the inspected benchmarks. For PEMS OOF replicas,
checkpoint selection uses a 6\% training-internal validation interval;
the official validation and test splits are not used for replica fitting
or checkpoint selection.

Repeated runs reuse the residual mappings fitted for the initial run.
For Sundial, runs with the channel-wise estimator refit validation
calibration; runs with the factorized estimator reuse the complete
adapter. These repetitions vary inference or adaptation, not the
pretrained foundation-model weights.

\paragraph{Computing environment.}
Independent experiment jobs run across eight NVIDIA GeForce RTX 4090
GPUs with 24\,GB each. Model-specific environments follow the released
implementations: VPNet uses Python 3.8.20, PyTorch 1.12.1, and CUDA 11.3;
CGTFra uses Python 3.9.24, PyTorch 2.4.1, and CUDA 12.1.
The foundation-model environments use PyTorch 2.7.1.
Chronos-2 and TimesFM~2.5 use Transformers 4.57.6, and Sundial uses
Transformers 4.40.1. The corresponding packages are
\texttt{chronos-forecasting} 2.3.2 and \texttt{timesfm} 2.0.2.
\morarevised{Inference efficiency across datasets and backbones is reported
in Appendix~\ref{app:computational_efficiency}.}

\subsection{Repeated Runs, Metrics, and Aggregation}
\label{app:experimental_reporting}

The long-horizon task-specific experiments use one run per horizon
on ETTh1, ETTm1, ETTm2, and Weather, and three backbone seeds on
ETTh2, ECL, Traffic, and Solar.
DLinear, PatchTST, TimeFilter, and VPNet use seed~2021 for the
single-run datasets and seeds~2021--2023 for repeated datasets.
CGTFra uses seed~2023 and seeds~2023--2025, respectively.
iTransformer uses seed~2021 for single-run settings except
ETTm2 at $H=96$, which uses seed~2023; repeated datasets use
seeds~2023--2025.
PEMS retains one local seed per backbone: 2021 for
TimeFilter/PatchTST/DLinear/VPNet and 2023 for iTransformer/CGTFra.

Chronos-2 and TimesFM~2.5 each use one fixed pretrained checkpoint.
Sundial uses sampling seed~2023 on ETTh1, ETTm1, ETTm2, and
Weather, and seeds~2023--2025 on ETTh2, ECL, Traffic, and Solar.
\morarevised{We report MSE and MAE averaged across runs for each setting.}

For $N$ queries, MSE and MAE average over queries, forecast steps,
and variables:
\begin{equation}
\begin{aligned}
 \mathrm{MSE}&=\frac{1}{NHC}
    \sum_{i,t,c}(y_{i,t,c}-\widehat y_{i,t,c})^2,\\
 \mathrm{MAE}&=\frac{1}{NHC}
    \sum_{i,t,c}|y_{i,t,c}-\widehat y_{i,t,c}|.
\end{aligned}
\label{eq:experimental_metrics}
\end{equation}
Local baseline--MORA pairs use identical queries, targets, and
normalization. Let $E^{(0)}_{h,s}$ and $E_{h,s}$ denote the baseline
and adapted errors at horizon $h$ for run $s$.
We first average the $S_h$ runs within a horizon, then give each
horizon equal weight:
\begin{equation}
 \overline E^{(0)}
 =\frac{1}{|\mathcal H|}\sum_{h\in\mathcal H}
      \frac{1}{S_h}\sum_{s=1}^{S_h}E^{(0)}_{h,s},
 \qquad
 \overline E
 =\frac{1}{|\mathcal H|}\sum_{h\in\mathcal H}
      \frac{1}{S_h}\sum_{s=1}^{S_h}E_{h,s}.
 \label{eq:table_error_aggregation}
\end{equation}
For either metric, the reported improvement is
\begin{equation}
 \mathrm{Imp.}(E)=100\,
 \frac{\overline E^{(0)}-\overline E}
      {\overline E^{(0)}}.
 \label{eq:main_table_improvement}
\end{equation}
Imp.\ is computed from unrounded mean errors, not by averaging
seed- or horizon-specific percentages. Dataset-level Imp.\ values
are then averaged equally for model-level summaries, so additional
runs do not increase a dataset's weight.

Long-horizon and foundation-model tables use local paired baselines.
For PEMS, Imp.\ uses the displayed reference score for the six quoted
entries and the local baseline otherwise. Thus, the literature-reference
comparison is distinct from the local paired comparison.
\morarevised{Win counts are computed from unrounded errors for each
dataset--backbone--horizon setting.}

\subsection{Variation Across Datasets and Metrics}
\label{app:result_variation}

\paragraph{Dataset-dependent gains.}
The benefit of residual adaptation depends on the dataset--backbone pair.
For TimeFilter, the horizon-averaged MSE reduction is 12.910\% on
Solar, 1.931\% on Traffic, and 0.508\% on ETTh2. CGTFra on ETTh1
and VPNet on ETTh2 increase both errors, while VPNet on Solar improves
MSE but not MAE. The full results retain these cases, so average gains
should not be read as improvements in every setting.

\paragraph{Metric-dependent gains.}
At $L=512$, all 24 foundation-model--dataset pairs improve in MSE,
but only 19 improve in MAE. For example, TimesFM~2.5 on Solar reduces
MSE by 21.222\% while increasing MAE by 4.503\%. Squared loss gives
larger errors greater weight than absolute loss. The two metrics
therefore capture different aspects of a correction, and a reduction
in MSE need not produce a reduction in MAE.

\paragraph{Reference-dependent gains.}
\morarevised{Using paired local baselines for all 96 PEMS settings
increases the average MSE reduction from 28.10\% to 28.88\% and
the average MAE reduction from 18.47\% to 19.09\%
(Table~\ref{tab:pems_protocol_summary}). The number of settings
with lower MSE also increases from 90 to 92 out of 96.
Both comparisons use identical \textsc{+MORA} forecasts and the
same aggregation rule; the only change is replacing the six
author-reported baseline scores with their corresponding local results.}

\FloatBarrier
\clearpage
\section{Additional Ablations}
\label{app:mora_ablation}

\begingroup
\setlength{\intextsep}{7pt}
\setlength{\abovecaptionskip}{3pt}
\setlength{\belowcaptionskip}{2pt}

Tables~\ref{tab:mora_ablation_components} and~\ref{tab:mora_ablation_posthoc}
use TimeFilter (seed~2021) with $L=96$, at $H=720$ and $H=96$,
respectively. \morarevised{Table~\ref{tab:mora_ablation_components}
presents three illustrative datasets from the component ablation study,
with ECL also using the channel-wise estimator in this comparison.}
Methods share the same frozen checkpoint and evaluation
windows within each setting.

\paragraph{Individual extrapolation sources.}
We remove each extrapolation source in turn, retain the learned estimate,
and refit the residual adapter (Table~\ref{tab:abl_observers}).
The study covers eight datasets, TimeFilter, iTransformer, PatchTST,
and DLinear, and $H\in\{96,720\}$ at $L=96$, using the channel-wise
estimator. For either error metric $E$, entries average the paired
change $100(E_{\mathrm{removed}}/E_{\mathrm{full}}-1)$ over matched settings;
positive values indicate higher error after removal.
Removing recent-pattern continuation or the static periodic template
increases both errors on average over their matched settings.

\begin{table}[!htbp]
\centering
\caption{Contribution of individual extrapolation sources. Entries
are mean relative error changes (\%) after removal and refitting.
The last two rows use the same 40 template-enabled settings.}
\label{tab:abl_observers}
\begingroup
\fontsize{9}{10.5}\selectfont
\setlength{\tabcolsep}{3pt}
\renewcommand{\arraystretch}{1.02}
\setlength{\aboverulesep}{1.5pt}
\setlength{\belowrulesep}{1.5pt}
\begin{tabular*}{\linewidth}{@{\extracolsep{\fill}}lrrr@{}}
\toprule[0.7pt]
\textbf{Estimate removed} & \textbf{Settings}
 & \textbf{MSE change} & \textbf{MAE change} \\
\midrule[0.45pt]
Current level & 64 & +0.038 & -0.173 \\
Recent-pattern continuation & 40 & +0.906 & +0.796 \\
Static periodic template & 40 & +1.603 & +0.380 \\
\bottomrule[0.7pt]
\end{tabular*}
\endgroup
\end{table}

\paragraph{Additional comparisons.}
Table~\ref{tab:abl_posthoc_expansion} extends the input--output
$\delta$-Adapter comparison to five further settings at $L=H=96$.
MORA achieves lower MSE in all five and lower MAE in three,
including both metrics on ECL/TimeFilter.
These results complement the three-dataset comparison in the main text.

\begin{table}[!htbp]
\centering
\caption{Additional comparisons with $\delta$-Adapter (input + output)
at $L=H=96$. MORA uses the factorized estimator on ECL and Traffic
and the channel-wise estimator elsewhere.}
\label{tab:abl_posthoc_expansion}
\begingroup
\fontsize{9}{10.5}\selectfont
\setlength{\tabcolsep}{2pt}
\renewcommand{\arraystretch}{1.02}
\setlength{\aboverulesep}{1.5pt}
\setlength{\belowrulesep}{1.5pt}
\begin{tabular*}{\linewidth}{@{\extracolsep{\fill}}llrrrrrr@{}}
\toprule[0.7pt]
\textbf{Dataset} & \textbf{Backbone}
 & \multicolumn{3}{c}{\textbf{MSE}}
 & \multicolumn{3}{c}{\textbf{MAE}} \\
\cmidrule(lr){3-5}\cmidrule(lr){6-8}
 & & Base & $\delta$ (I+O) & MORA & Base & $\delta$ (I+O) & MORA \\
\midrule[0.45pt]
ECL & PatchTST & 0.180 & 0.161 & \textbf{0.150} & 0.273 & \textbf{0.252} & 0.253 \\
Traffic & PatchTST & 0.459 & 0.432 & \textbf{0.431} & 0.298 & \textbf{0.278} & 0.285 \\
ETTm2 & PatchTST & 0.183 & 0.182 & \textbf{0.171} & 0.268 & 0.269 & \textbf{0.258} \\
ETTh2 & PatchTST & 0.299 & 0.298 & \textbf{0.285} & 0.349 & 0.348 & \textbf{0.337} \\
\addlinespace[2pt]
ECL & TimeFilter & 0.133 & 0.132 & \textbf{0.129} & 0.230 & 0.230 & \textbf{0.228} \\
\bottomrule[0.7pt]
\end{tabular*}
\endgroup
\end{table}

\paragraph{Comparison details.}
Affine calibration fits a channel-wise scale and offset on validation
data. Direct residual regression uses the original OOF training pairs
and reduced-rank ridge, matching MORA's predictive coefficient count
to within approximately 2\%.
The $\delta$-Adapter comparisons use the authors' released
implementations~\citep{liang2026the} with the backbone frozen and
online test-time updates disabled. Neural adapters use seed 2025;
input--output candidates (additive budgets 0.01 and 0.1, or a
multiplicative budget of 0.2) are selected by validation MSE.

\FloatBarrier
\endgroup

\clearpage
\section{Complete Forecasting Results}
\label{app:complete_results}
Tables~\ref{tab:complete_task}--\ref{tab:complete_pems} report the
horizon-level results for the main comparisons, with runs aggregated
as described in Appendix~\ref{app:experimental_reporting}.
Bold marks the lower displayed error within each baseline--MORA pair;
ties at the displayed precision are unbolded. Shaded rows give Imp.\ (\%).

\begingroup
\setlength{\intextsep}{6pt}
\setlength{\abovecaptionskip}{3pt}
\setlength{\belowcaptionskip}{2pt}
\newcommand{\AppScore}[2]{\ifnum#2=1\textbf{#1}\else#1\fi}
\newcommand{\AppSD}[1]{\,{\fontsize{6}{6.8}\selectfont$\pm$#1}}
\newcommand{\AppImp}[1]{%
  \ifdim#1pt<0pt\textcolor{MoraNegative}{#1}\else#1\fi}

\subsection{Task-Specific Forecasters}
\label{app:task_results}
\begin{table}[H]
\centering
\caption{Complete long-horizon results for six task-specific backbones at $L=96$.}
\label{tab:complete_task}
\label{tab:full_task:etth1}
\label{tab:full_task:etth2}
\label{tab:full_task:ettm1}
\label{tab:full_task:ettm2}
\label{tab:full_task:weather}
\label{tab:full_task:ecl}
\label{tab:full_task:traffic}
\label{tab:full_task:solar}
\fontsize{8}{8.6}\selectfont
\setlength{\tabcolsep}{1pt}
\renewcommand{\arraystretch}{1.00}
\setlength{\aboverulesep}{0.6pt}
\setlength{\belowrulesep}{0.6pt}
\begin{tabular}{@{}>{\raggedright\arraybackslash}p{30pt}>{\raggedright\arraybackslash}p{14pt}>{\raggedright\arraybackslash}p{34pt}>{\raggedleft\arraybackslash}p{\dimexpr(\linewidth-78pt-28\tabcolsep)/12\relax}>{\raggedleft\arraybackslash}p{\dimexpr(\linewidth-78pt-28\tabcolsep)/12\relax}>{\raggedleft\arraybackslash}p{\dimexpr(\linewidth-78pt-28\tabcolsep)/12\relax}>{\raggedleft\arraybackslash}p{\dimexpr(\linewidth-78pt-28\tabcolsep)/12\relax}>{\raggedleft\arraybackslash}p{\dimexpr(\linewidth-78pt-28\tabcolsep)/12\relax}>{\raggedleft\arraybackslash}p{\dimexpr(\linewidth-78pt-28\tabcolsep)/12\relax}>{\raggedleft\arraybackslash}p{\dimexpr(\linewidth-78pt-28\tabcolsep)/12\relax}>{\raggedleft\arraybackslash}p{\dimexpr(\linewidth-78pt-28\tabcolsep)/12\relax}>{\raggedleft\arraybackslash}p{\dimexpr(\linewidth-78pt-28\tabcolsep)/12\relax}>{\raggedleft\arraybackslash}p{\dimexpr(\linewidth-78pt-28\tabcolsep)/12\relax}>{\raggedleft\arraybackslash}p{\dimexpr(\linewidth-78pt-28\tabcolsep)/12\relax}>{\raggedleft\arraybackslash}p{\dimexpr(\linewidth-78pt-28\tabcolsep)/12\relax}@{}}
\toprule[0.8pt]
\textbf{Dataset} & $H$ & Forecast & \multicolumn{2}{c}{\textbf{TimeFilter}} & \multicolumn{2}{c}{\textbf{iTransformer}} & \multicolumn{2}{c}{\textbf{CGTFra}} & \multicolumn{2}{c}{\textbf{PatchTST}} & \multicolumn{2}{c}{\textbf{DLinear}} & \multicolumn{2}{c}{\textbf{VPNet}} \\
\cmidrule(lr){4-5}\cmidrule(lr){6-7}\cmidrule(lr){8-9}\cmidrule(lr){10-11}\cmidrule(lr){12-13}\cmidrule(lr){14-15}
 & & & MSE & MAE & MSE & MAE & MSE & MAE & MSE & MAE & MSE & MAE & MSE & MAE \\
\midrule
\multirow{12}{*}{\rotatebox[origin=c]{90}{\textbf{ETTh1}}} & \multirow{3}{*}{96} & Base & \AppScore{0.369}{0} & \AppScore{0.394}{0} & \AppScore{0.386}{0} & \AppScore{0.404}{0} & \AppScore{0.373}{0} & \AppScore{0.388}{1} & \AppScore{0.379}{0} & \AppScore{0.399}{0} & \AppScore{0.396}{0} & \AppScore{0.411}{0} & \AppScore{0.375}{0} & \AppScore{0.387}{0} \\
 &  & \textbf{+MORA} & \AppScore{0.358}{1} & \AppScore{0.388}{1} & \AppScore{0.382}{1} & \AppScore{0.402}{1} & \AppScore{0.371}{1} & \AppScore{0.389}{0} & \AppScore{0.365}{1} & \AppScore{0.392}{1} & \AppScore{0.378}{1} & \AppScore{0.401}{1} & \AppScore{0.368}{1} & \AppScore{0.385}{1} \\
 &  & \cellcolor{MoraTint}Imp.\ (\%) & \cellcolor{MoraTint}\AppImp{2.890} & \cellcolor{MoraTint}\AppImp{1.499} & \cellcolor{MoraTint}\AppImp{1.046} & \cellcolor{MoraTint}\AppImp{0.529} & \cellcolor{MoraTint}\AppImp{0.442} & \cellcolor{MoraTint}\AppImp{-0.234} & \cellcolor{MoraTint}\AppImp{3.652} & \cellcolor{MoraTint}\AppImp{1.700} & \cellcolor{MoraTint}\AppImp{4.486} & \cellcolor{MoraTint}\AppImp{2.376} & \cellcolor{MoraTint}\AppImp{1.945} & \cellcolor{MoraTint}\AppImp{0.647} \\
\cmidrule[0.25pt]{2-15}
 & \multirow{3}{*}{192} & Base & \AppScore{0.412}{0} & \AppScore{0.421}{0} & \AppScore{0.442}{0} & \AppScore{0.436}{0} & \AppScore{0.429}{0} & \AppScore{0.422}{0} & \AppScore{0.425}{0} & \AppScore{0.431}{0} & \AppScore{0.447}{0} & \AppScore{0.442}{0} & \AppScore{0.428}{0} & \AppScore{0.419}{0} \\
 &  & \textbf{+MORA} & \AppScore{0.403}{1} & \AppScore{0.415}{1} & \AppScore{0.431}{1} & \AppScore{0.428}{1} & \AppScore{0.424}{1} & \AppScore{0.418}{1} & \AppScore{0.411}{1} & \AppScore{0.422}{1} & \AppScore{0.424}{1} & \AppScore{0.430}{1} & \AppScore{0.420}{1} & \AppScore{0.415}{1} \\
 &  & \cellcolor{MoraTint}Imp.\ (\%) & \cellcolor{MoraTint}\AppImp{2.304} & \cellcolor{MoraTint}\AppImp{1.442} & \cellcolor{MoraTint}\AppImp{2.565} & \cellcolor{MoraTint}\AppImp{1.966} & \cellcolor{MoraTint}\AppImp{1.012} & \cellcolor{MoraTint}\AppImp{0.815} & \cellcolor{MoraTint}\AppImp{3.418} & \cellcolor{MoraTint}\AppImp{2.178} & \cellcolor{MoraTint}\AppImp{5.117} & \cellcolor{MoraTint}\AppImp{2.840} & \cellcolor{MoraTint}\AppImp{1.843} & \cellcolor{MoraTint}\AppImp{0.940} \\
\cmidrule[0.25pt]{2-15}
 & \multirow{3}{*}{336} & Base & \AppScore{0.447}{0} & \AppScore{0.439}{0} & \AppScore{0.491}{0} & \AppScore{0.461}{0} & \AppScore{0.470}{0} & \AppScore{0.442}{0} & \AppScore{0.471}{0} & \AppScore{0.458}{0} & \AppScore{0.491}{0} & \AppScore{0.470}{0} & \AppScore{0.466}{0} & \AppScore{0.441}{0} \\
 &  & \textbf{+MORA} & \AppScore{0.438}{1} & \AppScore{0.435}{1} & \AppScore{0.484}{1} & \AppScore{0.455}{1} & \AppScore{0.470}{0} & \AppScore{0.442}{0} & \AppScore{0.452}{1} & \AppScore{0.448}{1} & \AppScore{0.459}{1} & \AppScore{0.452}{1} & \AppScore{0.462}{1} & \AppScore{0.438}{1} \\
 &  & \cellcolor{MoraTint}Imp.\ (\%) & \cellcolor{MoraTint}\AppImp{2.022} & \cellcolor{MoraTint}\AppImp{0.791} & \cellcolor{MoraTint}\AppImp{1.462} & \cellcolor{MoraTint}\AppImp{1.178} & \cellcolor{MoraTint}\AppImp{0.098} & \cellcolor{MoraTint}\AppImp{0.078} & \cellcolor{MoraTint}\AppImp{4.167} & \cellcolor{MoraTint}\AppImp{2.282} & \cellcolor{MoraTint}\AppImp{6.483} & \cellcolor{MoraTint}\AppImp{3.860} & \cellcolor{MoraTint}\AppImp{0.960} & \cellcolor{MoraTint}\AppImp{0.535} \\
\cmidrule[0.25pt]{2-15}
 & \multirow{3}{*}{720} & Base & \AppScore{0.445}{0} & \AppScore{0.456}{0} & \AppScore{0.499}{0} & \AppScore{0.487}{0} & \AppScore{0.470}{1} & \AppScore{0.463}{1} & \AppScore{0.518}{0} & \AppScore{0.505}{0} & \AppScore{0.515}{0} & \AppScore{0.513}{0} & \AppScore{0.476}{1} & \AppScore{0.468}{1} \\
 &  & \textbf{+MORA} & \AppScore{0.441}{1} & \AppScore{0.454}{1} & \AppScore{0.494}{1} & \AppScore{0.482}{1} & \AppScore{0.485}{0} & \AppScore{0.471}{0} & \AppScore{0.482}{1} & \AppScore{0.482}{1} & \AppScore{0.494}{1} & \AppScore{0.499}{1} & \AppScore{0.483}{0} & \AppScore{0.471}{0} \\
 &  & \cellcolor{MoraTint}Imp.\ (\%) & \cellcolor{MoraTint}\AppImp{0.900} & \cellcolor{MoraTint}\AppImp{0.280} & \cellcolor{MoraTint}\AppImp{0.984} & \cellcolor{MoraTint}\AppImp{1.078} & \cellcolor{MoraTint}\AppImp{-3.398} & \cellcolor{MoraTint}\AppImp{-1.857} & \cellcolor{MoraTint}\AppImp{6.912} & \cellcolor{MoraTint}\AppImp{4.532} & \cellcolor{MoraTint}\AppImp{4.052} & \cellcolor{MoraTint}\AppImp{2.584} & \cellcolor{MoraTint}\AppImp{-1.427} & \cellcolor{MoraTint}\AppImp{-0.731} \\
\midrule[0.45pt]
\multirow{12}{*}{\rotatebox[origin=c]{90}{\textbf{ETTh2}}} & \multirow{3}{*}{96} & Base & \AppScore{0.288}{0} & \AppScore{0.340}{0} & \AppScore{0.301}{0} & \AppScore{0.350}{0} & \AppScore{0.296}{0} & \AppScore{0.339}{0} & \AppScore{0.302}{0} & \AppScore{0.353}{0} & \AppScore{0.349}{0} & \AppScore{0.401}{0} & \AppScore{0.290}{0} & \AppScore{0.335}{0} \\
 &  & \textbf{+MORA} & \AppScore{0.283}{1} & \AppScore{0.336}{1} & \AppScore{0.292}{1} & \AppScore{0.344}{1} & \AppScore{0.288}{1} & \AppScore{0.335}{1} & \AppScore{0.285}{1} & \AppScore{0.338}{1} & \AppScore{0.300}{1} & \AppScore{0.355}{1} & \AppScore{0.288}{1} & \AppScore{0.334}{1} \\
 &  & \cellcolor{MoraTint}Imp.\ (\%) & \cellcolor{MoraTint}\AppImp{1.768} & \cellcolor{MoraTint}\AppImp{1.024} & \cellcolor{MoraTint}\AppImp{3.091} & \cellcolor{MoraTint}\AppImp{1.843} & \cellcolor{MoraTint}\AppImp{2.538} & \cellcolor{MoraTint}\AppImp{1.269} & \cellcolor{MoraTint}\AppImp{5.748} & \cellcolor{MoraTint}\AppImp{4.233} & \cellcolor{MoraTint}\AppImp{13.991} & \cellcolor{MoraTint}\AppImp{11.526} & \cellcolor{MoraTint}\AppImp{0.716} & \cellcolor{MoraTint}\AppImp{0.187} \\
\cmidrule[0.25pt]{2-15}
 & \multirow{3}{*}{192} & Base & \AppScore{0.381}{0} & \AppScore{0.406}{0} & \AppScore{0.380}{0} & \AppScore{0.399}{0} & \AppScore{0.368}{0} & \AppScore{0.386}{0} & \AppScore{0.383}{0} & \AppScore{0.403}{0} & \AppScore{0.475}{0} & \AppScore{0.475}{0} & \AppScore{0.358}{0} & \AppScore{0.377}{0} \\
 &  & \textbf{+MORA} & \AppScore{0.369}{1} & \AppScore{0.393}{1} & \AppScore{0.369}{1} & \AppScore{0.391}{1} & \AppScore{0.362}{1} & \AppScore{0.381}{1} & \AppScore{0.364}{1} & \AppScore{0.388}{1} & \AppScore{0.404}{1} & \AppScore{0.421}{1} & \AppScore{0.356}{1} & \AppScore{0.376}{1} \\
 &  & \cellcolor{MoraTint}Imp.\ (\%) & \cellcolor{MoraTint}\AppImp{3.060} & \cellcolor{MoraTint}\AppImp{3.108} & \cellcolor{MoraTint}\AppImp{2.847} & \cellcolor{MoraTint}\AppImp{1.950} & \cellcolor{MoraTint}\AppImp{1.755} & \cellcolor{MoraTint}\AppImp{1.193} & \cellcolor{MoraTint}\AppImp{4.942} & \cellcolor{MoraTint}\AppImp{3.803} & \cellcolor{MoraTint}\AppImp{15.018} & \cellcolor{MoraTint}\AppImp{11.344} & \cellcolor{MoraTint}\AppImp{0.592} & \cellcolor{MoraTint}\AppImp{0.175} \\
\cmidrule[0.25pt]{2-15}
 & \multirow{3}{*}{336} & Base & \AppScore{0.408}{0} & \AppScore{0.431}{0} & \AppScore{0.423}{0} & \AppScore{0.432}{0} & \AppScore{0.418}{0} & \AppScore{0.425}{0} & \AppScore{0.429}{0} & \AppScore{0.438}{0} & \AppScore{0.599}{0} & \AppScore{0.544}{0} & \AppScore{0.397}{1} & \AppScore{0.409}{1} \\
 &  & \textbf{+MORA} & \AppScore{0.403}{1} & \AppScore{0.421}{1} & \AppScore{0.413}{1} & \AppScore{0.424}{1} & \AppScore{0.413}{1} & \AppScore{0.421}{1} & \AppScore{0.410}{1} & \AppScore{0.424}{1} & \AppScore{0.541}{1} & \AppScore{0.507}{1} & \AppScore{0.409}{0} & \AppScore{0.417}{0} \\
 &  & \cellcolor{MoraTint}Imp.\ (\%) & \cellcolor{MoraTint}\AppImp{1.323} & \cellcolor{MoraTint}\AppImp{2.224} & \cellcolor{MoraTint}\AppImp{2.438} & \cellcolor{MoraTint}\AppImp{1.891} & \cellcolor{MoraTint}\AppImp{1.130} & \cellcolor{MoraTint}\AppImp{0.777} & \cellcolor{MoraTint}\AppImp{4.487} & \cellcolor{MoraTint}\AppImp{3.206} & \cellcolor{MoraTint}\AppImp{9.598} & \cellcolor{MoraTint}\AppImp{6.871} & \cellcolor{MoraTint}\AppImp{-3.032} & \cellcolor{MoraTint}\AppImp{-1.933} \\
\cmidrule[0.25pt]{2-15}
 & \multirow{3}{*}{720} & Base & \AppScore{0.410}{1} & \AppScore{0.436}{1} & \AppScore{0.429}{0} & \AppScore{0.447}{0} & \AppScore{0.422}{0} & \AppScore{0.439}{0} & \AppScore{0.444}{0} & \AppScore{0.460}{0} & \AppScore{0.825}{0} & \AppScore{0.655}{0} & \AppScore{0.393}{1} & \AppScore{0.419}{1} \\
 &  & \textbf{+MORA} & \AppScore{0.424}{0} & \AppScore{0.446}{0} & \AppScore{0.415}{1} & \AppScore{0.435}{1} & \AppScore{0.415}{1} & \AppScore{0.434}{1} & \AppScore{0.410}{1} & \AppScore{0.438}{1} & \AppScore{0.701}{1} & \AppScore{0.583}{1} & \AppScore{0.400}{0} & \AppScore{0.430}{0} \\
 &  & \cellcolor{MoraTint}Imp.\ (\%) & \cellcolor{MoraTint}\AppImp{-3.559} & \cellcolor{MoraTint}\AppImp{-2.469} & \cellcolor{MoraTint}\AppImp{3.249} & \cellcolor{MoraTint}\AppImp{2.558} & \cellcolor{MoraTint}\AppImp{1.751} & \cellcolor{MoraTint}\AppImp{1.088} & \cellcolor{MoraTint}\AppImp{7.678} & \cellcolor{MoraTint}\AppImp{4.741} & \cellcolor{MoraTint}\AppImp{15.004} & \cellcolor{MoraTint}\AppImp{10.999} & \cellcolor{MoraTint}\AppImp{-1.824} & \cellcolor{MoraTint}\AppImp{-2.483} \\
\midrule[0.45pt]
\multirow{12}{*}{\rotatebox[origin=c]{90}{\textbf{ETTm1}}} & \multirow{3}{*}{96} & Base & \AppScore{0.313}{0} & \AppScore{0.354}{0} & \AppScore{0.341}{0} & \AppScore{0.375}{0} & \AppScore{0.321}{0} & \AppScore{0.349}{0} & \AppScore{0.323}{0} & \AppScore{0.364}{0} & \AppScore{0.344}{0} & \AppScore{0.373}{0} & \AppScore{0.312}{0} & \AppScore{0.341}{0} \\
 &  & \textbf{+MORA} & \AppScore{0.304}{1} & \AppScore{0.350}{1} & \AppScore{0.313}{1} & \AppScore{0.357}{1} & \AppScore{0.296}{1} & \AppScore{0.340}{1} & \AppScore{0.307}{1} & \AppScore{0.353}{1} & \AppScore{0.310}{1} & \AppScore{0.351}{1} & \AppScore{0.297}{1} & \AppScore{0.340}{1} \\
 &  & \cellcolor{MoraTint}Imp.\ (\%) & \cellcolor{MoraTint}\AppImp{3.056} & \cellcolor{MoraTint}\AppImp{1.275} & \cellcolor{MoraTint}\AppImp{8.057} & \cellcolor{MoraTint}\AppImp{4.800} & \cellcolor{MoraTint}\AppImp{7.705} & \cellcolor{MoraTint}\AppImp{2.537} & \cellcolor{MoraTint}\AppImp{5.183} & \cellcolor{MoraTint}\AppImp{3.112} & \cellcolor{MoraTint}\AppImp{10.051} & \cellcolor{MoraTint}\AppImp{5.962} & \cellcolor{MoraTint}\AppImp{4.663} & \cellcolor{MoraTint}\AppImp{0.352} \\
\cmidrule[0.25pt]{2-15}
 & \multirow{3}{*}{192} & Base & \AppScore{0.355}{0} & \AppScore{0.380}{0} & \AppScore{0.380}{0} & \AppScore{0.394}{0} & \AppScore{0.369}{0} & \AppScore{0.373}{0} & \AppScore{0.369}{0} & \AppScore{0.388}{0} & \AppScore{0.380}{0} & \AppScore{0.391}{0} & \AppScore{0.362}{0} & \AppScore{0.372}{0} \\
 &  & \textbf{+MORA} & \AppScore{0.338}{1} & \AppScore{0.373}{1} & \AppScore{0.350}{1} & \AppScore{0.380}{1} & \AppScore{0.346}{1} & \AppScore{0.367}{1} & \AppScore{0.346}{1} & \AppScore{0.379}{1} & \AppScore{0.343}{1} & \AppScore{0.376}{1} & \AppScore{0.340}{1} & \AppScore{0.367}{1} \\
 &  & \cellcolor{MoraTint}Imp.\ (\%) & \cellcolor{MoraTint}\AppImp{4.919} & \cellcolor{MoraTint}\AppImp{1.684} & \cellcolor{MoraTint}\AppImp{7.791} & \cellcolor{MoraTint}\AppImp{3.549} & \cellcolor{MoraTint}\AppImp{6.327} & \cellcolor{MoraTint}\AppImp{1.601} & \cellcolor{MoraTint}\AppImp{6.190} & \cellcolor{MoraTint}\AppImp{2.504} & \cellcolor{MoraTint}\AppImp{9.807} & \cellcolor{MoraTint}\AppImp{3.929} & \cellcolor{MoraTint}\AppImp{6.013} & \cellcolor{MoraTint}\AppImp{1.229} \\
\cmidrule[0.25pt]{2-15}
 & \multirow{3}{*}{336} & Base & \AppScore{0.386}{0} & \AppScore{0.403}{0} & \AppScore{0.420}{0} & \AppScore{0.419}{0} & \AppScore{0.397}{0} & \AppScore{0.394}{0} & \AppScore{0.399}{0} & \AppScore{0.409}{0} & \AppScore{0.414}{0} & \AppScore{0.415}{0} & \AppScore{0.386}{0} & \AppScore{0.390}{0} \\
 &  & \textbf{+MORA} & \AppScore{0.365}{1} & \AppScore{0.393}{1} & \AppScore{0.379}{1} & \AppScore{0.403}{1} & \AppScore{0.372}{1} & \AppScore{0.390}{1} & \AppScore{0.369}{1} & \AppScore{0.399}{1} & \AppScore{0.368}{1} & \AppScore{0.394}{1} & \AppScore{0.363}{1} & \AppScore{0.385}{1} \\
 &  & \cellcolor{MoraTint}Imp.\ (\%) & \cellcolor{MoraTint}\AppImp{5.417} & \cellcolor{MoraTint}\AppImp{2.597} & \cellcolor{MoraTint}\AppImp{9.639} & \cellcolor{MoraTint}\AppImp{3.855} & \cellcolor{MoraTint}\AppImp{6.318} & \cellcolor{MoraTint}\AppImp{1.128} & \cellcolor{MoraTint}\AppImp{7.586} & \cellcolor{MoraTint}\AppImp{2.505} & \cellcolor{MoraTint}\AppImp{11.071} & \cellcolor{MoraTint}\AppImp{4.986} & \cellcolor{MoraTint}\AppImp{5.916} & \cellcolor{MoraTint}\AppImp{1.117} \\
\cmidrule[0.25pt]{2-15}
 & \multirow{3}{*}{720} & Base & \AppScore{0.452}{0} & \AppScore{0.437}{0} & \AppScore{0.486}{0} & \AppScore{0.456}{0} & \AppScore{0.485}{0} & \AppScore{0.442}{0} & \AppScore{0.458}{0} & \AppScore{0.445}{0} & \AppScore{0.475}{0} & \AppScore{0.453}{0} & \AppScore{0.445}{0} & \AppScore{0.425}{0} \\
 &  & \textbf{+MORA} & \AppScore{0.422}{1} & \AppScore{0.426}{1} & \AppScore{0.432}{1} & \AppScore{0.434}{1} & \AppScore{0.428}{1} & \AppScore{0.422}{1} & \AppScore{0.426}{1} & \AppScore{0.429}{1} & \AppScore{0.427}{1} & \AppScore{0.426}{1} & \AppScore{0.418}{1} & \AppScore{0.418}{1} \\
 &  & \cellcolor{MoraTint}Imp.\ (\%) & \cellcolor{MoraTint}\AppImp{6.667} & \cellcolor{MoraTint}\AppImp{2.628} & \cellcolor{MoraTint}\AppImp{11.114} & \cellcolor{MoraTint}\AppImp{4.873} & \cellcolor{MoraTint}\AppImp{11.800} & \cellcolor{MoraTint}\AppImp{4.609} & \cellcolor{MoraTint}\AppImp{7.127} & \cellcolor{MoraTint}\AppImp{3.567} & \cellcolor{MoraTint}\AppImp{10.072} & \cellcolor{MoraTint}\AppImp{5.898} & \cellcolor{MoraTint}\AppImp{5.968} & \cellcolor{MoraTint}\AppImp{1.611} \\
\midrule[0.45pt]
\multirow{12}{*}{\rotatebox[origin=c]{90}{\textbf{ETTm2}}} & \multirow{3}{*}{96} & Base & \AppScore{0.169}{0} & \AppScore{0.255}{0} & \AppScore{0.185}{0} & \AppScore{0.272}{0} & \AppScore{0.172}{0} & \AppScore{0.249}{0} & \AppScore{0.183}{0} & \AppScore{0.268}{0} & \AppScore{0.195}{0} & \AppScore{0.295}{0} & \AppScore{0.173}{0} & \AppScore{0.247}{0} \\
 &  & \textbf{+MORA} & \AppScore{0.165}{1} & \AppScore{0.252}{1} & \AppScore{0.169}{1} & \AppScore{0.254}{1} & \AppScore{0.165}{1} & \AppScore{0.247}{1} & \AppScore{0.171}{1} & \AppScore{0.258}{1} & \AppScore{0.175}{1} & \AppScore{0.265}{1} & \AppScore{0.167}{1} & \AppScore{0.244}{1} \\
 &  & \cellcolor{MoraTint}Imp.\ (\%) & \cellcolor{MoraTint}\AppImp{1.887} & \cellcolor{MoraTint}\AppImp{1.126} & \cellcolor{MoraTint}\AppImp{8.865} & \cellcolor{MoraTint}\AppImp{6.487} & \cellcolor{MoraTint}\AppImp{4.080} & \cellcolor{MoraTint}\AppImp{1.056} & \cellcolor{MoraTint}\AppImp{6.539} & \cellcolor{MoraTint}\AppImp{3.623} & \cellcolor{MoraTint}\AppImp{10.288} & \cellcolor{MoraTint}\AppImp{10.081} & \cellcolor{MoraTint}\AppImp{3.457} & \cellcolor{MoraTint}\AppImp{0.988} \\
\cmidrule[0.25pt]{2-15}
 & \multirow{3}{*}{192} & Base & \AppScore{0.235}{0} & \AppScore{0.298}{0} & \AppScore{0.253}{0} & \AppScore{0.311}{0} & \AppScore{0.239}{0} & \AppScore{0.294}{0} & \AppScore{0.247}{0} & \AppScore{0.309}{0} & \AppScore{0.285}{0} & \AppScore{0.363}{0} & \AppScore{0.236}{0} & \AppScore{0.291}{0} \\
 &  & \textbf{+MORA} & \AppScore{0.229}{1} & \AppScore{0.296}{1} & \AppScore{0.233}{1} & \AppScore{0.300}{1} & \AppScore{0.229}{1} & \AppScore{0.290}{1} & \AppScore{0.233}{1} & \AppScore{0.302}{1} & \AppScore{0.251}{1} & \AppScore{0.323}{1} & \AppScore{0.229}{1} & \AppScore{0.288}{1} \\
 &  & \cellcolor{MoraTint}Imp.\ (\%) & \cellcolor{MoraTint}\AppImp{2.355} & \cellcolor{MoraTint}\AppImp{0.971} & \cellcolor{MoraTint}\AppImp{7.956} & \cellcolor{MoraTint}\AppImp{3.390} & \cellcolor{MoraTint}\AppImp{4.505} & \cellcolor{MoraTint}\AppImp{1.269} & \cellcolor{MoraTint}\AppImp{5.720} & \cellcolor{MoraTint}\AppImp{2.290} & \cellcolor{MoraTint}\AppImp{11.955} & \cellcolor{MoraTint}\AppImp{11.012} & \cellcolor{MoraTint}\AppImp{3.033} & \cellcolor{MoraTint}\AppImp{1.022} \\
\cmidrule[0.25pt]{2-15}
 & \multirow{3}{*}{336} & Base & \AppScore{0.293}{0} & \AppScore{0.336}{0} & \AppScore{0.316}{0} & \AppScore{0.353}{0} & \AppScore{0.304}{0} & \AppScore{0.337}{0} & \AppScore{0.318}{0} & \AppScore{0.352}{0} & \AppScore{0.379}{0} & \AppScore{0.424}{0} & \AppScore{0.301}{0} & \AppScore{0.333}{0} \\
 &  & \textbf{+MORA} & \AppScore{0.286}{1} & \AppScore{0.333}{1} & \AppScore{0.293}{1} & \AppScore{0.341}{1} & \AppScore{0.289}{1} & \AppScore{0.330}{1} & \AppScore{0.299}{1} & \AppScore{0.345}{1} & \AppScore{0.336}{1} & \AppScore{0.380}{1} & \AppScore{0.290}{1} & \AppScore{0.329}{1} \\
 &  & \cellcolor{MoraTint}Imp.\ (\%) & \cellcolor{MoraTint}\AppImp{2.205} & \cellcolor{MoraTint}\AppImp{0.783} & \cellcolor{MoraTint}\AppImp{7.219} & \cellcolor{MoraTint}\AppImp{3.481} & \cellcolor{MoraTint}\AppImp{5.011} & \cellcolor{MoraTint}\AppImp{2.020} & \cellcolor{MoraTint}\AppImp{5.806} & \cellcolor{MoraTint}\AppImp{1.904} & \cellcolor{MoraTint}\AppImp{11.226} & \cellcolor{MoraTint}\AppImp{10.345} & \cellcolor{MoraTint}\AppImp{3.688} & \cellcolor{MoraTint}\AppImp{1.153} \\
\cmidrule[0.25pt]{2-15}
 & \multirow{3}{*}{720} & Base & \AppScore{0.390}{0} & \AppScore{0.393}{0} & \AppScore{0.406}{0} & \AppScore{0.403}{0} & \AppScore{0.397}{0} & \AppScore{0.393}{0} & \AppScore{0.426}{0} & \AppScore{0.414}{0} & \AppScore{0.548}{0} & \AppScore{0.520}{0} & \AppScore{0.387}{0} & \AppScore{0.388}{0} \\
 &  & \textbf{+MORA} & \AppScore{0.376}{1} & \AppScore{0.388}{1} & \AppScore{0.384}{1} & \AppScore{0.392}{1} & \AppScore{0.377}{1} & \AppScore{0.385}{1} & \AppScore{0.397}{1} & \AppScore{0.402}{1} & \AppScore{0.451}{1} & \AppScore{0.449}{1} & \AppScore{0.373}{1} & \AppScore{0.382}{1} \\
 &  & \cellcolor{MoraTint}Imp.\ (\%) & \cellcolor{MoraTint}\AppImp{3.454} & \cellcolor{MoraTint}\AppImp{1.318} & \cellcolor{MoraTint}\AppImp{5.592} & \cellcolor{MoraTint}\AppImp{2.757} & \cellcolor{MoraTint}\AppImp{4.915} & \cellcolor{MoraTint}\AppImp{2.166} & \cellcolor{MoraTint}\AppImp{6.805} & \cellcolor{MoraTint}\AppImp{2.896} & \cellcolor{MoraTint}\AppImp{17.850} & \cellcolor{MoraTint}\AppImp{13.590} & \cellcolor{MoraTint}\AppImp{3.440} & \cellcolor{MoraTint}\AppImp{1.525} \\
\bottomrule[0.8pt]
\end{tabular}
\end{table}

\clearpage
\begin{table}[H]
\centering
\parbox{\linewidth}{\small Table~\ref{tab:complete_task} (continued, part 2 of 2). Complete long-horizon results for six task-specific backbones at $L=96$.}\par\vspace{3pt}
\fontsize{8}{8.6}\selectfont
\setlength{\tabcolsep}{1pt}
\renewcommand{\arraystretch}{1.00}
\setlength{\aboverulesep}{0.6pt}
\setlength{\belowrulesep}{0.6pt}
\begin{tabular}{@{}>{\raggedright\arraybackslash}p{30pt}>{\raggedright\arraybackslash}p{14pt}>{\raggedright\arraybackslash}p{34pt}>{\raggedleft\arraybackslash}p{\dimexpr(\linewidth-78pt-28\tabcolsep)/12\relax}>{\raggedleft\arraybackslash}p{\dimexpr(\linewidth-78pt-28\tabcolsep)/12\relax}>{\raggedleft\arraybackslash}p{\dimexpr(\linewidth-78pt-28\tabcolsep)/12\relax}>{\raggedleft\arraybackslash}p{\dimexpr(\linewidth-78pt-28\tabcolsep)/12\relax}>{\raggedleft\arraybackslash}p{\dimexpr(\linewidth-78pt-28\tabcolsep)/12\relax}>{\raggedleft\arraybackslash}p{\dimexpr(\linewidth-78pt-28\tabcolsep)/12\relax}>{\raggedleft\arraybackslash}p{\dimexpr(\linewidth-78pt-28\tabcolsep)/12\relax}>{\raggedleft\arraybackslash}p{\dimexpr(\linewidth-78pt-28\tabcolsep)/12\relax}>{\raggedleft\arraybackslash}p{\dimexpr(\linewidth-78pt-28\tabcolsep)/12\relax}>{\raggedleft\arraybackslash}p{\dimexpr(\linewidth-78pt-28\tabcolsep)/12\relax}>{\raggedleft\arraybackslash}p{\dimexpr(\linewidth-78pt-28\tabcolsep)/12\relax}>{\raggedleft\arraybackslash}p{\dimexpr(\linewidth-78pt-28\tabcolsep)/12\relax}@{}}
\toprule[0.8pt]
\textbf{Dataset} & $H$ & Forecast & \multicolumn{2}{c}{\textbf{TimeFilter}} & \multicolumn{2}{c}{\textbf{iTransformer}} & \multicolumn{2}{c}{\textbf{CGTFra}} & \multicolumn{2}{c}{\textbf{PatchTST}} & \multicolumn{2}{c}{\textbf{DLinear}} & \multicolumn{2}{c}{\textbf{VPNet}} \\
\cmidrule(lr){4-5}\cmidrule(lr){6-7}\cmidrule(lr){8-9}\cmidrule(lr){10-11}\cmidrule(lr){12-13}\cmidrule(lr){14-15}
 & & & MSE & MAE & MSE & MAE & MSE & MAE & MSE & MAE & MSE & MAE & MSE & MAE \\
\midrule
\multirow{12}{*}{\rotatebox[origin=c]{90}{\textbf{Weather}}} & \multirow{3}{*}{96} & Base & \AppScore{0.153}{0} & \AppScore{0.199}{0} & \AppScore{0.174}{0} & \AppScore{0.213}{0} & \AppScore{0.153}{0} & \AppScore{0.190}{1} & \AppScore{0.176}{0} & \AppScore{0.217}{0} & \AppScore{0.195}{0} & \AppScore{0.255}{0} & \AppScore{0.154}{0} & \AppScore{0.192}{0} \\
 &  & \textbf{+MORA} & \AppScore{0.148}{1} & \AppScore{0.198}{1} & \AppScore{0.151}{1} & \AppScore{0.210}{1} & \AppScore{0.144}{1} & \AppScore{0.191}{0} & \AppScore{0.151}{1} & \AppScore{0.212}{1} & \AppScore{0.151}{1} & \AppScore{0.219}{1} & \AppScore{0.149}{1} & \AppScore{0.191}{1} \\
 &  & \cellcolor{MoraTint}Imp.\ (\%) & \cellcolor{MoraTint}\AppImp{3.505} & \cellcolor{MoraTint}\AppImp{0.523} & \cellcolor{MoraTint}\AppImp{13.153} & \cellcolor{MoraTint}\AppImp{1.671} & \cellcolor{MoraTint}\AppImp{5.384} & \cellcolor{MoraTint}\AppImp{-0.091} & \cellcolor{MoraTint}\AppImp{13.943} & \cellcolor{MoraTint}\AppImp{2.322} & \cellcolor{MoraTint}\AppImp{22.545} & \cellcolor{MoraTint}\AppImp{13.879} & \cellcolor{MoraTint}\AppImp{3.417} & \cellcolor{MoraTint}\AppImp{0.872} \\
\cmidrule[0.25pt]{2-15}
 & \multirow{3}{*}{192} & Base & \AppScore{0.202}{0} & \AppScore{0.246}{0} & \AppScore{0.223}{0} & \AppScore{0.255}{0} & \AppScore{0.205}{0} & \AppScore{0.240}{0} & \AppScore{0.222}{0} & \AppScore{0.257}{0} & \AppScore{0.240}{0} & \AppScore{0.299}{0} & \AppScore{0.202}{0} & \AppScore{0.240}{0} \\
 &  & \textbf{+MORA} & \AppScore{0.194}{1} & \AppScore{0.243}{1} & \AppScore{0.194}{1} & \AppScore{0.251}{1} & \AppScore{0.190}{1} & \AppScore{0.238}{1} & \AppScore{0.194}{1} & \AppScore{0.253}{1} & \AppScore{0.204}{1} & \AppScore{0.272}{1} & \AppScore{0.195}{1} & \AppScore{0.236}{1} \\
 &  & \cellcolor{MoraTint}Imp.\ (\%) & \cellcolor{MoraTint}\AppImp{4.144} & \cellcolor{MoraTint}\AppImp{1.450} & \cellcolor{MoraTint}\AppImp{12.943} & \cellcolor{MoraTint}\AppImp{1.735} & \cellcolor{MoraTint}\AppImp{7.612} & \cellcolor{MoraTint}\AppImp{1.003} & \cellcolor{MoraTint}\AppImp{12.354} & \cellcolor{MoraTint}\AppImp{1.580} & \cellcolor{MoraTint}\AppImp{14.817} & \cellcolor{MoraTint}\AppImp{9.169} & \cellcolor{MoraTint}\AppImp{3.391} & \cellcolor{MoraTint}\AppImp{1.568} \\
\cmidrule[0.25pt]{2-15}
 & \multirow{3}{*}{336} & Base & \AppScore{0.259}{0} & \AppScore{0.289}{0} & \AppScore{0.281}{0} & \AppScore{0.298}{0} & \AppScore{0.260}{0} & \AppScore{0.282}{0} & \AppScore{0.280}{0} & \AppScore{0.299}{0} & \AppScore{0.284}{0} & \AppScore{0.335}{0} & \AppScore{0.259}{0} & \AppScore{0.281}{0} \\
 &  & \textbf{+MORA} & \AppScore{0.247}{1} & \AppScore{0.285}{1} & \AppScore{0.247}{1} & \AppScore{0.293}{1} & \AppScore{0.241}{1} & \AppScore{0.275}{1} & \AppScore{0.247}{1} & \AppScore{0.295}{1} & \AppScore{0.258}{1} & \AppScore{0.313}{1} & \AppScore{0.247}{1} & \AppScore{0.278}{1} \\
 &  & \cellcolor{MoraTint}Imp.\ (\%) & \cellcolor{MoraTint}\AppImp{4.839} & \cellcolor{MoraTint}\AppImp{1.462} & \cellcolor{MoraTint}\AppImp{12.270} & \cellcolor{MoraTint}\AppImp{1.779} & \cellcolor{MoraTint}\AppImp{7.612} & \cellcolor{MoraTint}\AppImp{2.394} & \cellcolor{MoraTint}\AppImp{11.930} & \cellcolor{MoraTint}\AppImp{1.201} & \cellcolor{MoraTint}\AppImp{9.358} & \cellcolor{MoraTint}\AppImp{6.615} & \cellcolor{MoraTint}\AppImp{4.557} & \cellcolor{MoraTint}\AppImp{0.999} \\
\cmidrule[0.25pt]{2-15}
 & \multirow{3}{*}{720} & Base & \AppScore{0.345}{0} & \AppScore{0.344}{0} & \AppScore{0.358}{0} & \AppScore{0.350}{0} & \AppScore{0.342}{0} & \AppScore{0.336}{0} & \AppScore{0.354}{0} & \AppScore{0.348}{0} & \AppScore{0.348}{0} & \AppScore{0.385}{0} & \AppScore{0.337}{0} & \AppScore{0.336}{0} \\
 &  & \textbf{+MORA} & \AppScore{0.319}{1} & \AppScore{0.335}{1} & \AppScore{0.331}{1} & \AppScore{0.345}{1} & \AppScore{0.315}{1} & \AppScore{0.326}{1} & \AppScore{0.330}{1} & \AppScore{0.347}{1} & \AppScore{0.330}{1} & \AppScore{0.374}{1} & \AppScore{0.322}{1} & \AppScore{0.332}{1} \\
 &  & \cellcolor{MoraTint}Imp.\ (\%) & \cellcolor{MoraTint}\AppImp{7.736} & \cellcolor{MoraTint}\AppImp{2.708} & \cellcolor{MoraTint}\AppImp{7.663} & \cellcolor{MoraTint}\AppImp{1.341} & \cellcolor{MoraTint}\AppImp{7.888} & \cellcolor{MoraTint}\AppImp{2.815} & \cellcolor{MoraTint}\AppImp{6.860} & \cellcolor{MoraTint}\AppImp{0.141} & \cellcolor{MoraTint}\AppImp{5.220} & \cellcolor{MoraTint}\AppImp{2.983} & \cellcolor{MoraTint}\AppImp{4.263} & \cellcolor{MoraTint}\AppImp{1.257} \\
\midrule[0.45pt]
\multirow{12}{*}{\rotatebox[origin=c]{90}{\textbf{ECL}}} & \multirow{3}{*}{96} & Base & \AppScore{0.135}{0} & \AppScore{0.233}{0} & \AppScore{0.148}{0} & \AppScore{0.240}{0} & \AppScore{0.139}{0} & \AppScore{0.228}{0} & \AppScore{0.180}{0} & \AppScore{0.273}{0} & \AppScore{0.195}{0} & \AppScore{0.277}{0} & \AppScore{0.135}{0} & \AppScore{0.224}{0} \\
 &  & \textbf{+MORA} & \AppScore{0.130}{1} & \AppScore{0.229}{1} & \AppScore{0.136}{1} & \AppScore{0.234}{1} & \AppScore{0.132}{1} & \AppScore{0.226}{1} & \AppScore{0.150}{1} & \AppScore{0.253}{1} & \AppScore{0.156}{1} & \AppScore{0.258}{1} & \AppScore{0.130}{1} & \AppScore{0.223}{1} \\
 &  & \cellcolor{MoraTint}Imp.\ (\%) & \cellcolor{MoraTint}\AppImp{3.618} & \cellcolor{MoraTint}\AppImp{1.409} & \cellcolor{MoraTint}\AppImp{8.454} & \cellcolor{MoraTint}\AppImp{2.368} & \cellcolor{MoraTint}\AppImp{4.623} & \cellcolor{MoraTint}\AppImp{1.083} & \cellcolor{MoraTint}\AppImp{16.788} & \cellcolor{MoraTint}\AppImp{7.136} & \cellcolor{MoraTint}\AppImp{19.679} & \cellcolor{MoraTint}\AppImp{6.946} & \cellcolor{MoraTint}\AppImp{3.873} & \cellcolor{MoraTint}\AppImp{0.783} \\
\cmidrule[0.25pt]{2-15}
 & \multirow{3}{*}{192} & Base & \AppScore{0.158}{0} & \AppScore{0.253}{0} & \AppScore{0.166}{0} & \AppScore{0.257}{0} & \AppScore{0.155}{0} & \AppScore{0.243}{0} & \AppScore{0.188}{0} & \AppScore{0.280}{0} & \AppScore{0.194}{0} & \AppScore{0.280}{0} & \AppScore{0.152}{0} & \AppScore{0.240}{0} \\
 &  & \textbf{+MORA} & \AppScore{0.150}{1} & \AppScore{0.247}{1} & \AppScore{0.152}{1} & \AppScore{0.249}{1} & \AppScore{0.148}{1} & \AppScore{0.240}{1} & \AppScore{0.160}{1} & \AppScore{0.261}{1} & \AppScore{0.162}{1} & \AppScore{0.262}{1} & \AppScore{0.146}{1} & \AppScore{0.238}{1} \\
 &  & \cellcolor{MoraTint}Imp.\ (\%) & \cellcolor{MoraTint}\AppImp{5.580} & \cellcolor{MoraTint}\AppImp{2.411} & \cellcolor{MoraTint}\AppImp{8.171} & \cellcolor{MoraTint}\AppImp{2.842} & \cellcolor{MoraTint}\AppImp{4.441} & \cellcolor{MoraTint}\AppImp{1.107} & \cellcolor{MoraTint}\AppImp{14.528} & \cellcolor{MoraTint}\AppImp{6.697} & \cellcolor{MoraTint}\AppImp{16.359} & \cellcolor{MoraTint}\AppImp{6.497} & \cellcolor{MoraTint}\AppImp{4.164} & \cellcolor{MoraTint}\AppImp{1.025} \\
\cmidrule[0.25pt]{2-15}
 & \multirow{3}{*}{336} & Base & \AppScore{0.166}{0} & \AppScore{0.264}{0} & \AppScore{0.179}{0} & \AppScore{0.271}{0} & \AppScore{0.170}{0} & \AppScore{0.260}{0} & \AppScore{0.204}{0} & \AppScore{0.296}{0} & \AppScore{0.207}{0} & \AppScore{0.296}{0} & \AppScore{0.169}{0} & \AppScore{0.258}{0} \\
 &  & \textbf{+MORA} & \AppScore{0.158}{1} & \AppScore{0.259}{1} & \AppScore{0.164}{1} & \AppScore{0.263}{1} & \AppScore{0.162}{1} & \AppScore{0.256}{1} & \AppScore{0.175}{1} & \AppScore{0.276}{1} & \AppScore{0.177}{1} & \AppScore{0.278}{1} & \AppScore{0.160}{1} & \AppScore{0.254}{1} \\
 &  & \cellcolor{MoraTint}Imp.\ (\%) & \cellcolor{MoraTint}\AppImp{4.588} & \cellcolor{MoraTint}\AppImp{1.933} & \cellcolor{MoraTint}\AppImp{8.115} & \cellcolor{MoraTint}\AppImp{2.967} & \cellcolor{MoraTint}\AppImp{4.841} & \cellcolor{MoraTint}\AppImp{1.394} & \cellcolor{MoraTint}\AppImp{14.147} & \cellcolor{MoraTint}\AppImp{6.584} & \cellcolor{MoraTint}\AppImp{14.562} & \cellcolor{MoraTint}\AppImp{6.061} & \cellcolor{MoraTint}\AppImp{5.388} & \cellcolor{MoraTint}\AppImp{1.466} \\
\cmidrule[0.25pt]{2-15}
 & \multirow{3}{*}{720} & Base & \AppScore{0.196}{0} & \AppScore{0.294}{0} & \AppScore{0.209}{0} & \AppScore{0.299}{0} & \AppScore{0.204}{0} & \AppScore{0.288}{0} & \AppScore{0.245}{0} & \AppScore{0.328}{0} & \AppScore{0.242}{0} & \AppScore{0.329}{0} & \AppScore{0.200}{0} & \AppScore{0.287}{0} \\
 &  & \textbf{+MORA} & \AppScore{0.186}{1} & \AppScore{0.288}{1} & \AppScore{0.190}{1} & \AppScore{0.288}{1} & \AppScore{0.194}{1} & \AppScore{0.286}{1} & \AppScore{0.212}{1} & \AppScore{0.307}{1} & \AppScore{0.214}{1} & \AppScore{0.312}{1} & \AppScore{0.189}{1} & \AppScore{0.283}{1} \\
 &  & \cellcolor{MoraTint}Imp.\ (\%) & \cellcolor{MoraTint}\AppImp{5.255} & \cellcolor{MoraTint}\AppImp{1.973} & \cellcolor{MoraTint}\AppImp{9.358} & \cellcolor{MoraTint}\AppImp{3.639} & \cellcolor{MoraTint}\AppImp{4.633} & \cellcolor{MoraTint}\AppImp{0.579} & \cellcolor{MoraTint}\AppImp{13.510} & \cellcolor{MoraTint}\AppImp{6.194} & \cellcolor{MoraTint}\AppImp{11.696} & \cellcolor{MoraTint}\AppImp{5.201} & \cellcolor{MoraTint}\AppImp{5.362} & \cellcolor{MoraTint}\AppImp{1.285} \\
\midrule[0.45pt]
\multirow{12}{*}{\rotatebox[origin=c]{90}{\textbf{Traffic}}} & \multirow{3}{*}{96} & Base & \AppScore{0.375}{0} & \AppScore{0.251}{0} & \AppScore{0.392}{0} & \AppScore{0.267}{0} & \AppScore{0.390}{0} & \AppScore{0.240}{0} & \AppScore{0.460}{0} & \AppScore{0.297}{0} & \AppScore{0.650}{0} & \AppScore{0.397}{0} & \AppScore{0.384}{0} & \AppScore{0.257}{0} \\
 &  & \textbf{+MORA} & \AppScore{0.368}{1} & \AppScore{0.249}{1} & \AppScore{0.380}{1} & \AppScore{0.262}{1} & \AppScore{0.384}{1} & \AppScore{0.239}{1} & \AppScore{0.432}{1} & \AppScore{0.285}{1} & \AppScore{0.530}{1} & \AppScore{0.340}{1} & \AppScore{0.376}{1} & \AppScore{0.255}{1} \\
 &  & \cellcolor{MoraTint}Imp.\ (\%) & \cellcolor{MoraTint}\AppImp{1.890} & \cellcolor{MoraTint}\AppImp{1.050} & \cellcolor{MoraTint}\AppImp{2.931} & \cellcolor{MoraTint}\AppImp{2.075} & \cellcolor{MoraTint}\AppImp{1.496} & \cellcolor{MoraTint}\AppImp{0.255} & \cellcolor{MoraTint}\AppImp{5.969} & \cellcolor{MoraTint}\AppImp{4.234} & \cellcolor{MoraTint}\AppImp{18.453} & \cellcolor{MoraTint}\AppImp{14.303} & \cellcolor{MoraTint}\AppImp{2.078} & \cellcolor{MoraTint}\AppImp{0.980} \\
\cmidrule[0.25pt]{2-15}
 & \multirow{3}{*}{192} & Base & \AppScore{0.396}{0} & \AppScore{0.262}{0} & \AppScore{0.415}{0} & \AppScore{0.278}{0} & \AppScore{0.422}{0} & \AppScore{0.249}{0} & \AppScore{0.467}{0} & \AppScore{0.301}{0} & \AppScore{0.599}{0} & \AppScore{0.371}{0} & \AppScore{0.408}{0} & \AppScore{0.267}{0} \\
 &  & \textbf{+MORA} & \AppScore{0.387}{1} & \AppScore{0.259}{1} & \AppScore{0.401}{1} & \AppScore{0.271}{1} & \AppScore{0.415}{1} & \AppScore{0.248}{1} & \AppScore{0.439}{1} & \AppScore{0.288}{1} & \AppScore{0.501}{1} & \AppScore{0.324}{1} & \AppScore{0.401}{1} & \AppScore{0.264}{1} \\
 &  & \cellcolor{MoraTint}Imp.\ (\%) & \cellcolor{MoraTint}\AppImp{2.224} & \cellcolor{MoraTint}\AppImp{1.346} & \cellcolor{MoraTint}\AppImp{3.356} & \cellcolor{MoraTint}\AppImp{2.517} & \cellcolor{MoraTint}\AppImp{1.666} & \cellcolor{MoraTint}\AppImp{0.207} & \cellcolor{MoraTint}\AppImp{6.001} & \cellcolor{MoraTint}\AppImp{4.410} & \cellcolor{MoraTint}\AppImp{16.338} & \cellcolor{MoraTint}\AppImp{12.698} & \cellcolor{MoraTint}\AppImp{1.879} & \cellcolor{MoraTint}\AppImp{1.107} \\
\cmidrule[0.25pt]{2-15}
 & \multirow{3}{*}{336} & Base & \AppScore{0.415}{0} & \AppScore{0.271}{0} & \AppScore{0.427}{0} & \AppScore{0.284}{0} & \AppScore{0.438}{0} & \AppScore{0.261}{0} & \AppScore{0.482}{0} & \AppScore{0.307}{0} & \AppScore{0.606}{0} & \AppScore{0.374}{0} & \AppScore{0.427}{0} & \AppScore{0.275}{0} \\
 &  & \textbf{+MORA} & \AppScore{0.406}{1} & \AppScore{0.268}{1} & \AppScore{0.415}{1} & \AppScore{0.278}{1} & \AppScore{0.430}{1} & \AppScore{0.261}{0} & \AppScore{0.457}{1} & \AppScore{0.296}{1} & \AppScore{0.522}{1} & \AppScore{0.336}{1} & \AppScore{0.419}{1} & \AppScore{0.273}{1} \\
 &  & \cellcolor{MoraTint}Imp.\ (\%) & \cellcolor{MoraTint}\AppImp{2.080} & \cellcolor{MoraTint}\AppImp{1.194} & \cellcolor{MoraTint}\AppImp{2.897} & \cellcolor{MoraTint}\AppImp{1.949} & \cellcolor{MoraTint}\AppImp{1.841} & \cellcolor{MoraTint}\AppImp{0.104} & \cellcolor{MoraTint}\AppImp{5.302} & \cellcolor{MoraTint}\AppImp{3.463} & \cellcolor{MoraTint}\AppImp{13.813} & \cellcolor{MoraTint}\AppImp{10.232} & \cellcolor{MoraTint}\AppImp{2.046} & \cellcolor{MoraTint}\AppImp{0.879} \\
\cmidrule[0.25pt]{2-15}
 & \multirow{3}{*}{720} & Base & \AppScore{0.447}{0} & \AppScore{0.291}{0} & \AppScore{0.457}{0} & \AppScore{0.300}{0} & \AppScore{0.471}{0} & \AppScore{0.277}{1} & \AppScore{0.516}{0} & \AppScore{0.325}{0} & \AppScore{0.646}{0} & \AppScore{0.396}{0} & \AppScore{0.465}{0} & \AppScore{0.296}{0} \\
 &  & \textbf{+MORA} & \AppScore{0.440}{1} & \AppScore{0.289}{1} & \AppScore{0.448}{1} & \AppScore{0.297}{1} & \AppScore{0.465}{1} & \AppScore{0.278}{0} & \AppScore{0.498}{1} & \AppScore{0.317}{1} & \AppScore{0.594}{1} & \AppScore{0.373}{1} & \AppScore{0.458}{1} & \AppScore{0.294}{1} \\
 &  & \cellcolor{MoraTint}Imp.\ (\%) & \cellcolor{MoraTint}\AppImp{1.569} & \cellcolor{MoraTint}\AppImp{0.653} & \cellcolor{MoraTint}\AppImp{1.969} & \cellcolor{MoraTint}\AppImp{0.890} & \cellcolor{MoraTint}\AppImp{1.398} & \cellcolor{MoraTint}\AppImp{-0.190} & \cellcolor{MoraTint}\AppImp{3.494} & \cellcolor{MoraTint}\AppImp{2.416} & \cellcolor{MoraTint}\AppImp{8.006} & \cellcolor{MoraTint}\AppImp{5.631} & \cellcolor{MoraTint}\AppImp{1.587} & \cellcolor{MoraTint}\AppImp{0.599} \\
\midrule[0.45pt]
\multirow{12}{*}{\rotatebox[origin=c]{90}{\textbf{Solar}}} & \multirow{3}{*}{96} & Base & \AppScore{0.197}{0} & \AppScore{0.232}{0} & \AppScore{0.203}{0} & \AppScore{0.237}{0} & \AppScore{0.192}{0} & \AppScore{0.207}{1} & \AppScore{0.223}{0} & \AppScore{0.278}{0} & \AppScore{0.290}{0} & \AppScore{0.377}{0} & \AppScore{0.176}{0} & \AppScore{0.190}{1} \\
 &  & \textbf{+MORA} & \AppScore{0.171}{1} & \AppScore{0.225}{1} & \AppScore{0.171}{1} & \AppScore{0.228}{1} & \AppScore{0.175}{1} & \AppScore{0.209}{0} & \AppScore{0.174}{1} & \AppScore{0.248}{1} & \AppScore{0.177}{1} & \AppScore{0.253}{1} & \AppScore{0.170}{1} & \AppScore{0.192}{0} \\
 &  & \cellcolor{MoraTint}Imp.\ (\%) & \cellcolor{MoraTint}\AppImp{13.130} & \cellcolor{MoraTint}\AppImp{2.871} & \cellcolor{MoraTint}\AppImp{15.631} & \cellcolor{MoraTint}\AppImp{3.670} & \cellcolor{MoraTint}\AppImp{8.900} & \cellcolor{MoraTint}\AppImp{-0.762} & \cellcolor{MoraTint}\AppImp{22.035} & \cellcolor{MoraTint}\AppImp{10.796} & \cellcolor{MoraTint}\AppImp{38.906} & \cellcolor{MoraTint}\AppImp{32.980} & \cellcolor{MoraTint}\AppImp{3.163} & \cellcolor{MoraTint}\AppImp{-0.814} \\
\cmidrule[0.25pt]{2-15}
 & \multirow{3}{*}{192} & Base & \AppScore{0.228}{0} & \AppScore{0.257}{0} & \AppScore{0.238}{0} & \AppScore{0.263}{0} & \AppScore{0.221}{0} & \AppScore{0.227}{1} & \AppScore{0.260}{0} & \AppScore{0.303}{0} & \AppScore{0.320}{0} & \AppScore{0.398}{0} & \AppScore{0.195}{0} & \AppScore{0.206}{1} \\
 &  & \textbf{+MORA} & \AppScore{0.201}{1} & \AppScore{0.251}{1} & \AppScore{0.194}{1} & \AppScore{0.249}{1} & \AppScore{0.204}{1} & \AppScore{0.228}{0} & \AppScore{0.198}{1} & \AppScore{0.266}{1} & \AppScore{0.200}{1} & \AppScore{0.271}{1} & \AppScore{0.189}{1} & \AppScore{0.207}{0} \\
 &  & \cellcolor{MoraTint}Imp.\ (\%) & \cellcolor{MoraTint}\AppImp{12.056} & \cellcolor{MoraTint}\AppImp{2.465} & \cellcolor{MoraTint}\AppImp{18.301} & \cellcolor{MoraTint}\AppImp{5.446} & \cellcolor{MoraTint}\AppImp{7.695} & \cellcolor{MoraTint}\AppImp{-0.314} & \cellcolor{MoraTint}\AppImp{23.851} & \cellcolor{MoraTint}\AppImp{12.277} & \cellcolor{MoraTint}\AppImp{37.287} & \cellcolor{MoraTint}\AppImp{31.772} & \cellcolor{MoraTint}\AppImp{3.342} & \cellcolor{MoraTint}\AppImp{-0.476} \\
\cmidrule[0.25pt]{2-15}
 & \multirow{3}{*}{336} & Base & \AppScore{0.239}{0} & \AppScore{0.269}{0} & \AppScore{0.250}{0} & \AppScore{0.274}{0} & \AppScore{0.241}{0} & \AppScore{0.242}{0} & \AppScore{0.277}{0} & \AppScore{0.308}{0} & \AppScore{0.352}{0} & \AppScore{0.415}{0} & \AppScore{0.217}{0} & \AppScore{0.221}{0} \\
 &  & \textbf{+MORA} & \AppScore{0.204}{1} & \AppScore{0.258}{1} & \AppScore{0.207}{1} & \AppScore{0.259}{1} & \AppScore{0.221}{1} & \AppScore{0.239}{1} & \AppScore{0.217}{1} & \AppScore{0.279}{1} & \AppScore{0.221}{1} & \AppScore{0.281}{1} & \AppScore{0.211}{1} & \AppScore{0.221}{0} \\
 &  & \cellcolor{MoraTint}Imp.\ (\%) & \cellcolor{MoraTint}\AppImp{14.671} & \cellcolor{MoraTint}\AppImp{4.224} & \cellcolor{MoraTint}\AppImp{17.291} & \cellcolor{MoraTint}\AppImp{5.522} & \cellcolor{MoraTint}\AppImp{8.019} & \cellcolor{MoraTint}\AppImp{1.464} & \cellcolor{MoraTint}\AppImp{21.616} & \cellcolor{MoraTint}\AppImp{9.454} & \cellcolor{MoraTint}\AppImp{37.231} & \cellcolor{MoraTint}\AppImp{32.337} & \cellcolor{MoraTint}\AppImp{3.018} & \cellcolor{MoraTint}\AppImp{0.071} \\
\cmidrule[0.25pt]{2-15}
 & \multirow{3}{*}{720} & Base & \AppScore{0.244}{0} & \AppScore{0.274}{0} & \AppScore{0.250}{0} & \AppScore{0.276}{0} & \AppScore{0.248}{0} & \AppScore{0.245}{0} & \AppScore{0.274}{0} & \AppScore{0.309}{0} & \AppScore{0.357}{0} & \AppScore{0.413}{0} & \AppScore{0.229}{0} & \AppScore{0.227}{1} \\
 &  & \textbf{+MORA} & \AppScore{0.215}{1} & \AppScore{0.267}{1} & \AppScore{0.220}{1} & \AppScore{0.269}{1} & \AppScore{0.239}{1} & \AppScore{0.244}{1} & \AppScore{0.232}{1} & \AppScore{0.285}{1} & \AppScore{0.247}{1} & \AppScore{0.306}{1} & \AppScore{0.227}{1} & \AppScore{0.228}{0} \\
 &  & \cellcolor{MoraTint}Imp.\ (\%) & \cellcolor{MoraTint}\AppImp{11.808} & \cellcolor{MoraTint}\AppImp{2.572} & \cellcolor{MoraTint}\AppImp{12.184} & \cellcolor{MoraTint}\AppImp{2.467} & \cellcolor{MoraTint}\AppImp{3.622} & \cellcolor{MoraTint}\AppImp{0.441} & \cellcolor{MoraTint}\AppImp{15.471} & \cellcolor{MoraTint}\AppImp{7.764} & \cellcolor{MoraTint}\AppImp{30.627} & \cellcolor{MoraTint}\AppImp{25.852} & \cellcolor{MoraTint}\AppImp{0.986} & \cellcolor{MoraTint}\AppImp{-0.326} \\
\bottomrule[0.8pt]
\end{tabular}
\end{table}

\clearpage

\subsection{Foundation Models at \texorpdfstring{$L=96$}{L=96}}
\label{app:foundation_results_96}
\begin{table}[H]
\centering
\caption{Complete long-horizon results for frozen foundation models at $L=96$.}
\label{tab:complete_foundation96}
\label{tab:full_foundation_96:etth1}
\label{tab:full_foundation_96:etth2}
\label{tab:full_foundation_96:ettm1}
\label{tab:full_foundation_96:ettm2}
\label{tab:full_foundation_96:weather}
\label{tab:full_foundation_96:ecl}
\label{tab:full_foundation_96:traffic}
\label{tab:full_foundation_96:solar}
\fontsize{8}{8.6}\selectfont
\setlength{\tabcolsep}{1pt}
\renewcommand{\arraystretch}{1.00}
\setlength{\aboverulesep}{0.6pt}
\setlength{\belowrulesep}{0.6pt}
\begin{tabular}{@{}>{\raggedright\arraybackslash}p{30pt}>{\raggedright\arraybackslash}p{14pt}>{\raggedright\arraybackslash}p{34pt}>{\raggedleft\arraybackslash}p{\dimexpr(\linewidth-78pt-16\tabcolsep)/6\relax}>{\raggedleft\arraybackslash}p{\dimexpr(\linewidth-78pt-16\tabcolsep)/6\relax}>{\raggedleft\arraybackslash}p{\dimexpr(\linewidth-78pt-16\tabcolsep)/6\relax}>{\raggedleft\arraybackslash}p{\dimexpr(\linewidth-78pt-16\tabcolsep)/6\relax}>{\raggedleft\arraybackslash}p{\dimexpr(\linewidth-78pt-16\tabcolsep)/6\relax}>{\raggedleft\arraybackslash}p{\dimexpr(\linewidth-78pt-16\tabcolsep)/6\relax}@{}}
\toprule[0.8pt]
\textbf{Dataset} & $H$ & Forecast & \multicolumn{2}{c}{\textbf{Chronos-2}} & \multicolumn{2}{c}{\textbf{Sundial}} & \multicolumn{2}{c}{\textbf{TimesFM 2.5}} \\
\cmidrule(lr){4-5}\cmidrule(lr){6-7}\cmidrule(lr){8-9}
 & & & MSE & MAE & MSE & MAE & MSE & MAE \\
\midrule
\multirow{12}{*}{\rotatebox[origin=c]{90}{\textbf{ETTh1}}} & \multirow{3}{*}{96} & Base & \AppScore{0.424}{0} & \AppScore{0.397}{0} & \AppScore{0.406}{0} & \AppScore{0.396}{0} & \AppScore{0.429}{0} & \AppScore{0.402}{0} \\
 &  & \textbf{+MORA} & \AppScore{0.394}{1} & \AppScore{0.393}{1} & \AppScore{0.376}{1} & \AppScore{0.392}{1} & \AppScore{0.379}{1} & \AppScore{0.394}{1} \\
 &  & \cellcolor{MoraTint}Imp.\ (\%) & \cellcolor{MoraTint}\AppImp{6.967} & \cellcolor{MoraTint}\AppImp{1.111} & \cellcolor{MoraTint}\AppImp{7.240} & \cellcolor{MoraTint}\AppImp{1.176} & \cellcolor{MoraTint}\AppImp{11.662} & \cellcolor{MoraTint}\AppImp{2.190} \\
\cmidrule[0.25pt]{2-9}
 & \multirow{3}{*}{192} & Base & \AppScore{0.484}{0} & \AppScore{0.431}{0} & \AppScore{0.469}{0} & \AppScore{0.430}{0} & \AppScore{0.485}{0} & \AppScore{0.432}{0} \\
 &  & \textbf{+MORA} & \AppScore{0.441}{1} & \AppScore{0.419}{1} & \AppScore{0.423}{1} & \AppScore{0.419}{1} & \AppScore{0.424}{1} & \AppScore{0.420}{1} \\
 &  & \cellcolor{MoraTint}Imp.\ (\%) & \cellcolor{MoraTint}\AppImp{8.880} & \cellcolor{MoraTint}\AppImp{2.795} & \cellcolor{MoraTint}\AppImp{9.854} & \cellcolor{MoraTint}\AppImp{2.636} & \cellcolor{MoraTint}\AppImp{12.682} & \cellcolor{MoraTint}\AppImp{2.680} \\
\cmidrule[0.25pt]{2-9}
 & \multirow{3}{*}{336} & Base & \AppScore{0.536}{0} & \AppScore{0.456}{0} & \AppScore{0.516}{0} & \AppScore{0.451}{0} & \AppScore{0.526}{0} & \AppScore{0.450}{0} \\
 &  & \textbf{+MORA} & \AppScore{0.492}{1} & \AppScore{0.444}{1} & \AppScore{0.479}{1} & \AppScore{0.442}{1} & \AppScore{0.466}{1} & \AppScore{0.441}{1} \\
 &  & \cellcolor{MoraTint}Imp.\ (\%) & \cellcolor{MoraTint}\AppImp{8.227} & \cellcolor{MoraTint}\AppImp{2.607} & \cellcolor{MoraTint}\AppImp{7.068} & \cellcolor{MoraTint}\AppImp{1.884} & \cellcolor{MoraTint}\AppImp{11.358} & \cellcolor{MoraTint}\AppImp{2.106} \\
\cmidrule[0.25pt]{2-9}
 & \multirow{3}{*}{720} & Base & \AppScore{0.526}{0} & \AppScore{0.471}{0} & \AppScore{0.500}{0} & \AppScore{0.461}{0} & \AppScore{0.498}{0} & \AppScore{0.457}{1} \\
 &  & \textbf{+MORA} & \AppScore{0.497}{1} & \AppScore{0.467}{1} & \AppScore{0.466}{1} & \AppScore{0.457}{1} & \AppScore{0.472}{1} & \AppScore{0.462}{0} \\
 &  & \cellcolor{MoraTint}Imp.\ (\%) & \cellcolor{MoraTint}\AppImp{5.431} & \cellcolor{MoraTint}\AppImp{0.988} & \cellcolor{MoraTint}\AppImp{6.852} & \cellcolor{MoraTint}\AppImp{0.758} & \cellcolor{MoraTint}\AppImp{5.130} & \cellcolor{MoraTint}\AppImp{-1.050} \\
\midrule[0.45pt]
\multirow{12}{*}{\rotatebox[origin=c]{90}{\textbf{ETTh2}}} & \multirow{3}{*}{96} & Base & \AppScore{0.317}{0} & \AppScore{0.344}{0} & \AppScore{0.317}{0} & \AppScore{0.353}{0} & \AppScore{0.330}{0} & \AppScore{0.345}{0} \\
 &  & \textbf{+MORA} & \AppScore{0.295}{1} & \AppScore{0.331}{1} & \AppScore{0.288}{1} & \AppScore{0.337}{1} & \AppScore{0.298}{1} & \AppScore{0.335}{1} \\
 &  & \cellcolor{MoraTint}Imp.\ (\%) & \cellcolor{MoraTint}\AppImp{7.021} & \cellcolor{MoraTint}\AppImp{3.977} & \cellcolor{MoraTint}\AppImp{9.078} & \cellcolor{MoraTint}\AppImp{4.622} & \cellcolor{MoraTint}\AppImp{9.758} & \cellcolor{MoraTint}\AppImp{2.901} \\
\cmidrule[0.25pt]{2-9}
 & \multirow{3}{*}{192} & Base & \AppScore{0.413}{0} & \AppScore{0.400}{0} & \AppScore{0.405}{0} & \AppScore{0.405}{0} & \AppScore{0.420}{0} & \AppScore{0.402}{0} \\
 &  & \textbf{+MORA} & \AppScore{0.379}{1} & \AppScore{0.382}{1} & \AppScore{0.375}{1} & \AppScore{0.388}{1} & \AppScore{0.396}{1} & \AppScore{0.391}{1} \\
 &  & \cellcolor{MoraTint}Imp.\ (\%) & \cellcolor{MoraTint}\AppImp{8.174} & \cellcolor{MoraTint}\AppImp{4.417} & \cellcolor{MoraTint}\AppImp{7.521} & \cellcolor{MoraTint}\AppImp{4.031} & \cellcolor{MoraTint}\AppImp{5.806} & \cellcolor{MoraTint}\AppImp{2.620} \\
\cmidrule[0.25pt]{2-9}
 & \multirow{3}{*}{336} & Base & \AppScore{0.449}{0} & \AppScore{0.433}{0} & \AppScore{0.438}{0} & \AppScore{0.436}{0} & \AppScore{0.453}{0} & \AppScore{0.434}{0} \\
 &  & \textbf{+MORA} & \AppScore{0.419}{1} & \AppScore{0.417}{1} & \AppScore{0.418}{1} & \AppScore{0.424}{1} & \AppScore{0.434}{1} & \AppScore{0.425}{1} \\
 &  & \cellcolor{MoraTint}Imp.\ (\%) & \cellcolor{MoraTint}\AppImp{6.601} & \cellcolor{MoraTint}\AppImp{3.812} & \cellcolor{MoraTint}\AppImp{4.479} & \cellcolor{MoraTint}\AppImp{2.620} & \cellcolor{MoraTint}\AppImp{4.285} & \cellcolor{MoraTint}\AppImp{2.192} \\
\cmidrule[0.25pt]{2-9}
 & \multirow{3}{*}{720} & Base & \AppScore{0.443}{0} & \AppScore{0.442}{0} & \AppScore{0.437}{1} & \AppScore{0.445}{1} & \AppScore{0.450}{1} & \AppScore{0.443}{1} \\
 &  & \textbf{+MORA} & \AppScore{0.441}{1} & \AppScore{0.441}{1} & \AppScore{0.474}{0} & \AppScore{0.472}{0} & \AppScore{0.462}{0} & \AppScore{0.465}{0} \\
 &  & \cellcolor{MoraTint}Imp.\ (\%) & \cellcolor{MoraTint}\AppImp{0.487} & \cellcolor{MoraTint}\AppImp{0.303} & \cellcolor{MoraTint}\AppImp{-8.556} & \cellcolor{MoraTint}\AppImp{-5.959} & \cellcolor{MoraTint}\AppImp{-2.575} & \cellcolor{MoraTint}\AppImp{-4.912} \\
\midrule[0.45pt]
\multirow{12}{*}{\rotatebox[origin=c]{90}{\textbf{ETTm1}}} & \multirow{3}{*}{96} & Base & \AppScore{0.592}{0} & \AppScore{0.452}{0} & \AppScore{0.478}{0} & \AppScore{0.420}{0} & \AppScore{0.841}{0} & \AppScore{0.560}{0} \\
 &  & \textbf{+MORA} & \AppScore{0.312}{1} & \AppScore{0.356}{1} & \AppScore{0.317}{1} & \AppScore{0.356}{1} & \AppScore{0.311}{1} & \AppScore{0.354}{1} \\
 &  & \cellcolor{MoraTint}Imp.\ (\%) & \cellcolor{MoraTint}\AppImp{47.221} & \cellcolor{MoraTint}\AppImp{21.329} & \cellcolor{MoraTint}\AppImp{33.785} & \cellcolor{MoraTint}\AppImp{15.219} & \cellcolor{MoraTint}\AppImp{63.051} & \cellcolor{MoraTint}\AppImp{36.912} \\
\cmidrule[0.25pt]{2-9}
 & \multirow{3}{*}{192} & Base & \AppScore{0.607}{0} & \AppScore{0.467}{0} & \AppScore{0.527}{0} & \AppScore{0.449}{0} & \AppScore{0.881}{0} & \AppScore{0.582}{0} \\
 &  & \textbf{+MORA} & \AppScore{0.352}{1} & \AppScore{0.379}{1} & \AppScore{0.353}{1} & \AppScore{0.379}{1} & \AppScore{0.358}{1} & \AppScore{0.380}{1} \\
 &  & \cellcolor{MoraTint}Imp.\ (\%) & \cellcolor{MoraTint}\AppImp{42.092} & \cellcolor{MoraTint}\AppImp{18.864} & \cellcolor{MoraTint}\AppImp{32.871} & \cellcolor{MoraTint}\AppImp{15.731} & \cellcolor{MoraTint}\AppImp{59.363} & \cellcolor{MoraTint}\AppImp{34.684} \\
\cmidrule[0.25pt]{2-9}
 & \multirow{3}{*}{336} & Base & \AppScore{0.618}{0} & \AppScore{0.482}{0} & \AppScore{0.548}{0} & \AppScore{0.467}{0} & \AppScore{0.872}{0} & \AppScore{0.588}{0} \\
 &  & \textbf{+MORA} & \AppScore{0.376}{1} & \AppScore{0.399}{1} & \AppScore{0.378}{1} & \AppScore{0.398}{1} & \AppScore{0.381}{1} & \AppScore{0.397}{1} \\
 &  & \cellcolor{MoraTint}Imp.\ (\%) & \cellcolor{MoraTint}\AppImp{39.086} & \cellcolor{MoraTint}\AppImp{17.133} & \cellcolor{MoraTint}\AppImp{31.103} & \cellcolor{MoraTint}\AppImp{14.681} & \cellcolor{MoraTint}\AppImp{56.328} & \cellcolor{MoraTint}\AppImp{32.427} \\
\cmidrule[0.25pt]{2-9}
 & \multirow{3}{*}{720} & Base & \AppScore{0.646}{0} & \AppScore{0.508}{0} & \AppScore{0.596}{0} & \AppScore{0.494}{0} & \AppScore{0.905}{0} & \AppScore{0.607}{0} \\
 &  & \textbf{+MORA} & \AppScore{0.425}{1} & \AppScore{0.429}{1} & \AppScore{0.428}{1} & \AppScore{0.429}{1} & \AppScore{0.431}{1} & \AppScore{0.429}{1} \\
 &  & \cellcolor{MoraTint}Imp.\ (\%) & \cellcolor{MoraTint}\AppImp{34.188} & \cellcolor{MoraTint}\AppImp{15.544} & \cellcolor{MoraTint}\AppImp{28.205} & \cellcolor{MoraTint}\AppImp{13.086} & \cellcolor{MoraTint}\AppImp{52.357} & \cellcolor{MoraTint}\AppImp{29.275} \\
\midrule[0.45pt]
\multirow{12}{*}{\rotatebox[origin=c]{90}{\textbf{ETTm2}}} & \multirow{3}{*}{96} & Base & \AppScore{0.210}{0} & \AppScore{0.273}{0} & \AppScore{0.201}{0} & \AppScore{0.276}{0} & \AppScore{0.257}{0} & \AppScore{0.312}{0} \\
 &  & \textbf{+MORA} & \AppScore{0.174}{1} & \AppScore{0.250}{1} & \AppScore{0.169}{1} & \AppScore{0.254}{1} & \AppScore{0.174}{1} & \AppScore{0.255}{1} \\
 &  & \cellcolor{MoraTint}Imp.\ (\%) & \cellcolor{MoraTint}\AppImp{16.965} & \cellcolor{MoraTint}\AppImp{8.321} & \cellcolor{MoraTint}\AppImp{15.637} & \cellcolor{MoraTint}\AppImp{8.134} & \cellcolor{MoraTint}\AppImp{32.146} & \cellcolor{MoraTint}\AppImp{18.140} \\
\cmidrule[0.25pt]{2-9}
 & \multirow{3}{*}{192} & Base & \AppScore{0.274}{0} & \AppScore{0.314}{0} & \AppScore{0.266}{0} & \AppScore{0.318}{0} & \AppScore{0.333}{0} & \AppScore{0.357}{0} \\
 &  & \textbf{+MORA} & \AppScore{0.236}{1} & \AppScore{0.291}{1} & \AppScore{0.233}{1} & \AppScore{0.299}{1} & \AppScore{0.238}{1} & \AppScore{0.299}{1} \\
 &  & \cellcolor{MoraTint}Imp.\ (\%) & \cellcolor{MoraTint}\AppImp{13.757} & \cellcolor{MoraTint}\AppImp{7.339} & \cellcolor{MoraTint}\AppImp{12.434} & \cellcolor{MoraTint}\AppImp{6.183} & \cellcolor{MoraTint}\AppImp{28.510} & \cellcolor{MoraTint}\AppImp{16.374} \\
\cmidrule[0.25pt]{2-9}
 & \multirow{3}{*}{336} & Base & \AppScore{0.340}{0} & \AppScore{0.355}{0} & \AppScore{0.326}{0} & \AppScore{0.355}{0} & \AppScore{0.395}{0} & \AppScore{0.393}{0} \\
 &  & \textbf{+MORA} & \AppScore{0.298}{1} & \AppScore{0.332}{1} & \AppScore{0.293}{1} & \AppScore{0.337}{1} & \AppScore{0.298}{1} & \AppScore{0.338}{1} \\
 &  & \cellcolor{MoraTint}Imp.\ (\%) & \cellcolor{MoraTint}\AppImp{12.324} & \cellcolor{MoraTint}\AppImp{6.508} & \cellcolor{MoraTint}\AppImp{10.226} & \cellcolor{MoraTint}\AppImp{5.103} & \cellcolor{MoraTint}\AppImp{24.502} & \cellcolor{MoraTint}\AppImp{14.135} \\
\cmidrule[0.25pt]{2-9}
 & \multirow{3}{*}{720} & Base & \AppScore{0.448}{0} & \AppScore{0.412}{0} & \AppScore{0.424}{0} & \AppScore{0.409}{0} & \AppScore{0.501}{0} & \AppScore{0.448}{0} \\
 &  & \textbf{+MORA} & \AppScore{0.393}{1} & \AppScore{0.386}{1} & \AppScore{0.389}{1} & \AppScore{0.390}{1} & \AppScore{0.397}{1} & \AppScore{0.394}{1} \\
 &  & \cellcolor{MoraTint}Imp.\ (\%) & \cellcolor{MoraTint}\AppImp{12.262} & \cellcolor{MoraTint}\AppImp{6.111} & \cellcolor{MoraTint}\AppImp{8.263} & \cellcolor{MoraTint}\AppImp{4.723} & \cellcolor{MoraTint}\AppImp{20.878} & \cellcolor{MoraTint}\AppImp{12.128} \\
\bottomrule[0.8pt]
\end{tabular}
\end{table}

\clearpage
\begin{table}[H]
\centering
\parbox{\linewidth}{\small Table~\ref{tab:complete_foundation96} (continued, part 2 of 2). Complete long-horizon results for frozen foundation models at $L=96$.}\par\vspace{3pt}
\fontsize{8}{8.6}\selectfont
\setlength{\tabcolsep}{1pt}
\renewcommand{\arraystretch}{1.00}
\setlength{\aboverulesep}{0.6pt}
\setlength{\belowrulesep}{0.6pt}
\begin{tabular}{@{}>{\raggedright\arraybackslash}p{30pt}>{\raggedright\arraybackslash}p{14pt}>{\raggedright\arraybackslash}p{34pt}>{\raggedleft\arraybackslash}p{\dimexpr(\linewidth-78pt-16\tabcolsep)/6\relax}>{\raggedleft\arraybackslash}p{\dimexpr(\linewidth-78pt-16\tabcolsep)/6\relax}>{\raggedleft\arraybackslash}p{\dimexpr(\linewidth-78pt-16\tabcolsep)/6\relax}>{\raggedleft\arraybackslash}p{\dimexpr(\linewidth-78pt-16\tabcolsep)/6\relax}>{\raggedleft\arraybackslash}p{\dimexpr(\linewidth-78pt-16\tabcolsep)/6\relax}>{\raggedleft\arraybackslash}p{\dimexpr(\linewidth-78pt-16\tabcolsep)/6\relax}@{}}
\toprule[0.8pt]
\textbf{Dataset} & $H$ & Forecast & \multicolumn{2}{c}{\textbf{Chronos-2}} & \multicolumn{2}{c}{\textbf{Sundial}} & \multicolumn{2}{c}{\textbf{TimesFM 2.5}} \\
\cmidrule(lr){4-5}\cmidrule(lr){6-7}\cmidrule(lr){8-9}
 & & & MSE & MAE & MSE & MAE & MSE & MAE \\
\midrule
\multirow{12}{*}{\rotatebox[origin=c]{90}{\textbf{Weather}}} & \multirow{3}{*}{96} & Base & \AppScore{0.244}{0} & \AppScore{0.230}{0} & \AppScore{0.195}{0} & \AppScore{0.235}{0} & \AppScore{0.207}{0} & \AppScore{0.229}{0} \\
 &  & \textbf{+MORA} & \AppScore{0.171}{1} & \AppScore{0.224}{1} & \AppScore{0.151}{1} & \AppScore{0.218}{1} & \AppScore{0.152}{1} & \AppScore{0.210}{1} \\
 &  & \cellcolor{MoraTint}Imp.\ (\%) & \cellcolor{MoraTint}\AppImp{29.872} & \cellcolor{MoraTint}\AppImp{2.529} & \cellcolor{MoraTint}\AppImp{22.542} & \cellcolor{MoraTint}\AppImp{7.029} & \cellcolor{MoraTint}\AppImp{26.717} & \cellcolor{MoraTint}\AppImp{8.175} \\
\cmidrule[0.25pt]{2-9}
 & \multirow{3}{*}{192} & Base & \AppScore{0.273}{0} & \AppScore{0.270}{0} & \AppScore{0.254}{0} & \AppScore{0.283}{0} & \AppScore{0.264}{0} & \AppScore{0.277}{0} \\
 &  & \textbf{+MORA} & \AppScore{0.206}{1} & \AppScore{0.267}{1} & \AppScore{0.196}{1} & \AppScore{0.266}{1} & \AppScore{0.194}{1} & \AppScore{0.258}{1} \\
 &  & \cellcolor{MoraTint}Imp.\ (\%) & \cellcolor{MoraTint}\AppImp{24.501} & \cellcolor{MoraTint}\AppImp{1.258} & \cellcolor{MoraTint}\AppImp{22.674} & \cellcolor{MoraTint}\AppImp{6.140} & \cellcolor{MoraTint}\AppImp{26.376} & \cellcolor{MoraTint}\AppImp{6.830} \\
\cmidrule[0.25pt]{2-9}
 & \multirow{3}{*}{336} & Base & \AppScore{0.319}{0} & \AppScore{0.305}{1} & \AppScore{0.309}{0} & \AppScore{0.322}{0} & \AppScore{0.336}{0} & \AppScore{0.318}{0} \\
 &  & \textbf{+MORA} & \AppScore{0.256}{1} & \AppScore{0.306}{0} & \AppScore{0.251}{1} & \AppScore{0.309}{1} & \AppScore{0.248}{1} & \AppScore{0.302}{1} \\
 &  & \cellcolor{MoraTint}Imp.\ (\%) & \cellcolor{MoraTint}\AppImp{19.551} & \cellcolor{MoraTint}\AppImp{-0.134} & \cellcolor{MoraTint}\AppImp{18.685} & \cellcolor{MoraTint}\AppImp{3.811} & \cellcolor{MoraTint}\AppImp{26.109} & \cellcolor{MoraTint}\AppImp{4.913} \\
\cmidrule[0.25pt]{2-9}
 & \multirow{3}{*}{720} & Base & \AppScore{0.392}{0} & \AppScore{0.357}{1} & \AppScore{0.382}{0} & \AppScore{0.369}{1} & \AppScore{0.450}{0} & \AppScore{0.374}{0} \\
 &  & \textbf{+MORA} & \AppScore{0.339}{1} & \AppScore{0.362}{0} & \AppScore{0.331}{1} & \AppScore{0.370}{0} & \AppScore{0.327}{1} & \AppScore{0.360}{1} \\
 &  & \cellcolor{MoraTint}Imp.\ (\%) & \cellcolor{MoraTint}\AppImp{13.635} & \cellcolor{MoraTint}\AppImp{-1.391} & \cellcolor{MoraTint}\AppImp{13.228} & \cellcolor{MoraTint}\AppImp{-0.046} & \cellcolor{MoraTint}\AppImp{27.335} & \cellcolor{MoraTint}\AppImp{3.647} \\
\midrule[0.45pt]
\multirow{12}{*}{\rotatebox[origin=c]{90}{\textbf{ECL}}} & \multirow{3}{*}{96} & Base & \AppScore{0.214}{0} & \AppScore{0.274}{0} & \AppScore{0.180}{0} & \AppScore{0.261}{0} & \AppScore{0.223}{0} & \AppScore{0.293}{0} \\
 &  & \textbf{+MORA} & \AppScore{0.159}{1} & \AppScore{0.253}{1} & \AppScore{0.149}{1} & \AppScore{0.247}{1} & \AppScore{0.164}{1} & \AppScore{0.264}{1} \\
 &  & \cellcolor{MoraTint}Imp.\ (\%) & \cellcolor{MoraTint}\AppImp{25.886} & \cellcolor{MoraTint}\AppImp{7.622} & \cellcolor{MoraTint}\AppImp{17.197} & \cellcolor{MoraTint}\AppImp{5.210} & \cellcolor{MoraTint}\AppImp{26.459} & \cellcolor{MoraTint}\AppImp{9.869} \\
\cmidrule[0.25pt]{2-9}
 & \multirow{3}{*}{192} & Base & \AppScore{0.213}{0} & \AppScore{0.279}{0} & \AppScore{0.187}{0} & \AppScore{0.269}{0} & \AppScore{0.220}{0} & \AppScore{0.292}{0} \\
 &  & \textbf{+MORA} & \AppScore{0.167}{1} & \AppScore{0.260}{1} & \AppScore{0.159}{1} & \AppScore{0.256}{1} & \AppScore{0.171}{1} & \AppScore{0.268}{1} \\
 &  & \cellcolor{MoraTint}Imp.\ (\%) & \cellcolor{MoraTint}\AppImp{21.610} & \cellcolor{MoraTint}\AppImp{6.879} & \cellcolor{MoraTint}\AppImp{14.650} & \cellcolor{MoraTint}\AppImp{4.846} & \cellcolor{MoraTint}\AppImp{22.592} & \cellcolor{MoraTint}\AppImp{8.431} \\
\cmidrule[0.25pt]{2-9}
 & \multirow{3}{*}{336} & Base & \AppScore{0.230}{0} & \AppScore{0.295}{0} & \AppScore{0.204}{0} & \AppScore{0.286}{0} & \AppScore{0.236}{0} & \AppScore{0.307}{0} \\
 &  & \textbf{+MORA} & \AppScore{0.184}{1} & \AppScore{0.275}{1} & \AppScore{0.174}{1} & \AppScore{0.272}{1} & \AppScore{0.186}{1} & \AppScore{0.283}{1} \\
 &  & \cellcolor{MoraTint}Imp.\ (\%) & \cellcolor{MoraTint}\AppImp{20.155} & \cellcolor{MoraTint}\AppImp{6.719} & \cellcolor{MoraTint}\AppImp{14.334} & \cellcolor{MoraTint}\AppImp{4.920} & \cellcolor{MoraTint}\AppImp{20.969} & \cellcolor{MoraTint}\AppImp{7.820} \\
\cmidrule[0.25pt]{2-9}
 & \multirow{3}{*}{720} & Base & \AppScore{0.279}{0} & \AppScore{0.329}{0} & \AppScore{0.245}{0} & \AppScore{0.319}{0} & \AppScore{0.278}{0} & \AppScore{0.341}{0} \\
 &  & \textbf{+MORA} & \AppScore{0.224}{1} & \AppScore{0.307}{1} & \AppScore{0.211}{1} & \AppScore{0.304}{1} & \AppScore{0.226}{1} & \AppScore{0.317}{1} \\
 &  & \cellcolor{MoraTint}Imp.\ (\%) & \cellcolor{MoraTint}\AppImp{19.458} & \cellcolor{MoraTint}\AppImp{6.412} & \cellcolor{MoraTint}\AppImp{13.908} & \cellcolor{MoraTint}\AppImp{4.926} & \cellcolor{MoraTint}\AppImp{18.830} & \cellcolor{MoraTint}\AppImp{7.000} \\
\midrule[0.45pt]
\multirow{12}{*}{\rotatebox[origin=c]{90}{\textbf{Traffic}}} & \multirow{3}{*}{96} & Base & \AppScore{0.574}{0} & \AppScore{0.310}{0} & \AppScore{0.605}{0} & \AppScore{0.353}{0} & \AppScore{0.602}{0} & \AppScore{0.338}{0} \\
 &  & \textbf{+MORA} & \AppScore{0.497}{1} & \AppScore{0.295}{1} & \AppScore{0.513}{1} & \AppScore{0.317}{1} & \AppScore{0.530}{1} & \AppScore{0.311}{1} \\
 &  & \cellcolor{MoraTint}Imp.\ (\%) & \cellcolor{MoraTint}\AppImp{13.280} & \cellcolor{MoraTint}\AppImp{5.018} & \cellcolor{MoraTint}\AppImp{15.189} & \cellcolor{MoraTint}\AppImp{10.148} & \cellcolor{MoraTint}\AppImp{11.970} & \cellcolor{MoraTint}\AppImp{8.010} \\
\cmidrule[0.25pt]{2-9}
 & \multirow{3}{*}{192} & Base & \AppScore{0.550}{0} & \AppScore{0.301}{0} & \AppScore{0.582}{0} & \AppScore{0.344}{0} & \AppScore{0.575}{0} & \AppScore{0.324}{0} \\
 &  & \textbf{+MORA} & \AppScore{0.491}{1} & \AppScore{0.289}{1} & \AppScore{0.502}{1} & \AppScore{0.313}{1} & \AppScore{0.515}{1} & \AppScore{0.303}{1} \\
 &  & \cellcolor{MoraTint}Imp.\ (\%) & \cellcolor{MoraTint}\AppImp{10.781} & \cellcolor{MoraTint}\AppImp{4.186} & \cellcolor{MoraTint}\AppImp{13.652} & \cellcolor{MoraTint}\AppImp{9.223} & \cellcolor{MoraTint}\AppImp{10.328} & \cellcolor{MoraTint}\AppImp{6.483} \\
\cmidrule[0.25pt]{2-9}
 & \multirow{3}{*}{336} & Base & \AppScore{0.566}{0} & \AppScore{0.309}{0} & \AppScore{0.596}{0} & \AppScore{0.351}{0} & \AppScore{0.592}{0} & \AppScore{0.329}{0} \\
 &  & \textbf{+MORA} & \AppScore{0.513}{1} & \AppScore{0.298}{1} & \AppScore{0.524}{1} & \AppScore{0.324}{1} & \AppScore{0.538}{1} & \AppScore{0.313}{1} \\
 &  & \cellcolor{MoraTint}Imp.\ (\%) & \cellcolor{MoraTint}\AppImp{9.313} & \cellcolor{MoraTint}\AppImp{3.519} & \cellcolor{MoraTint}\AppImp{12.173} & \cellcolor{MoraTint}\AppImp{7.749} & \cellcolor{MoraTint}\AppImp{9.023} & \cellcolor{MoraTint}\AppImp{5.011} \\
\cmidrule[0.25pt]{2-9}
 & \multirow{3}{*}{720} & Base & \AppScore{0.606}{0} & \AppScore{0.331}{0} & \AppScore{0.635}{0} & \AppScore{0.372}{0} & \AppScore{0.628}{0} & \AppScore{0.346}{0} \\
 &  & \textbf{+MORA} & \AppScore{0.569}{1} & \AppScore{0.324}{1} & \AppScore{0.585}{1} & \AppScore{0.354}{1} & \AppScore{0.594}{1} & \AppScore{0.337}{1} \\
 &  & \cellcolor{MoraTint}Imp.\ (\%) & \cellcolor{MoraTint}\AppImp{6.057} & \cellcolor{MoraTint}\AppImp{2.142} & \cellcolor{MoraTint}\AppImp{7.832} & \cellcolor{MoraTint}\AppImp{4.801} & \cellcolor{MoraTint}\AppImp{5.372} & \cellcolor{MoraTint}\AppImp{2.600} \\
\midrule[0.45pt]
\multirow{12}{*}{\rotatebox[origin=c]{90}{\textbf{Solar}}} & \multirow{3}{*}{96} & Base & \AppScore{0.640}{0} & \AppScore{0.343}{0} & \AppScore{0.625}{0} & \AppScore{0.502}{0} & \AppScore{0.875}{0} & \AppScore{0.503}{0} \\
 &  & \textbf{+MORA} & \AppScore{0.186}{1} & \AppScore{0.265}{1} & \AppScore{0.177}{1} & \AppScore{0.254}{1} & \AppScore{0.184}{1} & \AppScore{0.254}{1} \\
 &  & \cellcolor{MoraTint}Imp.\ (\%) & \cellcolor{MoraTint}\AppImp{70.987} & \cellcolor{MoraTint}\AppImp{22.878} & \cellcolor{MoraTint}\AppImp{71.722} & \cellcolor{MoraTint}\AppImp{49.450} & \cellcolor{MoraTint}\AppImp{78.941} & \cellcolor{MoraTint}\AppImp{49.516} \\
\cmidrule[0.25pt]{2-9}
 & \multirow{3}{*}{192} & Base & \AppScore{0.527}{0} & \AppScore{0.324}{0} & \AppScore{0.680}{0} & \AppScore{0.534}{0} & \AppScore{0.937}{0} & \AppScore{0.528}{0} \\
 &  & \textbf{+MORA} & \AppScore{0.208}{1} & \AppScore{0.285}{1} & \AppScore{0.201}{1} & \AppScore{0.272}{1} & \AppScore{0.211}{1} & \AppScore{0.274}{1} \\
 &  & \cellcolor{MoraTint}Imp.\ (\%) & \cellcolor{MoraTint}\AppImp{60.569} & \cellcolor{MoraTint}\AppImp{12.318} & \cellcolor{MoraTint}\AppImp{70.495} & \cellcolor{MoraTint}\AppImp{49.014} & \cellcolor{MoraTint}\AppImp{77.473} & \cellcolor{MoraTint}\AppImp{48.213} \\
\cmidrule[0.25pt]{2-9}
 & \multirow{3}{*}{336} & Base & \AppScore{0.494}{0} & \AppScore{0.323}{0} & \AppScore{0.684}{0} & \AppScore{0.537}{0} & \AppScore{0.909}{0} & \AppScore{0.520}{0} \\
 &  & \textbf{+MORA} & \AppScore{0.221}{1} & \AppScore{0.289}{1} & \AppScore{0.222}{1} & \AppScore{0.285}{1} & \AppScore{0.234}{1} & \AppScore{0.287}{1} \\
 &  & \cellcolor{MoraTint}Imp.\ (\%) & \cellcolor{MoraTint}\AppImp{55.344} & \cellcolor{MoraTint}\AppImp{10.342} & \cellcolor{MoraTint}\AppImp{67.609} & \cellcolor{MoraTint}\AppImp{46.916} & \cellcolor{MoraTint}\AppImp{74.258} & \cellcolor{MoraTint}\AppImp{44.733} \\
\cmidrule[0.25pt]{2-9}
 & \multirow{3}{*}{720} & Base & \AppScore{0.451}{0} & \AppScore{0.319}{0} & \AppScore{0.676}{0} & \AppScore{0.545}{0} & \AppScore{0.930}{0} & \AppScore{0.528}{0} \\
 &  & \textbf{+MORA} & \AppScore{0.246}{1} & \AppScore{0.307}{1} & \AppScore{0.243}{1} & \AppScore{0.301}{1} & \AppScore{0.258}{1} & \AppScore{0.303}{1} \\
 &  & \cellcolor{MoraTint}Imp.\ (\%) & \cellcolor{MoraTint}\AppImp{45.493} & \cellcolor{MoraTint}\AppImp{3.786} & \cellcolor{MoraTint}\AppImp{64.109} & \cellcolor{MoraTint}\AppImp{44.763} & \cellcolor{MoraTint}\AppImp{72.301} & \cellcolor{MoraTint}\AppImp{42.712} \\
\bottomrule[0.8pt]
\end{tabular}
\end{table}

\clearpage

\subsection{Foundation Models at \texorpdfstring{$L=512$}{L=512}}
\label{app:foundation_results_512}
\begin{table}[H]
\centering
\caption{Complete long-horizon results for frozen foundation models at $L=512$.}
\label{tab:complete_foundation512}
\label{tab:full_foundation_512:etth1}
\label{tab:full_foundation_512:etth2}
\label{tab:full_foundation_512:ettm1}
\label{tab:full_foundation_512:ettm2}
\label{tab:full_foundation_512:weather}
\label{tab:full_foundation_512:ecl}
\label{tab:full_foundation_512:traffic}
\label{tab:full_foundation_512:solar}
\fontsize{8}{8.6}\selectfont
\setlength{\tabcolsep}{1pt}
\renewcommand{\arraystretch}{1.00}
\setlength{\aboverulesep}{0.6pt}
\setlength{\belowrulesep}{0.6pt}
\begin{tabular}{@{}>{\raggedright\arraybackslash}p{30pt}>{\raggedright\arraybackslash}p{14pt}>{\raggedright\arraybackslash}p{34pt}>{\raggedleft\arraybackslash}p{\dimexpr(\linewidth-78pt-16\tabcolsep)/6\relax}>{\raggedleft\arraybackslash}p{\dimexpr(\linewidth-78pt-16\tabcolsep)/6\relax}>{\raggedleft\arraybackslash}p{\dimexpr(\linewidth-78pt-16\tabcolsep)/6\relax}>{\raggedleft\arraybackslash}p{\dimexpr(\linewidth-78pt-16\tabcolsep)/6\relax}>{\raggedleft\arraybackslash}p{\dimexpr(\linewidth-78pt-16\tabcolsep)/6\relax}>{\raggedleft\arraybackslash}p{\dimexpr(\linewidth-78pt-16\tabcolsep)/6\relax}@{}}
\toprule[0.8pt]
\textbf{Dataset} & $H$ & Forecast & \multicolumn{2}{c}{\textbf{Chronos-2}} & \multicolumn{2}{c}{\textbf{Sundial}} & \multicolumn{2}{c}{\textbf{TimesFM 2.5}} \\
\cmidrule(lr){4-5}\cmidrule(lr){6-7}\cmidrule(lr){8-9}
 & & & MSE & MAE & MSE & MAE & MSE & MAE \\
\midrule
\multirow{12}{*}{\rotatebox[origin=c]{90}{\textbf{ETTh1}}} & \multirow{3}{*}{96} & Base & \AppScore{0.371}{0} & \AppScore{0.375}{1} & \AppScore{0.356}{0} & \AppScore{0.383}{0} & \AppScore{0.384}{0} & \AppScore{0.382}{0} \\
 &  & \textbf{+MORA} & \AppScore{0.350}{1} & \AppScore{0.376}{0} & \AppScore{0.343}{1} & \AppScore{0.378}{1} & \AppScore{0.354}{1} & \AppScore{0.380}{1} \\
 &  & \cellcolor{MoraTint}Imp.\ (\%) & \cellcolor{MoraTint}\AppImp{5.652} & \cellcolor{MoraTint}\AppImp{-0.287} & \cellcolor{MoraTint}\AppImp{3.593} & \cellcolor{MoraTint}\AppImp{1.318} & \cellcolor{MoraTint}\AppImp{7.781} & \cellcolor{MoraTint}\AppImp{0.545} \\
\cmidrule[0.25pt]{2-9}
 & \multirow{3}{*}{192} & Base & \AppScore{0.414}{0} & \AppScore{0.406}{0} & \AppScore{0.390}{0} & \AppScore{0.406}{0} & \AppScore{0.428}{0} & \AppScore{0.408}{1} \\
 &  & \textbf{+MORA} & \AppScore{0.392}{1} & \AppScore{0.406}{0} & \AppScore{0.378}{1} & \AppScore{0.401}{1} & \AppScore{0.401}{1} & \AppScore{0.409}{0} \\
 &  & \cellcolor{MoraTint}Imp.\ (\%) & \cellcolor{MoraTint}\AppImp{5.260} & \cellcolor{MoraTint}\AppImp{-0.093} & \cellcolor{MoraTint}\AppImp{3.025} & \cellcolor{MoraTint}\AppImp{1.106} & \cellcolor{MoraTint}\AppImp{6.117} & \cellcolor{MoraTint}\AppImp{-0.190} \\
\cmidrule[0.25pt]{2-9}
 & \multirow{3}{*}{336} & Base & \AppScore{0.439}{0} & \AppScore{0.423}{1} & \AppScore{0.403}{0} & \AppScore{0.415}{0} & \AppScore{0.446}{0} & \AppScore{0.419}{1} \\
 &  & \textbf{+MORA} & \AppScore{0.421}{1} & \AppScore{0.427}{0} & \AppScore{0.396}{1} & \AppScore{0.414}{1} & \AppScore{0.427}{1} & \AppScore{0.424}{0} \\
 &  & \cellcolor{MoraTint}Imp.\ (\%) & \cellcolor{MoraTint}\AppImp{4.179} & \cellcolor{MoraTint}\AppImp{-0.906} & \cellcolor{MoraTint}\AppImp{1.857} & \cellcolor{MoraTint}\AppImp{0.266} & \cellcolor{MoraTint}\AppImp{4.340} & \cellcolor{MoraTint}\AppImp{-1.193} \\
\cmidrule[0.25pt]{2-9}
 & \multirow{3}{*}{720} & Base & \AppScore{0.449}{0} & \AppScore{0.445}{1} & \AppScore{0.438}{0} & \AppScore{0.444}{0} & \AppScore{0.428}{0} & \AppScore{0.427}{1} \\
 &  & \textbf{+MORA} & \AppScore{0.445}{1} & \AppScore{0.452}{0} & \AppScore{0.431}{1} & \AppScore{0.443}{1} & \AppScore{0.428}{0} & \AppScore{0.437}{0} \\
 &  & \cellcolor{MoraTint}Imp.\ (\%) & \cellcolor{MoraTint}\AppImp{0.925} & \cellcolor{MoraTint}\AppImp{-1.602} & \cellcolor{MoraTint}\AppImp{1.603} & \cellcolor{MoraTint}\AppImp{0.134} & \cellcolor{MoraTint}\AppImp{-0.148} & \cellcolor{MoraTint}\AppImp{-2.455} \\
\midrule[0.45pt]
\multirow{12}{*}{\rotatebox[origin=c]{90}{\textbf{ETTh2}}} & \multirow{3}{*}{96} & Base & \AppScore{0.280}{0} & \AppScore{0.320}{0} & \AppScore{0.279}{0} & \AppScore{0.334}{0} & \AppScore{0.306}{0} & \AppScore{0.338}{0} \\
 &  & \textbf{+MORA} & \AppScore{0.273}{1} & \AppScore{0.316}{1} & \AppScore{0.271}{1} & \AppScore{0.325}{1} & \AppScore{0.295}{1} & \AppScore{0.332}{1} \\
 &  & \cellcolor{MoraTint}Imp.\ (\%) & \cellcolor{MoraTint}\AppImp{2.535} & \cellcolor{MoraTint}\AppImp{1.068} & \cellcolor{MoraTint}\AppImp{2.938} & \cellcolor{MoraTint}\AppImp{2.550} & \cellcolor{MoraTint}\AppImp{3.730} & \cellcolor{MoraTint}\AppImp{1.796} \\
\cmidrule[0.25pt]{2-9}
 & \multirow{3}{*}{192} & Base & \AppScore{0.351}{0} & \AppScore{0.369}{0} & \AppScore{0.342}{0} & \AppScore{0.376}{0} & \AppScore{0.376}{0} & \AppScore{0.384}{0} \\
 &  & \textbf{+MORA} & \AppScore{0.342}{1} & \AppScore{0.364}{1} & \AppScore{0.338}{1} & \AppScore{0.371}{1} & \AppScore{0.365}{1} & \AppScore{0.379}{1} \\
 &  & \cellcolor{MoraTint}Imp.\ (\%) & \cellcolor{MoraTint}\AppImp{2.717} & \cellcolor{MoraTint}\AppImp{1.316} & \cellcolor{MoraTint}\AppImp{1.211} & \cellcolor{MoraTint}\AppImp{1.468} & \cellcolor{MoraTint}\AppImp{2.858} & \cellcolor{MoraTint}\AppImp{1.444} \\
\cmidrule[0.25pt]{2-9}
 & \multirow{3}{*}{336} & Base & \AppScore{0.377}{0} & \AppScore{0.397}{0} & \AppScore{0.361}{1} & \AppScore{0.395}{1} & \AppScore{0.401}{0} & \AppScore{0.409}{0} \\
 &  & \textbf{+MORA} & \AppScore{0.371}{1} & \AppScore{0.395}{1} & \AppScore{0.367}{0} & \AppScore{0.397}{0} & \AppScore{0.394}{1} & \AppScore{0.405}{1} \\
 &  & \cellcolor{MoraTint}Imp.\ (\%) & \cellcolor{MoraTint}\AppImp{1.726} & \cellcolor{MoraTint}\AppImp{0.402} & \cellcolor{MoraTint}\AppImp{-1.778} & \cellcolor{MoraTint}\AppImp{-0.482} & \cellcolor{MoraTint}\AppImp{1.713} & \cellcolor{MoraTint}\AppImp{0.875} \\
\cmidrule[0.25pt]{2-9}
 & \multirow{3}{*}{720} & Base & \AppScore{0.398}{0} & \AppScore{0.424}{1} & \AppScore{0.392}{0} & \AppScore{0.426}{0} & \AppScore{0.405}{1} & \AppScore{0.422}{0} \\
 &  & \textbf{+MORA} & \AppScore{0.393}{1} & \AppScore{0.434}{0} & \AppScore{0.384}{1} & \AppScore{0.420}{1} & \AppScore{0.406}{0} & \AppScore{0.418}{1} \\
 &  & \cellcolor{MoraTint}Imp.\ (\%) & \cellcolor{MoraTint}\AppImp{1.362} & \cellcolor{MoraTint}\AppImp{-2.310} & \cellcolor{MoraTint}\AppImp{1.997} & \cellcolor{MoraTint}\AppImp{1.358} & \cellcolor{MoraTint}\AppImp{-0.195} & \cellcolor{MoraTint}\AppImp{0.756} \\
\midrule[0.45pt]
\multirow{12}{*}{\rotatebox[origin=c]{90}{\textbf{ETTm1}}} & \multirow{3}{*}{96} & Base & \AppScore{0.308}{0} & \AppScore{0.323}{0} & \AppScore{0.292}{0} & \AppScore{0.325}{0} & \AppScore{0.313}{0} & \AppScore{0.329}{0} \\
 &  & \textbf{+MORA} & \AppScore{0.277}{1} & \AppScore{0.315}{1} & \AppScore{0.275}{1} & \AppScore{0.323}{1} & \AppScore{0.284}{1} & \AppScore{0.326}{1} \\
 &  & \cellcolor{MoraTint}Imp.\ (\%) & \cellcolor{MoraTint}\AppImp{10.024} & \cellcolor{MoraTint}\AppImp{2.463} & \cellcolor{MoraTint}\AppImp{5.920} & \cellcolor{MoraTint}\AppImp{0.676} & \cellcolor{MoraTint}\AppImp{9.169} & \cellcolor{MoraTint}\AppImp{1.008} \\
\cmidrule[0.25pt]{2-9}
 & \multirow{3}{*}{192} & Base & \AppScore{0.362}{0} & \AppScore{0.355}{0} & \AppScore{0.342}{0} & \AppScore{0.357}{0} & \AppScore{0.361}{0} & \AppScore{0.359}{0} \\
 &  & \textbf{+MORA} & \AppScore{0.318}{1} & \AppScore{0.345}{1} & \AppScore{0.318}{1} & \AppScore{0.353}{1} & \AppScore{0.321}{1} & \AppScore{0.353}{1} \\
 &  & \cellcolor{MoraTint}Imp.\ (\%) & \cellcolor{MoraTint}\AppImp{12.170} & \cellcolor{MoraTint}\AppImp{2.873} & \cellcolor{MoraTint}\AppImp{7.018} & \cellcolor{MoraTint}\AppImp{1.176} & \cellcolor{MoraTint}\AppImp{10.897} & \cellcolor{MoraTint}\AppImp{1.813} \\
\cmidrule[0.25pt]{2-9}
 & \multirow{3}{*}{336} & Base & \AppScore{0.395}{0} & \AppScore{0.381}{0} & \AppScore{0.378}{0} & \AppScore{0.380}{0} & \AppScore{0.394}{0} & \AppScore{0.382}{0} \\
 &  & \textbf{+MORA} & \AppScore{0.345}{1} & \AppScore{0.370}{1} & \AppScore{0.350}{1} & \AppScore{0.375}{1} & \AppScore{0.351}{1} & \AppScore{0.374}{1} \\
 &  & \cellcolor{MoraTint}Imp.\ (\%) & \cellcolor{MoraTint}\AppImp{12.803} & \cellcolor{MoraTint}\AppImp{2.734} & \cellcolor{MoraTint}\AppImp{7.557} & \cellcolor{MoraTint}\AppImp{1.434} & \cellcolor{MoraTint}\AppImp{10.820} & \cellcolor{MoraTint}\AppImp{2.128} \\
\cmidrule[0.25pt]{2-9}
 & \multirow{3}{*}{720} & Base & \AppScore{0.484}{0} & \AppScore{0.430}{0} & \AppScore{0.451}{0} & \AppScore{0.418}{0} & \AppScore{0.456}{0} & \AppScore{0.418}{0} \\
 &  & \textbf{+MORA} & \AppScore{0.411}{1} & \AppScore{0.412}{1} & \AppScore{0.411}{1} & \AppScore{0.409}{1} & \AppScore{0.409}{1} & \AppScore{0.411}{1} \\
 &  & \cellcolor{MoraTint}Imp.\ (\%) & \cellcolor{MoraTint}\AppImp{14.997} & \cellcolor{MoraTint}\AppImp{4.306} & \cellcolor{MoraTint}\AppImp{8.910} & \cellcolor{MoraTint}\AppImp{2.086} & \cellcolor{MoraTint}\AppImp{10.337} & \cellcolor{MoraTint}\AppImp{1.501} \\
\midrule[0.45pt]
\multirow{12}{*}{\rotatebox[origin=c]{90}{\textbf{ETTm2}}} & \multirow{3}{*}{96} & Base & \AppScore{0.170}{0} & \AppScore{0.237}{0} & \AppScore{0.176}{0} & \AppScore{0.255}{0} & \AppScore{0.179}{0} & \AppScore{0.246}{0} \\
 &  & \textbf{+MORA} & \AppScore{0.161}{1} & \AppScore{0.233}{1} & \AppScore{0.158}{1} & \AppScore{0.244}{1} & \AppScore{0.168}{1} & \AppScore{0.241}{1} \\
 &  & \cellcolor{MoraTint}Imp.\ (\%) & \cellcolor{MoraTint}\AppImp{5.504} & \cellcolor{MoraTint}\AppImp{1.841} & \cellcolor{MoraTint}\AppImp{10.200} & \cellcolor{MoraTint}\AppImp{4.244} & \cellcolor{MoraTint}\AppImp{6.448} & \cellcolor{MoraTint}\AppImp{2.062} \\
\cmidrule[0.25pt]{2-9}
 & \multirow{3}{*}{192} & Base & \AppScore{0.241}{0} & \AppScore{0.286}{0} & \AppScore{0.239}{0} & \AppScore{0.298}{0} & \AppScore{0.249}{0} & \AppScore{0.293}{0} \\
 &  & \textbf{+MORA} & \AppScore{0.224}{1} & \AppScore{0.277}{1} & \AppScore{0.216}{1} & \AppScore{0.285}{1} & \AppScore{0.231}{1} & \AppScore{0.285}{1} \\
 &  & \cellcolor{MoraTint}Imp.\ (\%) & \cellcolor{MoraTint}\AppImp{6.835} & \cellcolor{MoraTint}\AppImp{2.823} & \cellcolor{MoraTint}\AppImp{9.776} & \cellcolor{MoraTint}\AppImp{4.402} & \cellcolor{MoraTint}\AppImp{7.006} & \cellcolor{MoraTint}\AppImp{2.958} \\
\cmidrule[0.25pt]{2-9}
 & \multirow{3}{*}{336} & Base & \AppScore{0.306}{0} & \AppScore{0.329}{0} & \AppScore{0.297}{0} & \AppScore{0.336}{0} & \AppScore{0.314}{0} & \AppScore{0.336}{0} \\
 &  & \textbf{+MORA} & \AppScore{0.277}{1} & \AppScore{0.315}{1} & \AppScore{0.272}{1} & \AppScore{0.323}{1} & \AppScore{0.294}{1} & \AppScore{0.326}{1} \\
 &  & \cellcolor{MoraTint}Imp.\ (\%) & \cellcolor{MoraTint}\AppImp{9.327} & \cellcolor{MoraTint}\AppImp{4.403} & \cellcolor{MoraTint}\AppImp{8.294} & \cellcolor{MoraTint}\AppImp{3.623} & \cellcolor{MoraTint}\AppImp{6.455} & \cellcolor{MoraTint}\AppImp{2.956} \\
\cmidrule[0.25pt]{2-9}
 & \multirow{3}{*}{720} & Base & \AppScore{0.404}{0} & \AppScore{0.390}{0} & \AppScore{0.387}{0} & \AppScore{0.393}{0} & \AppScore{0.420}{0} & \AppScore{0.396}{0} \\
 &  & \textbf{+MORA} & \AppScore{0.366}{1} & \AppScore{0.373}{1} & \AppScore{0.364}{1} & \AppScore{0.381}{1} & \AppScore{0.399}{1} & \AppScore{0.387}{1} \\
 &  & \cellcolor{MoraTint}Imp.\ (\%) & \cellcolor{MoraTint}\AppImp{9.249} & \cellcolor{MoraTint}\AppImp{4.478} & \cellcolor{MoraTint}\AppImp{6.138} & \cellcolor{MoraTint}\AppImp{3.056} & \cellcolor{MoraTint}\AppImp{5.089} & \cellcolor{MoraTint}\AppImp{2.445} \\
\bottomrule[0.8pt]
\end{tabular}
\end{table}

\clearpage
\begin{table}[H]
\centering
\parbox{\linewidth}{\small Table~\ref{tab:complete_foundation512} (continued, part 2 of 2). Complete long-horizon results for frozen foundation models at $L=512$.}\par\vspace{3pt}
\fontsize{8}{8.6}\selectfont
\setlength{\tabcolsep}{1pt}
\renewcommand{\arraystretch}{1.00}
\setlength{\aboverulesep}{0.6pt}
\setlength{\belowrulesep}{0.6pt}
\begin{tabular}{@{}>{\raggedright\arraybackslash}p{30pt}>{\raggedright\arraybackslash}p{14pt}>{\raggedright\arraybackslash}p{34pt}>{\raggedleft\arraybackslash}p{\dimexpr(\linewidth-78pt-16\tabcolsep)/6\relax}>{\raggedleft\arraybackslash}p{\dimexpr(\linewidth-78pt-16\tabcolsep)/6\relax}>{\raggedleft\arraybackslash}p{\dimexpr(\linewidth-78pt-16\tabcolsep)/6\relax}>{\raggedleft\arraybackslash}p{\dimexpr(\linewidth-78pt-16\tabcolsep)/6\relax}>{\raggedleft\arraybackslash}p{\dimexpr(\linewidth-78pt-16\tabcolsep)/6\relax}>{\raggedleft\arraybackslash}p{\dimexpr(\linewidth-78pt-16\tabcolsep)/6\relax}@{}}
\toprule[0.8pt]
\textbf{Dataset} & $H$ & Forecast & \multicolumn{2}{c}{\textbf{Chronos-2}} & \multicolumn{2}{c}{\textbf{Sundial}} & \multicolumn{2}{c}{\textbf{TimesFM 2.5}} \\
\cmidrule(lr){4-5}\cmidrule(lr){6-7}\cmidrule(lr){8-9}
 & & & MSE & MAE & MSE & MAE & MSE & MAE \\
\midrule
\multirow{12}{*}{\rotatebox[origin=c]{90}{\textbf{Weather}}} & \multirow{3}{*}{96} & Base & \AppScore{0.156}{0} & \AppScore{0.176}{1} & \AppScore{0.160}{0} & \AppScore{0.207}{0} & \AppScore{0.163}{0} & \AppScore{0.194}{0} \\
 &  & \textbf{+MORA} & \AppScore{0.139}{1} & \AppScore{0.178}{0} & \AppScore{0.139}{1} & \AppScore{0.195}{1} & \AppScore{0.142}{1} & \AppScore{0.186}{1} \\
 &  & \cellcolor{MoraTint}Imp.\ (\%) & \cellcolor{MoraTint}\AppImp{10.563} & \cellcolor{MoraTint}\AppImp{-1.281} & \cellcolor{MoraTint}\AppImp{13.187} & \cellcolor{MoraTint}\AppImp{5.588} & \cellcolor{MoraTint}\AppImp{12.716} & \cellcolor{MoraTint}\AppImp{4.399} \\
\cmidrule[0.25pt]{2-9}
 & \multirow{3}{*}{192} & Base & \AppScore{0.211}{0} & \AppScore{0.226}{0} & \AppScore{0.213}{0} & \AppScore{0.255}{0} & \AppScore{0.216}{0} & \AppScore{0.242}{0} \\
 &  & \textbf{+MORA} & \AppScore{0.181}{1} & \AppScore{0.225}{1} & \AppScore{0.182}{1} & \AppScore{0.240}{1} & \AppScore{0.184}{1} & \AppScore{0.227}{1} \\
 &  & \cellcolor{MoraTint}Imp.\ (\%) & \cellcolor{MoraTint}\AppImp{14.208} & \cellcolor{MoraTint}\AppImp{0.505} & \cellcolor{MoraTint}\AppImp{14.873} & \cellcolor{MoraTint}\AppImp{6.002} & \cellcolor{MoraTint}\AppImp{14.831} & \cellcolor{MoraTint}\AppImp{6.405} \\
\cmidrule[0.25pt]{2-9}
 & \multirow{3}{*}{336} & Base & \AppScore{0.281}{0} & \AppScore{0.275}{0} & \AppScore{0.267}{0} & \AppScore{0.295}{0} & \AppScore{0.289}{0} & \AppScore{0.289}{0} \\
 &  & \textbf{+MORA} & \AppScore{0.229}{1} & \AppScore{0.269}{1} & \AppScore{0.231}{1} & \AppScore{0.280}{1} & \AppScore{0.232}{1} & \AppScore{0.267}{1} \\
 &  & \cellcolor{MoraTint}Imp.\ (\%) & \cellcolor{MoraTint}\AppImp{18.699} & \cellcolor{MoraTint}\AppImp{2.069} & \cellcolor{MoraTint}\AppImp{13.580} & \cellcolor{MoraTint}\AppImp{5.082} & \cellcolor{MoraTint}\AppImp{19.744} & \cellcolor{MoraTint}\AppImp{7.461} \\
\cmidrule[0.25pt]{2-9}
 & \multirow{3}{*}{720} & Base & \AppScore{0.383}{0} & \AppScore{0.332}{0} & \AppScore{0.336}{0} & \AppScore{0.344}{0} & \AppScore{0.385}{0} & \AppScore{0.345}{0} \\
 &  & \textbf{+MORA} & \AppScore{0.302}{1} & \AppScore{0.329}{1} & \AppScore{0.300}{1} & \AppScore{0.332}{1} & \AppScore{0.302}{1} & \AppScore{0.318}{1} \\
 &  & \cellcolor{MoraTint}Imp.\ (\%) & \cellcolor{MoraTint}\AppImp{21.168} & \cellcolor{MoraTint}\AppImp{0.937} & \cellcolor{MoraTint}\AppImp{10.945} & \cellcolor{MoraTint}\AppImp{3.359} & \cellcolor{MoraTint}\AppImp{21.631} & \cellcolor{MoraTint}\AppImp{7.811} \\
\midrule[0.45pt]
\multirow{12}{*}{\rotatebox[origin=c]{90}{\textbf{ECL}}} & \multirow{3}{*}{96} & Base & \AppScore{0.128}{0} & \AppScore{0.211}{0} & \AppScore{0.118}{0} & \AppScore{0.213}{0} & \AppScore{0.133}{0} & \AppScore{0.220}{0} \\
 &  & \textbf{+MORA} & \AppScore{0.123}{1} & \AppScore{0.210}{1} & \AppScore{0.116}{1} & \AppScore{0.212}{1} & \AppScore{0.127}{1} & \AppScore{0.218}{1} \\
 &  & \cellcolor{MoraTint}Imp.\ (\%) & \cellcolor{MoraTint}\AppImp{3.483} & \cellcolor{MoraTint}\AppImp{0.394} & \cellcolor{MoraTint}\AppImp{1.657} & \cellcolor{MoraTint}\AppImp{0.588} & \cellcolor{MoraTint}\AppImp{4.320} & \cellcolor{MoraTint}\AppImp{0.810} \\
\cmidrule[0.25pt]{2-9}
 & \multirow{3}{*}{192} & Base & \AppScore{0.146}{0} & \AppScore{0.229}{0} & \AppScore{0.135}{0} & \AppScore{0.231}{0} & \AppScore{0.150}{0} & \AppScore{0.236}{0} \\
 &  & \textbf{+MORA} & \AppScore{0.141}{1} & \AppScore{0.228}{1} & \AppScore{0.132}{1} & \AppScore{0.229}{1} & \AppScore{0.143}{1} & \AppScore{0.234}{1} \\
 &  & \cellcolor{MoraTint}Imp.\ (\%) & \cellcolor{MoraTint}\AppImp{3.732} & \cellcolor{MoraTint}\AppImp{0.538} & \cellcolor{MoraTint}\AppImp{2.126} & \cellcolor{MoraTint}\AppImp{0.754} & \cellcolor{MoraTint}\AppImp{4.480} & \cellcolor{MoraTint}\AppImp{1.014} \\
\cmidrule[0.25pt]{2-9}
 & \multirow{3}{*}{336} & Base & \AppScore{0.167}{0} & \AppScore{0.249}{0} & \AppScore{0.150}{0} & \AppScore{0.247}{0} & \AppScore{0.167}{0} & \AppScore{0.253}{0} \\
 &  & \textbf{+MORA} & \AppScore{0.158}{1} & \AppScore{0.247}{1} & \AppScore{0.146}{1} & \AppScore{0.245}{1} & \AppScore{0.159}{1} & \AppScore{0.251}{1} \\
 &  & \cellcolor{MoraTint}Imp.\ (\%) & \cellcolor{MoraTint}\AppImp{5.013} & \cellcolor{MoraTint}\AppImp{0.973} & \cellcolor{MoraTint}\AppImp{2.865} & \cellcolor{MoraTint}\AppImp{0.979} & \cellcolor{MoraTint}\AppImp{5.096} & \cellcolor{MoraTint}\AppImp{1.169} \\
\cmidrule[0.25pt]{2-9}
 & \multirow{3}{*}{720} & Base & \AppScore{0.210}{0} & \AppScore{0.287}{0} & \AppScore{0.188}{0} & \AppScore{0.280}{0} & \AppScore{0.206}{0} & \AppScore{0.287}{0} \\
 &  & \textbf{+MORA} & \AppScore{0.196}{1} & \AppScore{0.281}{1} & \AppScore{0.180}{1} & \AppScore{0.276}{1} & \AppScore{0.195}{1} & \AppScore{0.283}{1} \\
 &  & \cellcolor{MoraTint}Imp.\ (\%) & \cellcolor{MoraTint}\AppImp{6.869} & \cellcolor{MoraTint}\AppImp{1.912} & \cellcolor{MoraTint}\AppImp{4.257} & \cellcolor{MoraTint}\AppImp{1.461} & \cellcolor{MoraTint}\AppImp{5.696} & \cellcolor{MoraTint}\AppImp{1.545} \\
\midrule[0.45pt]
\multirow{12}{*}{\rotatebox[origin=c]{90}{\textbf{Traffic}}} & \multirow{3}{*}{96} & Base & \AppScore{0.353}{0} & \AppScore{0.219}{0} & \AppScore{0.425}{0} & \AppScore{0.270}{0} & \AppScore{0.355}{0} & \AppScore{0.231}{0} \\
 &  & \textbf{+MORA} & \AppScore{0.349}{1} & \AppScore{0.219}{0} & \AppScore{0.408}{1} & \AppScore{0.265}{1} & \AppScore{0.352}{1} & \AppScore{0.230}{1} \\
 &  & \cellcolor{MoraTint}Imp.\ (\%) & \cellcolor{MoraTint}\AppImp{0.965} & \cellcolor{MoraTint}\AppImp{-0.212} & \cellcolor{MoraTint}\AppImp{3.920} & \cellcolor{MoraTint}\AppImp{1.923} & \cellcolor{MoraTint}\AppImp{0.873} & \cellcolor{MoraTint}\AppImp{0.149} \\
\cmidrule[0.25pt]{2-9}
 & \multirow{3}{*}{192} & Base & \AppScore{0.379}{0} & \AppScore{0.231}{0} & \AppScore{0.448}{0} & \AppScore{0.283}{0} & \AppScore{0.371}{0} & \AppScore{0.238}{0} \\
 &  & \textbf{+MORA} & \AppScore{0.375}{1} & \AppScore{0.231}{0} & \AppScore{0.428}{1} & \AppScore{0.276}{1} & \AppScore{0.368}{1} & \AppScore{0.238}{0} \\
 &  & \cellcolor{MoraTint}Imp.\ (\%) & \cellcolor{MoraTint}\AppImp{1.106} & \cellcolor{MoraTint}\AppImp{-0.106} & \cellcolor{MoraTint}\AppImp{4.341} & \cellcolor{MoraTint}\AppImp{2.471} & \cellcolor{MoraTint}\AppImp{0.903} & \cellcolor{MoraTint}\AppImp{0.155} \\
\cmidrule[0.25pt]{2-9}
 & \multirow{3}{*}{336} & Base & \AppScore{0.396}{0} & \AppScore{0.240}{0} & \AppScore{0.462}{0} & \AppScore{0.292}{0} & \AppScore{0.382}{0} & \AppScore{0.244}{0} \\
 &  & \textbf{+MORA} & \AppScore{0.392}{1} & \AppScore{0.240}{0} & \AppScore{0.444}{1} & \AppScore{0.285}{1} & \AppScore{0.379}{1} & \AppScore{0.244}{0} \\
 &  & \cellcolor{MoraTint}Imp.\ (\%) & \cellcolor{MoraTint}\AppImp{1.006} & \cellcolor{MoraTint}\AppImp{-0.158} & \cellcolor{MoraTint}\AppImp{3.944} & \cellcolor{MoraTint}\AppImp{2.350} & \cellcolor{MoraTint}\AppImp{0.730} & \cellcolor{MoraTint}\AppImp{-0.022} \\
\cmidrule[0.25pt]{2-9}
 & \multirow{3}{*}{720} & Base & \AppScore{0.442}{0} & \AppScore{0.264}{1} & \AppScore{0.499}{0} & \AppScore{0.314}{0} & \AppScore{0.418}{0} & \AppScore{0.264}{0} \\
 &  & \textbf{+MORA} & \AppScore{0.440}{1} & \AppScore{0.265}{0} & \AppScore{0.485}{1} & \AppScore{0.308}{1} & \AppScore{0.416}{1} & \AppScore{0.264}{0} \\
 &  & \cellcolor{MoraTint}Imp.\ (\%) & \cellcolor{MoraTint}\AppImp{0.384} & \cellcolor{MoraTint}\AppImp{-0.205} & \cellcolor{MoraTint}\AppImp{2.837} & \cellcolor{MoraTint}\AppImp{1.799} & \cellcolor{MoraTint}\AppImp{0.519} & \cellcolor{MoraTint}\AppImp{0.052} \\
\midrule[0.45pt]
\multirow{12}{*}{\rotatebox[origin=c]{90}{\textbf{Solar}}} & \multirow{3}{*}{96} & Base & \AppScore{0.157}{0} & \AppScore{0.172}{0} & \AppScore{0.212}{0} & \AppScore{0.234}{0} & \AppScore{0.236}{0} & \AppScore{0.221}{1} \\
 &  & \textbf{+MORA} & \AppScore{0.157}{0} & \AppScore{0.172}{0} & \AppScore{0.190}{1} & \AppScore{0.234}{0} & \AppScore{0.186}{1} & \AppScore{0.231}{0} \\
 &  & \cellcolor{MoraTint}Imp.\ (\%) & \cellcolor{MoraTint}\AppImp{0.011} & \cellcolor{MoraTint}\AppImp{-0.032} & \cellcolor{MoraTint}\AppImp{10.075} & \cellcolor{MoraTint}\AppImp{-0.266} & \cellcolor{MoraTint}\AppImp{20.996} & \cellcolor{MoraTint}\AppImp{-4.608} \\
\cmidrule[0.25pt]{2-9}
 & \multirow{3}{*}{192} & Base & \AppScore{0.180}{0} & \AppScore{0.190}{0} & \AppScore{0.229}{0} & \AppScore{0.253}{0} & \AppScore{0.270}{0} & \AppScore{0.242}{1} \\
 &  & \textbf{+MORA} & \AppScore{0.180}{0} & \AppScore{0.190}{0} & \AppScore{0.209}{1} & \AppScore{0.253}{0} & \AppScore{0.217}{1} & \AppScore{0.253}{0} \\
 &  & \cellcolor{MoraTint}Imp.\ (\%) & \cellcolor{MoraTint}\AppImp{0.069} & \cellcolor{MoraTint}\AppImp{-0.119} & \cellcolor{MoraTint}\AppImp{8.832} & \cellcolor{MoraTint}\AppImp{-0.024} & \cellcolor{MoraTint}\AppImp{19.555} & \cellcolor{MoraTint}\AppImp{-4.666} \\
\cmidrule[0.25pt]{2-9}
 & \multirow{3}{*}{336} & Base & \AppScore{0.197}{0} & \AppScore{0.202}{0} & \AppScore{0.234}{0} & \AppScore{0.264}{0} & \AppScore{0.303}{0} & \AppScore{0.262}{1} \\
 &  & \textbf{+MORA} & \AppScore{0.197}{0} & \AppScore{0.202}{0} & \AppScore{0.214}{1} & \AppScore{0.264}{0} & \AppScore{0.241}{1} & \AppScore{0.274}{0} \\
 &  & \cellcolor{MoraTint}Imp.\ (\%) & \cellcolor{MoraTint}\AppImp{0.148} & \cellcolor{MoraTint}\AppImp{-0.214} & \cellcolor{MoraTint}\AppImp{8.562} & \cellcolor{MoraTint}\AppImp{0.161} & \cellcolor{MoraTint}\AppImp{20.378} & \cellcolor{MoraTint}\AppImp{-4.620} \\
\cmidrule[0.25pt]{2-9}
 & \multirow{3}{*}{720} & Base & \AppScore{0.214}{0} & \AppScore{0.210}{1} & \AppScore{0.243}{0} & \AppScore{0.277}{0} & \AppScore{0.317}{0} & \AppScore{0.274}{1} \\
 &  & \textbf{+MORA} & \AppScore{0.213}{1} & \AppScore{0.211}{0} & \AppScore{0.220}{1} & \AppScore{0.275}{1} & \AppScore{0.242}{1} & \AppScore{0.285}{0} \\
 &  & \cellcolor{MoraTint}Imp.\ (\%) & \cellcolor{MoraTint}\AppImp{0.405} & \cellcolor{MoraTint}\AppImp{-0.384} & \cellcolor{MoraTint}\AppImp{9.443} & \cellcolor{MoraTint}\AppImp{0.722} & \cellcolor{MoraTint}\AppImp{23.618} & \cellcolor{MoraTint}\AppImp{-4.162} \\
\bottomrule[0.8pt]
\end{tabular}
\end{table}

\clearpage

\subsection{Short-Horizon Forecasting on PEMS}
\label{app:pems_results}
\begin{table}[H]
\centering
\caption{Complete short-horizon results on PEMS at $L=96$.}
\label{tab:complete_pems}
\label{tab:full_pems:pems03}
\label{tab:full_pems:pems04}
\label{tab:full_pems:pems07}
\label{tab:full_pems:pems08}
\fontsize{8}{8.6}\selectfont
\setlength{\tabcolsep}{1pt}
\renewcommand{\arraystretch}{1.00}
\setlength{\aboverulesep}{0.6pt}
\setlength{\belowrulesep}{0.6pt}
\begin{tabular}{@{}>{\raggedright\arraybackslash}p{30pt}>{\raggedright\arraybackslash}p{14pt}>{\raggedright\arraybackslash}p{34pt}>{\raggedleft\arraybackslash}p{\dimexpr(\linewidth-78pt-28\tabcolsep)/12\relax}>{\raggedleft\arraybackslash}p{\dimexpr(\linewidth-78pt-28\tabcolsep)/12\relax}>{\raggedleft\arraybackslash}p{\dimexpr(\linewidth-78pt-28\tabcolsep)/12\relax}>{\raggedleft\arraybackslash}p{\dimexpr(\linewidth-78pt-28\tabcolsep)/12\relax}>{\raggedleft\arraybackslash}p{\dimexpr(\linewidth-78pt-28\tabcolsep)/12\relax}>{\raggedleft\arraybackslash}p{\dimexpr(\linewidth-78pt-28\tabcolsep)/12\relax}>{\raggedleft\arraybackslash}p{\dimexpr(\linewidth-78pt-28\tabcolsep)/12\relax}>{\raggedleft\arraybackslash}p{\dimexpr(\linewidth-78pt-28\tabcolsep)/12\relax}>{\raggedleft\arraybackslash}p{\dimexpr(\linewidth-78pt-28\tabcolsep)/12\relax}>{\raggedleft\arraybackslash}p{\dimexpr(\linewidth-78pt-28\tabcolsep)/12\relax}>{\raggedleft\arraybackslash}p{\dimexpr(\linewidth-78pt-28\tabcolsep)/12\relax}>{\raggedleft\arraybackslash}p{\dimexpr(\linewidth-78pt-28\tabcolsep)/12\relax}@{}}
\toprule[0.8pt]
\textbf{Dataset} & $H$ & Forecast & \multicolumn{2}{c}{\textbf{TimeFilter}} & \multicolumn{2}{c}{\textbf{iTransformer}} & \multicolumn{2}{c}{\textbf{CGTFra}} & \multicolumn{2}{c}{\textbf{PatchTST}} & \multicolumn{2}{c}{\textbf{DLinear}} & \multicolumn{2}{c}{\textbf{VPNet}} \\
\cmidrule(lr){4-5}\cmidrule(lr){6-7}\cmidrule(lr){8-9}\cmidrule(lr){10-11}\cmidrule(lr){12-13}\cmidrule(lr){14-15}
 & & & MSE & MAE & MSE & MAE & MSE & MAE & MSE & MAE & MSE & MAE & MSE & MAE \\
\midrule
\multirow{12}{*}{\rotatebox[origin=c]{90}{\textbf{PEMS03}}} & \multirow{3}{*}{12} & Base & \AppScore{0.063}{0} & \AppScore{0.164}{0} & \AppScore{0.068}{1} & \AppScore{0.174}{0} & \AppScore{0.060}{0} & \AppScore{0.160}{0} & \AppScore{0.084}{0} & \AppScore{0.191}{0} & \AppScore{0.107}{0} & \AppScore{0.219}{0} & \AppScore{0.066}{0} & \AppScore{0.171}{0} \\
 &  & \textbf{+MORA} & \AppScore{0.062}{1} & \AppScore{0.164}{0} & \AppScore{0.069}{0} & \AppScore{0.172}{1} & \AppScore{0.059}{1} & \AppScore{0.159}{1} & \AppScore{0.075}{1} & \AppScore{0.182}{1} & \AppScore{0.083}{1} & \AppScore{0.193}{1} & \AppScore{0.066}{0} & \AppScore{0.171}{0} \\
 &  & \cellcolor{MoraTint}Imp.\ (\%) & \cellcolor{MoraTint}\AppImp{1.704} & \cellcolor{MoraTint}\AppImp{-0.059} & \cellcolor{MoraTint}\AppImp{-0.168} & \cellcolor{MoraTint}\AppImp{1.204} & \cellcolor{MoraTint}\AppImp{1.745} & \cellcolor{MoraTint}\AppImp{0.576} & \cellcolor{MoraTint}\AppImp{10.531} & \cellcolor{MoraTint}\AppImp{4.782} & \cellcolor{MoraTint}\AppImp{22.063} & \cellcolor{MoraTint}\AppImp{11.975} & \cellcolor{MoraTint}\AppImp{-0.578} & \cellcolor{MoraTint}\AppImp{-0.060} \\
\cmidrule[0.25pt]{2-15}
 & \multirow{3}{*}{24} & Base & \AppScore{0.079}{0} & \AppScore{0.186}{0} & \AppScore{0.095}{0} & \AppScore{0.204}{0} & \AppScore{0.079}{0} & \AppScore{0.184}{0} & \AppScore{0.130}{0} & \AppScore{0.239}{0} & \AppScore{0.184}{0} & \AppScore{0.290}{0} & \AppScore{0.090}{0} & \AppScore{0.202}{0} \\
 &  & \textbf{+MORA} & \AppScore{0.077}{1} & \AppScore{0.184}{1} & \AppScore{0.088}{1} & \AppScore{0.192}{1} & \AppScore{0.075}{1} & \AppScore{0.180}{1} & \AppScore{0.098}{1} & \AppScore{0.208}{1} & \AppScore{0.109}{1} & \AppScore{0.219}{1} & \AppScore{0.084}{1} & \AppScore{0.194}{1} \\
 &  & \cellcolor{MoraTint}Imp.\ (\%) & \cellcolor{MoraTint}\AppImp{3.062} & \cellcolor{MoraTint}\AppImp{1.023} & \cellcolor{MoraTint}\AppImp{7.299} & \cellcolor{MoraTint}\AppImp{5.700} & \cellcolor{MoraTint}\AppImp{4.641} & \cellcolor{MoraTint}\AppImp{2.126} & \cellcolor{MoraTint}\AppImp{24.083} & \cellcolor{MoraTint}\AppImp{12.862} & \cellcolor{MoraTint}\AppImp{40.671} & \cellcolor{MoraTint}\AppImp{24.421} & \cellcolor{MoraTint}\AppImp{6.581} & \cellcolor{MoraTint}\AppImp{4.129} \\
\cmidrule[0.25pt]{2-15}
 & \multirow{3}{*}{48} & Base & \AppScore{0.109}{0} & \AppScore{0.222}{0} & \AppScore{0.125}{1}\MoraQuoted & \AppScore{0.236}{0}\MoraQuoted & \AppScore{0.119}{0}\MoraQuoted & \AppScore{0.228}{0}\MoraQuoted & \AppScore{0.231}{0} & \AppScore{0.324}{0} & \AppScore{0.303}{0} & \AppScore{0.389}{0} & \AppScore{0.125}{0} & \AppScore{0.239}{0} \\
 &  & \textbf{+MORA} & \AppScore{0.106}{1} & \AppScore{0.217}{1} & \AppScore{0.126}{0} & \AppScore{0.230}{1} & \AppScore{0.110}{1} & \AppScore{0.217}{1} & \AppScore{0.142}{1} & \AppScore{0.248}{1} & \AppScore{0.152}{1} & \AppScore{0.256}{1} & \AppScore{0.115}{1} & \AppScore{0.225}{1} \\
 &  & \cellcolor{MoraTint}Imp.\ (\%) & \cellcolor{MoraTint}\AppImp{3.046} & \cellcolor{MoraTint}\AppImp{1.989} & \cellcolor{MoraTint}\AppImp{-1.121} & \cellcolor{MoraTint}\AppImp{2.505} & \cellcolor{MoraTint}\AppImp{7.322} & \cellcolor{MoraTint}\AppImp{4.706} & \cellcolor{MoraTint}\AppImp{38.688} & \cellcolor{MoraTint}\AppImp{23.650} & \cellcolor{MoraTint}\AppImp{49.885} & \cellcolor{MoraTint}\AppImp{34.103} & \cellcolor{MoraTint}\AppImp{8.367} & \cellcolor{MoraTint}\AppImp{5.605} \\
\cmidrule[0.25pt]{2-15}
 & \multirow{3}{*}{96} & Base & \AppScore{0.154}{0} & \AppScore{0.269}{0} & \AppScore{0.171}{0} & \AppScore{0.281}{0} & \AppScore{0.178}{0} & \AppScore{0.281}{0} & \AppScore{0.379}{0} & \AppScore{0.427}{0} & \AppScore{0.387}{0} & \AppScore{0.468}{0} & \AppScore{0.185}{0} & \AppScore{0.297}{0} \\
 &  & \textbf{+MORA} & \AppScore{0.148}{1} & \AppScore{0.258}{1} & \AppScore{0.171}{0} & \AppScore{0.269}{1} & \AppScore{0.151}{1} & \AppScore{0.255}{1} & \AppScore{0.193}{1} & \AppScore{0.289}{1} & \AppScore{0.204}{1} & \AppScore{0.296}{1} & \AppScore{0.163}{1} & \AppScore{0.269}{1} \\
 &  & \cellcolor{MoraTint}Imp.\ (\%) & \cellcolor{MoraTint}\AppImp{3.519} & \cellcolor{MoraTint}\AppImp{3.970} & \cellcolor{MoraTint}\AppImp{-0.013} & \cellcolor{MoraTint}\AppImp{4.148} & \cellcolor{MoraTint}\AppImp{15.085} & \cellcolor{MoraTint}\AppImp{9.410} & \cellcolor{MoraTint}\AppImp{49.023} & \cellcolor{MoraTint}\AppImp{32.428} & \cellcolor{MoraTint}\AppImp{47.286} & \cellcolor{MoraTint}\AppImp{36.723} & \cellcolor{MoraTint}\AppImp{12.004} & \cellcolor{MoraTint}\AppImp{9.496} \\
\midrule[0.45pt]
\multirow{12}{*}{\rotatebox[origin=c]{90}{\textbf{PEMS04}}} & \multirow{3}{*}{12} & Base & \AppScore{0.068}{0} & \AppScore{0.167}{0} & \AppScore{0.081}{0} & \AppScore{0.189}{0} & \AppScore{0.070}{0} & \AppScore{0.169}{0} & \AppScore{0.098}{0} & \AppScore{0.207}{0} & \AppScore{0.111}{0} & \AppScore{0.226}{0} & \AppScore{0.073}{0} & \AppScore{0.177}{0} \\
 &  & \textbf{+MORA} & \AppScore{0.064}{1} & \AppScore{0.161}{1} & \AppScore{0.072}{1} & \AppScore{0.173}{1} & \AppScore{0.066}{1} & \AppScore{0.163}{1} & \AppScore{0.073}{1} & \AppScore{0.174}{1} & \AppScore{0.076}{1} & \AppScore{0.179}{1} & \AppScore{0.066}{1} & \AppScore{0.165}{1} \\
 &  & \cellcolor{MoraTint}Imp.\ (\%) & \cellcolor{MoraTint}\AppImp{5.406} & \cellcolor{MoraTint}\AppImp{3.684} & \cellcolor{MoraTint}\AppImp{11.134} & \cellcolor{MoraTint}\AppImp{8.489} & \cellcolor{MoraTint}\AppImp{5.853} & \cellcolor{MoraTint}\AppImp{3.903} & \cellcolor{MoraTint}\AppImp{25.557} & \cellcolor{MoraTint}\AppImp{15.758} & \cellcolor{MoraTint}\AppImp{31.532} & \cellcolor{MoraTint}\AppImp{20.921} & \cellcolor{MoraTint}\AppImp{9.127} & \cellcolor{MoraTint}\AppImp{6.719} \\
\cmidrule[0.25pt]{2-15}
 & \multirow{3}{*}{24} & Base & \AppScore{0.080}{0} & \AppScore{0.183}{0} & \AppScore{0.099}{0} & \AppScore{0.210}{0} & \AppScore{0.084}{0} & \AppScore{0.187}{0} & \AppScore{0.156}{0} & \AppScore{0.264}{0} & \AppScore{0.182}{0} & \AppScore{0.298}{0} & \AppScore{0.095}{0} & \AppScore{0.210}{0} \\
 &  & \textbf{+MORA} & \AppScore{0.072}{1} & \AppScore{0.171}{1} & \AppScore{0.082}{1} & \AppScore{0.185}{1} & \AppScore{0.074}{1} & \AppScore{0.174}{1} & \AppScore{0.086}{1} & \AppScore{0.189}{1} & \AppScore{0.093}{1} & \AppScore{0.195}{1} & \AppScore{0.078}{1} & \AppScore{0.181}{1} \\
 &  & \cellcolor{MoraTint}Imp.\ (\%) & \cellcolor{MoraTint}\AppImp{9.494} & \cellcolor{MoraTint}\AppImp{6.310} & \cellcolor{MoraTint}\AppImp{16.886} & \cellcolor{MoraTint}\AppImp{12.036} & \cellcolor{MoraTint}\AppImp{11.082} & \cellcolor{MoraTint}\AppImp{6.860} & \cellcolor{MoraTint}\AppImp{45.180} & \cellcolor{MoraTint}\AppImp{28.572} & \cellcolor{MoraTint}\AppImp{48.635} & \cellcolor{MoraTint}\AppImp{34.410} & \cellcolor{MoraTint}\AppImp{17.956} & \cellcolor{MoraTint}\AppImp{13.426} \\
\cmidrule[0.25pt]{2-15}
 & \multirow{3}{*}{48} & Base & \AppScore{0.100}{0} & \AppScore{0.209}{0} & \AppScore{0.120}{0}\MoraQuoted & \AppScore{0.233}{0}\MoraQuoted & \AppScore{0.113}{0} & \AppScore{0.220}{0} & \AppScore{0.281}{0} & \AppScore{0.364}{0} & \AppScore{0.305}{0} & \AppScore{0.404}{0} & \AppScore{0.123}{0} & \AppScore{0.241}{0} \\
 &  & \textbf{+MORA} & \AppScore{0.084}{1} & \AppScore{0.187}{1} & \AppScore{0.096}{1} & \AppScore{0.200}{1} & \AppScore{0.087}{1} & \AppScore{0.190}{1} & \AppScore{0.105}{1} & \AppScore{0.205}{1} & \AppScore{0.118}{1} & \AppScore{0.210}{1} & \AppScore{0.090}{1} & \AppScore{0.195}{1} \\
 &  & \cellcolor{MoraTint}Imp.\ (\%) & \cellcolor{MoraTint}\AppImp{16.173} & \cellcolor{MoraTint}\AppImp{10.641} & \cellcolor{MoraTint}\AppImp{20.076} & \cellcolor{MoraTint}\AppImp{13.950} & \cellcolor{MoraTint}\AppImp{22.971} & \cellcolor{MoraTint}\AppImp{13.611} & \cellcolor{MoraTint}\AppImp{62.411} & \cellcolor{MoraTint}\AppImp{43.517} & \cellcolor{MoraTint}\AppImp{61.151} & \cellcolor{MoraTint}\AppImp{48.033} & \cellcolor{MoraTint}\AppImp{27.454} & \cellcolor{MoraTint}\AppImp{19.035} \\
\cmidrule[0.25pt]{2-15}
 & \multirow{3}{*}{96} & Base & \AppScore{0.134}{0} & \AppScore{0.247}{0} & \AppScore{0.150}{0}\MoraQuoted & \AppScore{0.262}{0}\MoraQuoted & \AppScore{0.160}{0} & \AppScore{0.266}{0} & \AppScore{0.490}{0} & \AppScore{0.497}{0} & \AppScore{0.387}{0} & \AppScore{0.471}{0} & \AppScore{0.151}{0} & \AppScore{0.268}{0} \\
 &  & \textbf{+MORA} & \AppScore{0.097}{1} & \AppScore{0.202}{1} & \AppScore{0.109}{1} & \AppScore{0.213}{1} & \AppScore{0.101}{1} & \AppScore{0.207}{1} & \AppScore{0.122}{1} & \AppScore{0.217}{1} & \AppScore{0.131}{1} & \AppScore{0.219}{1} & \AppScore{0.101}{1} & \AppScore{0.207}{1} \\
 &  & \cellcolor{MoraTint}Imp.\ (\%) & \cellcolor{MoraTint}\AppImp{27.137} & \cellcolor{MoraTint}\AppImp{17.987} & \cellcolor{MoraTint}\AppImp{27.632} & \cellcolor{MoraTint}\AppImp{18.567} & \cellcolor{MoraTint}\AppImp{36.958} & \cellcolor{MoraTint}\AppImp{22.196} & \cellcolor{MoraTint}\AppImp{75.102} & \cellcolor{MoraTint}\AppImp{56.359} & \cellcolor{MoraTint}\AppImp{66.221} & \cellcolor{MoraTint}\AppImp{53.504} & \cellcolor{MoraTint}\AppImp{32.947} & \cellcolor{MoraTint}\AppImp{22.874} \\
\midrule[0.45pt]
\multirow{12}{*}{\rotatebox[origin=c]{90}{\textbf{PEMS07}}} & \multirow{3}{*}{12} & Base & \AppScore{0.055}{0} & \AppScore{0.150}{0} & \AppScore{0.067}{1} & \AppScore{0.165}{0} & \AppScore{0.056}{0} & \AppScore{0.147}{0} & \AppScore{0.079}{0} & \AppScore{0.184}{0} & \AppScore{0.100}{0} & \AppScore{0.220}{0} & \AppScore{0.064}{0} & \AppScore{0.167}{0} \\
 &  & \textbf{+MORA} & \AppScore{0.053}{1} & \AppScore{0.145}{1} & \AppScore{0.068}{0} & \AppScore{0.156}{1} & \AppScore{0.055}{1} & \AppScore{0.143}{1} & \AppScore{0.063}{1} & \AppScore{0.161}{1} & \AppScore{0.071}{1} & \AppScore{0.174}{1} & \AppScore{0.059}{1} & \AppScore{0.155}{1} \\
 &  & \cellcolor{MoraTint}Imp.\ (\%) & \cellcolor{MoraTint}\AppImp{3.918} & \cellcolor{MoraTint}\AppImp{2.748} & \cellcolor{MoraTint}\AppImp{-2.294} & \cellcolor{MoraTint}\AppImp{5.382} & \cellcolor{MoraTint}\AppImp{1.271} & \cellcolor{MoraTint}\AppImp{2.575} & \cellcolor{MoraTint}\AppImp{19.735} & \cellcolor{MoraTint}\AppImp{12.430} & \cellcolor{MoraTint}\AppImp{29.351} & \cellcolor{MoraTint}\AppImp{21.181} & \cellcolor{MoraTint}\AppImp{8.192} & \cellcolor{MoraTint}\AppImp{7.121} \\
\cmidrule[0.25pt]{2-15}
 & \multirow{3}{*}{24} & Base & \AppScore{0.068}{0} & \AppScore{0.166}{0} & \AppScore{0.087}{0} & \AppScore{0.190}{0} & \AppScore{0.075}{0} & \AppScore{0.167}{0} & \AppScore{0.133}{0} & \AppScore{0.240}{0} & \AppScore{0.186}{0} & \AppScore{0.305}{0} & \AppScore{0.092}{0} & \AppScore{0.204}{0} \\
 &  & \textbf{+MORA} & \AppScore{0.063}{1} & \AppScore{0.159}{1} & \AppScore{0.084}{1} & \AppScore{0.171}{1} & \AppScore{0.067}{1} & \AppScore{0.158}{1} & \AppScore{0.080}{1} & \AppScore{0.181}{1} & \AppScore{0.089}{1} & \AppScore{0.192}{1} & \AppScore{0.071}{1} & \AppScore{0.176}{1} \\
 &  & \cellcolor{MoraTint}Imp.\ (\%) & \cellcolor{MoraTint}\AppImp{7.450} & \cellcolor{MoraTint}\AppImp{4.720} & \cellcolor{MoraTint}\AppImp{3.459} & \cellcolor{MoraTint}\AppImp{9.633} & \cellcolor{MoraTint}\AppImp{10.056} & \cellcolor{MoraTint}\AppImp{5.688} & \cellcolor{MoraTint}\AppImp{40.006} & \cellcolor{MoraTint}\AppImp{24.766} & \cellcolor{MoraTint}\AppImp{52.014} & \cellcolor{MoraTint}\AppImp{36.994} & \cellcolor{MoraTint}\AppImp{22.476} & \cellcolor{MoraTint}\AppImp{14.099} \\
\cmidrule[0.25pt]{2-15}
 & \multirow{3}{*}{48} & Base & \AppScore{0.090}{0} & \AppScore{0.193}{0} & \AppScore{0.110}{0} & \AppScore{0.214}{0} & \AppScore{0.104}{0} & \AppScore{0.198}{0} & \AppScore{0.252}{0} & \AppScore{0.336}{0} & \AppScore{0.350}{0} & \AppScore{0.436}{0} & \AppScore{0.107}{0} & \AppScore{0.220}{0} \\
 &  & \textbf{+MORA} & \AppScore{0.076}{1} & \AppScore{0.175}{1} & \AppScore{0.098}{1} & \AppScore{0.187}{1} & \AppScore{0.085}{1} & \AppScore{0.175}{1} & \AppScore{0.096}{1} & \AppScore{0.195}{1} & \AppScore{0.107}{1} & \AppScore{0.203}{1} & \AppScore{0.080}{1} & \AppScore{0.182}{1} \\
 &  & \cellcolor{MoraTint}Imp.\ (\%) & \cellcolor{MoraTint}\AppImp{15.306} & \cellcolor{MoraTint}\AppImp{9.405} & \cellcolor{MoraTint}\AppImp{10.549} & \cellcolor{MoraTint}\AppImp{13.014} & \cellcolor{MoraTint}\AppImp{18.848} & \cellcolor{MoraTint}\AppImp{11.952} & \cellcolor{MoraTint}\AppImp{62.024} & \cellcolor{MoraTint}\AppImp{41.922} & \cellcolor{MoraTint}\AppImp{69.534} & \cellcolor{MoraTint}\AppImp{53.554} & \cellcolor{MoraTint}\AppImp{24.571} & \cellcolor{MoraTint}\AppImp{16.933} \\
\cmidrule[0.25pt]{2-15}
 & \multirow{3}{*}{96} & Base & \AppScore{0.122}{0} & \AppScore{0.229}{0} & \AppScore{0.141}{0} & \AppScore{0.247}{0} & \AppScore{0.148}{0} & \AppScore{0.243}{0} & \AppScore{0.416}{0} & \AppScore{0.444}{0} & \AppScore{0.517}{0} & \AppScore{0.564}{0} & \AppScore{0.139}{0} & \AppScore{0.252}{0} \\
 &  & \textbf{+MORA} & \AppScore{0.091}{1} & \AppScore{0.188}{1} & \AppScore{0.113}{1} & \AppScore{0.199}{1} & \AppScore{0.106}{1} & \AppScore{0.189}{1} & \AppScore{0.107}{1} & \AppScore{0.202}{1} & \AppScore{0.127}{1} & \AppScore{0.220}{1} & \AppScore{0.093}{1} & \AppScore{0.191}{1} \\
 &  & \cellcolor{MoraTint}Imp.\ (\%) & \cellcolor{MoraTint}\AppImp{25.250} & \cellcolor{MoraTint}\AppImp{17.545} & \cellcolor{MoraTint}\AppImp{19.493} & \cellcolor{MoraTint}\AppImp{19.324} & \cellcolor{MoraTint}\AppImp{28.425} & \cellcolor{MoraTint}\AppImp{22.191} & \cellcolor{MoraTint}\AppImp{74.292} & \cellcolor{MoraTint}\AppImp{54.527} & \cellcolor{MoraTint}\AppImp{75.436} & \cellcolor{MoraTint}\AppImp{60.985} & \cellcolor{MoraTint}\AppImp{33.118} & \cellcolor{MoraTint}\AppImp{24.199} \\
\midrule[0.45pt]
\multirow{12}{*}{\rotatebox[origin=c]{90}{\textbf{PEMS08}}} & \multirow{3}{*}{12} & Base & \AppScore{0.064}{0} & \AppScore{0.162}{0} & \AppScore{0.081}{0} & \AppScore{0.184}{0} & \AppScore{0.071}{0} & \AppScore{0.168}{0} & \AppScore{0.084}{0} & \AppScore{0.191}{0} & \AppScore{0.102}{0} & \AppScore{0.218}{0} & \AppScore{0.067}{0} & \AppScore{0.174}{0} \\
 &  & \textbf{+MORA} & \AppScore{0.060}{1} & \AppScore{0.157}{1} & \AppScore{0.070}{1} & \AppScore{0.168}{1} & \AppScore{0.067}{1} & \AppScore{0.161}{1} & \AppScore{0.069}{1} & \AppScore{0.170}{1} & \AppScore{0.074}{1} & \AppScore{0.181}{1} & \AppScore{0.062}{1} & \AppScore{0.163}{1} \\
 &  & \cellcolor{MoraTint}Imp.\ (\%) & \cellcolor{MoraTint}\AppImp{5.294} & \cellcolor{MoraTint}\AppImp{3.570} & \cellcolor{MoraTint}\AppImp{13.608} & \cellcolor{MoraTint}\AppImp{8.618} & \cellcolor{MoraTint}\AppImp{6.007} & \cellcolor{MoraTint}\AppImp{4.139} & \cellcolor{MoraTint}\AppImp{18.501} & \cellcolor{MoraTint}\AppImp{10.922} & \cellcolor{MoraTint}\AppImp{27.135} & \cellcolor{MoraTint}\AppImp{17.211} & \cellcolor{MoraTint}\AppImp{7.335} & \cellcolor{MoraTint}\AppImp{6.417} \\
\cmidrule[0.25pt]{2-15}
 & \multirow{3}{*}{24} & Base & \AppScore{0.079}{0} & \AppScore{0.182}{0} & \AppScore{0.118}{0} & \AppScore{0.221}{0} & \AppScore{0.096}{0} & \AppScore{0.194}{0} & \AppScore{0.131}{0} & \AppScore{0.239}{0} & \AppScore{0.169}{0} & \AppScore{0.287}{0} & \AppScore{0.086}{0} & \AppScore{0.193}{0} \\
 &  & \textbf{+MORA} & \AppScore{0.073}{1} & \AppScore{0.173}{1} & \AppScore{0.091}{1} & \AppScore{0.191}{1} & \AppScore{0.086}{1} & \AppScore{0.182}{1} & \AppScore{0.088}{1} & \AppScore{0.194}{1} & \AppScore{0.096}{1} & \AppScore{0.207}{1} & \AppScore{0.076}{1} & \AppScore{0.179}{1} \\
 &  & \cellcolor{MoraTint}Imp.\ (\%) & \cellcolor{MoraTint}\AppImp{7.611} & \cellcolor{MoraTint}\AppImp{4.865} & \cellcolor{MoraTint}\AppImp{22.635} & \cellcolor{MoraTint}\AppImp{13.575} & \cellcolor{MoraTint}\AppImp{9.965} & \cellcolor{MoraTint}\AppImp{6.118} & \cellcolor{MoraTint}\AppImp{32.501} & \cellcolor{MoraTint}\AppImp{18.715} & \cellcolor{MoraTint}\AppImp{43.043} & \cellcolor{MoraTint}\AppImp{27.909} & \cellcolor{MoraTint}\AppImp{10.998} & \cellcolor{MoraTint}\AppImp{7.132} \\
\cmidrule[0.25pt]{2-15}
 & \multirow{3}{*}{48} & Base & \AppScore{0.104}{0} & \AppScore{0.212}{0} & \AppScore{0.186}{0}\MoraQuoted & \AppScore{0.235}{0}\MoraQuoted & \AppScore{0.153}{0} & \AppScore{0.244}{0} & \AppScore{0.235}{0} & \AppScore{0.326}{0} & \AppScore{0.295}{0} & \AppScore{0.398}{0} & \AppScore{0.113}{0} & \AppScore{0.224}{0} \\
 &  & \textbf{+MORA} & \AppScore{0.091}{1} & \AppScore{0.196}{1} & \AppScore{0.134}{1} & \AppScore{0.227}{1} & \AppScore{0.121}{1} & \AppScore{0.215}{1} & \AppScore{0.118}{1} & \AppScore{0.226}{1} & \AppScore{0.129}{1} & \AppScore{0.239}{1} & \AppScore{0.096}{1} & \AppScore{0.203}{1} \\
 &  & \cellcolor{MoraTint}Imp.\ (\%) & \cellcolor{MoraTint}\AppImp{12.002} & \cellcolor{MoraTint}\AppImp{7.332} & \cellcolor{MoraTint}\AppImp{28.131} & \cellcolor{MoraTint}\AppImp{3.239} & \cellcolor{MoraTint}\AppImp{20.984} & \cellcolor{MoraTint}\AppImp{11.769} & \cellcolor{MoraTint}\AppImp{49.941} & \cellcolor{MoraTint}\AppImp{30.529} & \cellcolor{MoraTint}\AppImp{56.183} & \cellcolor{MoraTint}\AppImp{40.027} & \cellcolor{MoraTint}\AppImp{15.225} & \cellcolor{MoraTint}\AppImp{9.331} \\
\cmidrule[0.25pt]{2-15}
 & \multirow{3}{*}{96} & Base & \AppScore{0.145}{0} & \AppScore{0.257}{0} & \AppScore{0.221}{0}\MoraQuoted & \AppScore{0.267}{0}\MoraQuoted & \AppScore{0.259}{0} & \AppScore{0.296}{0} & \AppScore{0.392}{0} & \AppScore{0.429}{0} & \AppScore{0.390}{0} & \AppScore{0.479}{0} & \AppScore{0.159}{0} & \AppScore{0.270}{0} \\
 &  & \textbf{+MORA} & \AppScore{0.115}{1} & \AppScore{0.224}{1} & \AppScore{0.221}{0} & \AppScore{0.254}{1} & \AppScore{0.167}{1} & \AppScore{0.244}{1} & \AppScore{0.144}{1} & \AppScore{0.247}{1} & \AppScore{0.154}{1} & \AppScore{0.257}{1} & \AppScore{0.119}{1} & \AppScore{0.228}{1} \\
 &  & \cellcolor{MoraTint}Imp.\ (\%) & \cellcolor{MoraTint}\AppImp{20.679} & \cellcolor{MoraTint}\AppImp{12.789} & \cellcolor{MoraTint}\AppImp{-0.180} & \cellcolor{MoraTint}\AppImp{4.708} & \cellcolor{MoraTint}\AppImp{35.493} & \cellcolor{MoraTint}\AppImp{17.739} & \cellcolor{MoraTint}\AppImp{63.130} & \cellcolor{MoraTint}\AppImp{42.419} & \cellcolor{MoraTint}\AppImp{60.455} & \cellcolor{MoraTint}\AppImp{46.441} & \cellcolor{MoraTint}\AppImp{25.044} & \cellcolor{MoraTint}\AppImp{15.562} \\
\bottomrule[0.8pt]
\end{tabular}
\end{table}

\endgroup
\clearpage

\clearpage
\section{Additional Residual Analysis}
\label{app:residual_analysis}

\morarevised{This appendix analyzes residual energy across nine forecasters
on the eight long-term datasets at $L=H=96$. It also reports
history-based forecast-mean probes for the three forecasters
shown in Figure~\ref{fig:motivation}.}
The analysis retains the original chronological splits and
training-set normalization. DLinear and TimeFilter use seed~2021.
iTransformer uses seed~2021 on ETTh1, ETTm1, and Weather, and seed~2023
on the other five datasets. The linear probes are used only for this
analysis; MORA is fitted as described in Appendix~\ref{app:fitting_protocol}.

\subsection{Frequency Decomposition of Forecast Residuals}
\label{app:residual_spectrum_definition}

For test window $i$ and channel $c$, let
$\mathbf r_{i,c}=\mathbf y_{i,c}-\widehat{\mathbf y}^{(0)}_{i,c}$
be the backbone residual. Let $\mathbf D_H$ denote the orthonormal
DCT-II matrix over the forecast horizon. With $N$ test windows, the
contribution of frequency index $k$ to the overall forecasting MSE is
\begin{equation}
    E_k = \frac{1}{NHC}\sum_{i=1}^{N}\sum_{c=1}^{C}
    \left[(\mathbf D_H\mathbf r_{i,c})_k\right]^2,
    \qquad k=0,\ldots,H-1.
    \label{eq:diagnostic_frequency_energy}
\end{equation}
Orthonormality gives
\begin{equation}
    \mathrm{MSE}=\sum_{k=0}^{H-1}E_k,
    \qquad
    F_8=\frac{\sum_{k=0}^{7}E_k}{\sum_{k=0}^{H-1}E_k}.
    \label{eq:diagnostic_low_frequency_share}
\end{equation}
Thus, $F_8$ measures the share of residual energy in the first eight
modes, including the constant component. Display ranks 1--8 in
Figure~\ref{fig:motivation}(a) correspond to DCT indices 0--7.
The reported fractions use the full residual spectrum and all test
windows and channels; they are ratios of summed energies, not averages
of per-window energy fractions.

\begin{figure}[!htbp]
    \centering
    \includegraphics[width=\linewidth]{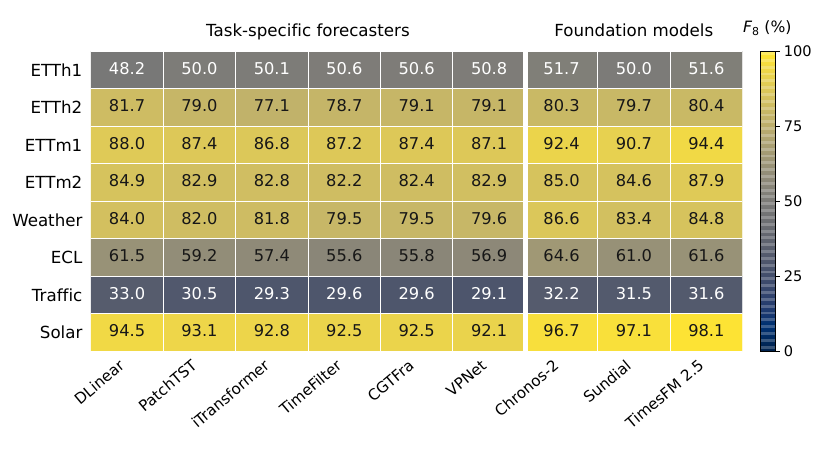}
    \caption{\morarevised{\textbf{Low-frequency residual energy across nine forecasters.}
    Each cell reports $F_8$ (\%) at $L=H=96$, computed over all test
    windows and channels for the corresponding setting.
    The first six columns show task-specific forecasters, and the
    last three show time-series foundation models.}}
    \label{fig:nine_model_residual_energy}
\end{figure}

\morarevised{Figure~\ref{fig:nine_model_residual_energy} shows similar
low-frequency residual-energy shares across the nine forecasters
on each dataset. The differences are more pronounced across
datasets, with the strongest concentration on Solar. Even on
Traffic, which has the lowest share, the first eight of the
96 DCT components account for 29.1--33.0\% of the residual energy.
This is about 3.5--4.0 times the 8.3\% share they would have if
energy were evenly distributed across components.
Table~\ref{tab:residual_diagnostics_across_datasets} reports $F_8$
alongside the history-based probe statistics for the three
forecasters shown in Figure~\ref{fig:motivation}.}
The diagnostic band is fixed and is distinct from MORA's tapered
correction subspace.

\FloatBarrier

\subsection{History-Based Forecast-Mean Probes}
\label{app:forecast_mean_probe}

Following Appendix~\ref{app:mean_probe_connection}, the probe targets
the signed future-mean residual:
\begin{equation}
    \bar r_{i,c}=\frac{1}{H}\mathbf 1_H^{\top}\mathbf r_{i,c},
    \qquad
    \mathbf h_{i,c}=
    \bigl[(\mathbf D_L\mathbf x_{i,c})_0,\ldots,
          (\mathbf D_L\mathbf x_{i,c})_7\bigr]^{\top}.
    \label{eq:diagnostic_history_and_mean_error}
\end{equation}
Here $\mathbf h_{i,c}$ contains eight coefficients of the complete
same-channel historical window, not eight selected time points.
A positive $\bar r_{i,c}$ indicates that the backbone underpredicts the
future mean; a negative value indicates overprediction.

For each backbone and channel, we fit a linear ridge probe with an
intercept on that backbone's validation windows. Let
$\widetilde{\mathbf h}_{i,c}$ denote the features standardized by their
validation means and standard deviations, with a scale floor of $10^{-8}$.
The probe predicts $p_{i,c}=a_c+\mathbf w_c^\top\widetilde{\mathbf h}_{i,c}$
by minimizing
\begin{equation}
    \frac{1}{N_{\mathrm{val}}}
    \sum_{i\in\mathcal I_{\mathrm{val}}}
    (\bar r_{i,c}-p_{i,c})^2+0.01\|\mathbf w_c\|_2^2,
    \label{eq:diagnostic_probe_fit}
\end{equation}
where $\mathcal I_{\mathrm{val}}$ contains the $N_{\mathrm{val}}$
validation windows. Predictions remain in the benchmark's standardized
units. The probe uses neither future observations nor calendar covariates;
its weights, normalization statistics, and group thresholds remain fixed
on test data. Test targets are used only to evaluate the associations.

For Figure~\ref{fig:motivation}(b), each channel's validation probe
scores define five groups through their 20th, 40th, 60th, and 80th
percentiles. We apply these same thresholds to test scores computed
from history. Within each group, we first average the estimated and
observed mean errors over test windows separately for each channel,
then average channels equally. To keep the channel composition the
same across the five groups, this visualization uses channels with
at least one test window in every group. The resulting coverage on
Weather is 19 of 21 channels for DLinear and iTransformer and 18 of
21 for TimeFilter. Coverage depends on the history-based group
assignments, not on the observed errors. The plotted points are group
means, not individual forecasts or a temporal trajectory.

For the cross-dataset summary, we compute the Pearson correlation
between the frozen probe output and $\bar r_{i,c}$ over test windows
within each channel, and report its median over channels with a
defined correlation. This uses all such channels, independently of
the five-group visualization's coverage criterion. Correlations are
defined for 320 of 321 channels on ECL and for all channels in the
other settings reported here.

\begin{table}[!htbp]
\centering
\caption{Residual diagnostics at $L=H=96$. $F_8$ is the DCT-8 share
of squared residual energy; $r_{\mathrm{med}}$ is the median
channel-wise test correlation of the history-based mean-residual probe.}
\label{tab:residual_diagnostics_across_datasets}
\begingroup
\fontsize{9}{10.5}\selectfont
\setlength{\tabcolsep}{4pt}
\renewcommand{\arraystretch}{1.05}
\begin{tabular*}{\linewidth}{@{\extracolsep{\fill}}lrrrrrr@{}}
\toprule
\textbf{Dataset} & \multicolumn{2}{c}{\textbf{DLinear}}
 & \multicolumn{2}{c}{\textbf{iTransformer}}
 & \multicolumn{2}{c}{\textbf{TimeFilter}} \\
\cmidrule(lr){2-3}\cmidrule(lr){4-5}\cmidrule(lr){6-7}
 & $F_8$ (\%) & $r_{\mathrm{med}}$ & $F_8$ (\%) & $r_{\mathrm{med}}$
 & $F_8$ (\%) & $r_{\mathrm{med}}$ \\
\midrule
ETTh1 & 48.195 & 0.115 & 50.060 & 0.264 & 50.558 & 0.219 \\
ETTh2 & 81.734 & 0.161 & 77.132 & 0.068 & 78.749 & 0.277 \\
ETTm1 & 88.041 & 0.008 & 86.825 & 0.128 & 87.196 & 0.122 \\
ETTm2 & 84.858 & 0.206 & 82.751 & 0.318 & 82.226 & 0.210 \\
Weather & 83.984 & 0.471 & 81.847 & 0.302 & 79.537 & 0.206 \\
ECL & 61.541 & 0.176 & 57.438 & 0.228 & 55.567 & 0.225 \\
Traffic & 32.993 & 0.393 & 29.319 & 0.285 & 29.598 & 0.240 \\
Solar & 94.530 & 0.057 & 92.786 & 0.158 & 92.490 & 0.101 \\
\bottomrule
\end{tabular*}
\endgroup
\end{table}

\paragraph{Interpretation.}
Spectral concentration and predictability answer different questions.
DLinear has substantial low-frequency residual energy on both Weather
and ETTm1, yet the corresponding probe correlations are 0.471 and 0.008.
Low-frequency energy alone therefore does not establish that an error
can be inferred from history. The probe provides separate evidence of
history-dependent variation in the future-mean residual, without
implying that every low-frequency component is predictable or that
all backbones share a common training failure. Since test windows
overlap, we report correlations and group means as descriptive
statistics, not independent-sample significance tests.

\FloatBarrier

\begin{figure}[!htbp]
    \centering
    \includegraphics[width=\linewidth]{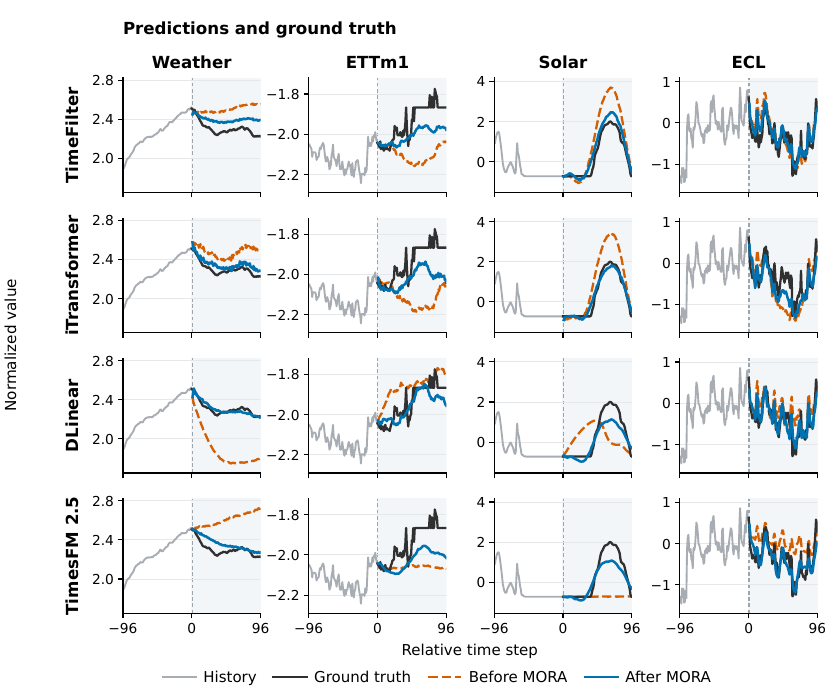}
    \caption{\morarevised{\textbf{Predictions before and after applying MORA across four forecasters.}
    Selected test windows at $L=H=96$. Each column shows the same
    history and ground truth across models. The shaded region marks
    the forecast interval, and relative time zero is the last observation.
    The TimeFilter row matches the upper row of
    Figure~\ref{fig:mora_timefilter_examples}.}}
    \label{fig:mora_all_raw_examples}
\end{figure}

\begin{figure}[!htbp]
    \centering
    \includegraphics[width=\linewidth]{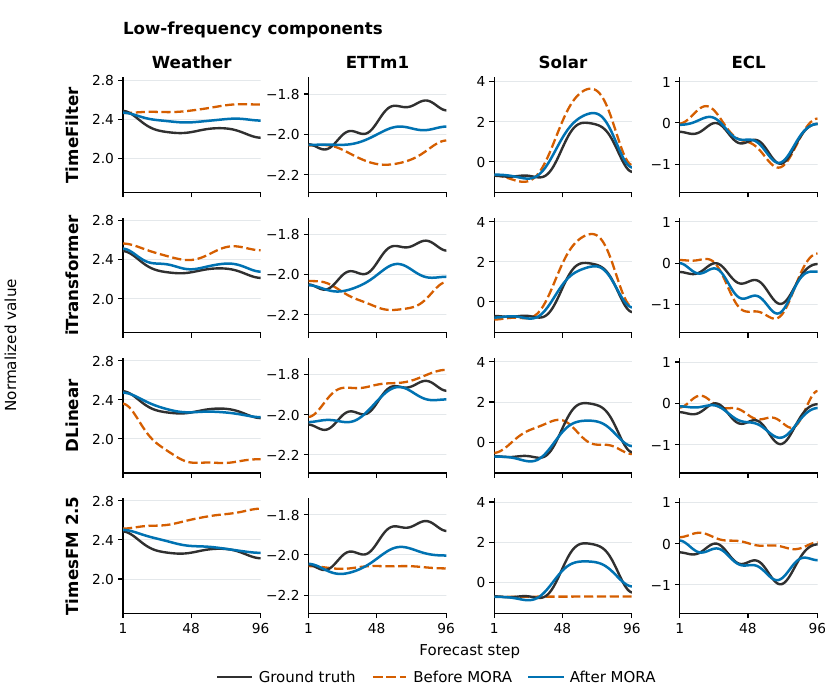}
    \caption{\morarevised{\textbf{Corresponding low-frequency components across four forecasters.}
    The future sequences in Figure~\ref{fig:mora_all_raw_examples}
    are reconstructed from the first eight DCT components, including
    the constant component. The same projection is applied to the
    ground truth and both predictions. Vertical limits match those
    of the original predictions within each dataset. The TimeFilter
    row matches the lower row of
    Figure~\ref{fig:mora_timefilter_examples}.}}
    \label{fig:mora_all_low_frequency_examples}
\end{figure}

\begin{morarevision}
\subsection{Predictions and Low-Frequency Components}
\label{app:prediction_examples}

Figures~\ref{fig:mora_all_raw_examples}
and~\ref{fig:mora_all_low_frequency_examples} extend the comparison
in Figure~\ref{fig:mora_timefilter_examples} to TimeFilter,
iTransformer, DLinear, and TimesFM~2.5 on Weather, ETTm1, Solar,
and ECL. Within each dataset, all four forecasters use the same
selected test window and channel. The two figures show the original
predictions and their low-frequency components, using the DCT-8
diagnostic projection defined in
Appendix~\ref{app:residual_spectrum_definition}.
Vertical limits are shared across models and between the two views
within each dataset. The TimeFilter panels reproduce the examples
in the main text. ECL uses MORA's factorized coarse estimator.
Across these examples, MORA reduces deviations in overall level
and slow variation, bringing the low-frequency components of the
four forecasters closer to those of the ground truth.
\end{morarevision}

\clearpage

\FloatBarrier

\clearpage
\begin{morarevision}
\section{Supplementary Evaluation Results}
\label{app:supplementary_evaluation_results}

\subsection{PEMS Baseline Comparisons}
\label{app:pems_baseline_comparisons}
\end{morarevision}

\begin{table}[!htbp]
\centering
\caption{\morarevised{Author-reported baselines and paired local results
for the six PEMS settings \mbox{marked with $\dagger$}.}}
\label{tab:pems_quoted_paired}
\begingroup
\fontsize{9}{10.5}\selectfont
\setlength{\tabcolsep}{3.5pt}
\renewcommand{\arraystretch}{1.04}
\begin{tabular*}{\linewidth}{@{\extracolsep{\fill}}llc*{6}{c}@{}}
\toprule
\multirow{2}{*}{\textbf{Dataset}} & \multirow{2}{*}{\textbf{Backbone}}
& \multirow{2}{*}{$H$}
& \multicolumn{2}{c}{\textbf{Author-reported}}
& \multicolumn{2}{c}{\textbf{Local baseline}}
& \multicolumn{2}{c}{\textbf{Local \textsc{+MORA}}} \\
\cmidrule(lr){4-5}\cmidrule(lr){6-7}\cmidrule(lr){8-9}
& & & MSE & MAE & MSE & MAE & MSE & MAE \\
\midrule
\multirow{2}{*}{PEMS03} & iTransformer & 48 & 0.125 & 0.236 & 0.135 & 0.247 & 0.126 & 0.230 \\
& CGTFra & 48 & 0.119 & 0.228 & 0.126 & 0.234 & 0.110 & 0.217 \\
\midrule[0.3pt]
\multirow{2}{*}{PEMS04} & iTransformer & 48 & 0.120 & 0.233 & 0.131 & 0.245 & 0.096 & 0.200 \\
& iTransformer & 96 & 0.150 & 0.262 & 0.165 & 0.276 & 0.109 & 0.213 \\
\midrule[0.3pt]
\multirow{2}{*}{PEMS08} & iTransformer & 48 & 0.186 & 0.235 & 0.205 & 0.293 & 0.134 & 0.227 \\
& iTransformer & 96 & 0.221 & 0.267 & 0.292 & 0.324 & 0.221 & 0.254 \\
\bottomrule
\end{tabular*}
\endgroup
\end{table}

\begin{table}[!htbp]
\centering
\caption{\morarevised{PEMS improvements under two baseline comparison
protocols, using identical local \textsc{+MORA} forecasts and the same
aggregation rule.}}
\label{tab:pems_protocol_summary}
\begingroup
\fontsize{9}{10.5}\selectfont
\setlength{\tabcolsep}{3.5pt}
\renewcommand{\arraystretch}{1.04}
\begin{tabular*}{\linewidth}{@{\extracolsep{\fill}}lcccc@{}}
\toprule
\textbf{Baseline protocol}
& \shortstack{\textbf{MSE reduction}\\(\%)}
& \shortstack{\textbf{MAE reduction}\\(\%)}
& \textbf{MSE wins} & \textbf{MAE wins} \\
\midrule
90 local + 6 author-reported & 28.10 & 18.47 & 90/96 & 94/96 \\
96 paired local & \textbf{28.88} & \textbf{19.09} & \textbf{92/96} & 94/96 \\
\bottomrule
\end{tabular*}
\endgroup
\end{table}

\FloatBarrier
\clearpage
\begin{morarevision}
\subsection{Computational Efficiency}
\label{app:computational_efficiency}

We measure inference efficiency on an NVIDIA GeForce RTX 4090 across
the eight long-term datasets at $H\in\{96,720\}$ and $B\in\{1,32\}$.
TimeFilter and CGTFra use $L=96$, while TimesFM~2.5 uses $L=512$.
TimeFilter and CGTFra use GPU-resident inputs and outputs.
TimesFM~2.5 uses its native prediction API with CPU-array inputs and
outputs, including the required transfers. Its $H=96$ predictions
are taken from the prefix of an $H=720$ rollout.

GFLOPs are operator-level estimates, counting one multiply--accumulate
as two FLOPs; latency and peak GPU memory are measured directly.
Peak memory refers to active GPU allocations.
Table~\ref{tab:efficiency_summary} summarizes the average overhead at
$H=720$, and Tables~\ref{tab:efficiency_h96}
and~\ref{tab:efficiency_h720} report the dataset-level results.
\end{morarevision}

\begin{table}[!htbp]
\centering
\caption{\morarevised{Average inference overhead across eight datasets at $H=720$. FLOPs and peak memory are reported at $B=1$; the last column reports additional latency per query at $B=32$. Percentage FLOPs increases are computed from dataset-averaged costs.}}
\label{tab:efficiency_summary}
\begingroup
\fontsize{8.5}{10}\selectfont
\setlength{\tabcolsep}{2.5pt}
\renewcommand{\arraystretch}{1.04}
\begin{tabular*}{\linewidth}{@{\extracolsep{\fill}}lccccc@{}}
\toprule
\textbf{Backbone} & \shortstack{\textbf{Extra}\\\textbf{GFLOPs}} & \shortstack{\textbf{FLOPs}\\\textbf{increase}} & \shortstack{\textbf{Extra peak}\\\textbf{memory (MiB)}} & \shortstack{$B=1$ \textbf{latency (ms)}\\Base $\to$ +MORA} & \shortstack{$B=32$ \textbf{extra}\\\textbf{ms/query}} \\
\midrule
TimeFilter & 0.08745 & 2.512\% & 27.222 & 2.698 $\to$ 3.547 & 0.119 \\
CGTFra & 0.08727 & 2.361\% & 27.289 & 3.595 $\to$ 4.375 & 0.118 \\
TimesFM 2.5 & 0.09778 & 0.00152\% & 45.092 & 639.908 $\to$ 641.442 & 0.464 \\
\bottomrule
\end{tabular*}
\endgroup
\end{table}

\FloatBarrier

\begin{table}[!htbp]
\centering
\caption{\morarevised{Inference costs across eight datasets at $H=96$. Extra GFLOPs and peak memory are reported at $B=1$. Each latency pair shows the baseline and +MORA costs for the entire batch at the indicated batch size.}}
\label{tab:efficiency_h96}
\begingroup
\fontsize{8}{9.4}\selectfont
\setlength{\tabcolsep}{2pt}
\renewcommand{\arraystretch}{1.04}
\begin{tabular*}{\linewidth}{@{\extracolsep{\fill}}llccccc@{}}
\toprule
\textbf{Backbone} & \textbf{Dataset} & \shortstack{\textbf{Extra}\\\textbf{GFLOPs}} & \shortstack{\textbf{FLOPs increase}\\(\%)} & \shortstack{\textbf{Extra memory}\\(MiB)} & \shortstack{$B=1$ (ms)\\Base $\to$ +MORA} & \shortstack{$B=32$ (ms)\\Base $\to$ +MORA} \\
\midrule
\multirow{8}{*}{TimeFilter} & ETTh1 & 0.00065 & 0.238 & 9.047 & 3.339 $\to$ 4.181 & 6.617 $\to$ 7.542 \\
 & ETTh2 & 0.00051 & 0.988 & 8.583 & 1.129 $\to$ 1.706 & 1.157 $\to$ 1.827 \\
 & ETTm1 & 0.00065 & 0.661 & 9.047 & 1.782 $\to$ 2.613 & 1.855 $\to$ 2.884 \\
 & ETTm2 & 0.00065 & 4.917 & 9.046 & 1.783 $\to$ 2.620 & 1.857 $\to$ 2.873 \\
 & Weather & 0.00162 & 8.384 & 10.619 & 3.347 $\to$ 4.086 & 3.437 $\to$ 4.386 \\
 & ECL & 0.01754 & 0.228 & 17.035 & 3.502 $\to$ 3.847 & 42.034 $\to$ 43.038 \\
 & Traffic & 0.04540 & 0.255 & 27.534 & 2.681 $\to$ 3.052 & 49.351 $\to$ 51.325 \\
 & Solar & 0.01200 & 2.033 & 28.831 & 3.372 $\to$ 4.419 & 5.169 $\to$ 6.885 \\
\midrule[0.3pt]
\multirow{8}{*}{CGTFra} & ETTh1 & 0.00040 & 17.023 & 8.540 & 2.426 $\to$ 2.882 & 2.486 $\to$ 3.027 \\
 & ETTh2 & 0.00051 & 10.375 & 8.583 & 2.423 $\to$ 2.977 & 2.468 $\to$ 3.114 \\
 & ETTm1 & 0.00072 & 11.762 & 9.305 & 3.490 $\to$ 4.419 & 3.587 $\to$ 4.761 \\
 & ETTm2 & 0.00040 & 28.065 & 8.540 & 2.401 $\to$ 2.857 & 2.473 $\to$ 3.008 \\
 & Weather & 0.00184 & 40.653 & 11.389 & 2.428 $\to$ 3.262 & 2.444 $\to$ 3.549 \\
 & ECL & 0.01754 & 2.296 & 16.475 & 3.606 $\to$ 3.933 & 4.192 $\to$ 5.197 \\
 & Traffic & 0.04540 & 0.167 & 27.202 & 6.022 $\to$ 6.386 & 53.810 $\to$ 55.941 \\
 & Solar & 0.01199 & 13.418 & 29.041 & 3.531 $\to$ 4.415 & 3.557 $\to$ 5.216 \\
\midrule[0.3pt]
\multirow{8}{*}{TimesFM 2.5} & ETTh1 & 0.00149 & 0.00063 & 10.455 & 400.304 $\to$ 401.657 & 514.840 $\to$ 516.794 \\
 & ETTh2 & 0.00068 & 0.00029 & 8.578 & 398.441 $\to$ 399.172 & 502.904 $\to$ 503.837 \\
 & ETTm1 & 0.00149 & 0.00063 & 10.455 & 393.809 $\to$ 395.190 & 516.495 $\to$ 518.471 \\
 & ETTm2 & 0.00149 & 0.00063 & 10.455 & 394.295 $\to$ 395.744 & 519.280 $\to$ 521.350 \\
 & Weather & 0.00377 & 0.00053 & 14.374 & 397.964 $\to$ 399.372 & 1976.759 $\to$ 1978.833 \\
 & ECL & 0.02655 & 0.00025 & 17.433 & 688.658 $\to$ 689.330 & 21581.336 $\to$ 21589.445 \\
 & Traffic & 0.06960 & 0.00020 & 28.799 & 1984.113 $\to$ 1985.031 & 54124.705 $\to$ 54149.678 \\
 & Solar & 0.02866 & 0.00062 & 53.141 & 445.449 $\to$ 447.040 & 9314.785 $\to$ 9319.383 \\
\bottomrule
\end{tabular*}
\endgroup
\end{table}

\FloatBarrier

\begin{table}[!htbp]
\centering
\caption{\morarevised{Inference costs across eight datasets at $H=720$. Extra GFLOPs and peak memory are reported at $B=1$. Each latency pair shows the baseline and +MORA costs for the entire batch at the indicated batch size.}}
\label{tab:efficiency_h720}
\begingroup
\fontsize{8}{9.4}\selectfont
\setlength{\tabcolsep}{2pt}
\renewcommand{\arraystretch}{1.04}
\begin{tabular*}{\linewidth}{@{\extracolsep{\fill}}llccccc@{}}
\toprule
\textbf{Backbone} & \textbf{Dataset} & \shortstack{\textbf{Extra}\\\textbf{GFLOPs}} & \shortstack{\textbf{FLOPs increase}\\(\%)} & \shortstack{\textbf{Extra memory}\\(MiB)} & \shortstack{$B=1$ (ms)\\Base $\to$ +MORA} & \shortstack{$B=32$ (ms)\\Base $\to$ +MORA} \\
\midrule
\multirow{8}{*}{TimeFilter} & ETTh1 & 0.00396 & 1.405 & 9.753 & 3.336 $\to$ 3.945 & 6.669 $\to$ 7.918 \\
 & ETTh2 & 0.00396 & 3.169 & 9.753 & 1.782 $\to$ 2.391 & 1.856 $\to$ 3.205 \\
 & ETTm1 & 0.00430 & 7.305 & 10.961 & 1.791 $\to$ 2.645 & 1.865 $\to$ 3.581 \\
 & ETTm2 & 0.00430 & 21.606 & 10.961 & 1.814 $\to$ 2.680 & 1.884 $\to$ 3.594 \\
 & Weather & 0.01012 & 38.865 & 15.140 & 3.339 $\to$ 4.104 & 3.409 $\to$ 4.867 \\
 & ECL & 0.16496 & 1.986 & 32.038 & 3.475 $\to$ 4.397 & 42.394 $\to$ 47.275 \\
 & Traffic & 0.43845 & 2.389 & 64.000 & 2.685 $\to$ 3.736 & 50.326 $\to$ 65.265 \\
 & Solar & 0.06955 & 10.255 & 65.169 & 3.366 $\to$ 4.477 & 5.150 $\to$ 8.310 \\
\midrule[0.3pt]
\multirow{8}{*}{CGTFra} & ETTh1 & 0.00304 & 4.626 & 9.642 & 3.533 $\to$ 3.999 & 3.627 $\to$ 4.633 \\
 & ETTh2 & 0.00396 & 23.400 & 9.753 & 2.438 $\to$ 3.007 & 2.487 $\to$ 3.797 \\
 & ETTm1 & 0.00448 & 31.309 & 11.629 & 3.548 $\to$ 4.488 & 3.628 $\to$ 5.469 \\
 & ETTm2 & 0.00304 & 10.174 & 9.642 & 3.568 $\to$ 4.028 & 3.724 $\to$ 4.711 \\
 & Weather & 0.01066 & 52.718 & 17.139 & 2.492 $\to$ 3.332 & 2.490 $\to$ 4.086 \\
 & ECL & 0.16496 & 12.863 & 32.341 & 3.603 $\to$ 4.552 & 4.971 $\to$ 9.926 \\
 & Traffic & 0.43845 & 1.583 & 62.998 & 5.945 $\to$ 7.003 & 54.739 $\to$ 69.985 \\
 & Solar & 0.06956 & 15.454 & 65.169 & 3.634 $\to$ 4.594 & 3.662 $\to$ 6.835 \\
\midrule[0.3pt]
\multirow{8}{*}{TimesFM 2.5} & ETTh1 & 0.00571 & 0.00243 & 22.129 & 390.594 $\to$ 392.066 & 519.381 $\to$ 522.131 \\
 & ETTh2 & 0.00332 & 0.00141 & 17.789 & 392.789 $\to$ 393.556 & 513.885 $\to$ 515.389 \\
 & ETTm1 & 0.00571 & 0.00243 & 22.129 & 398.034 $\to$ 399.438 & 517.112 $\to$ 519.883 \\
 & ETTm2 & 0.00571 & 0.00243 & 22.129 & 397.691 $\to$ 399.213 & 508.988 $\to$ 511.805 \\
 & Weather & 0.01399 & 0.00198 & 33.565 & 399.304 $\to$ 400.831 & 2066.051 $\to$ 2069.163 \\
 & ECL & 0.17401 & 0.00161 & 40.444 & 692.145 $\to$ 693.623 & 21448.834 $\to$ 21472.072 \\
 & Traffic & 0.46275 & 0.00134 & 70.355 & 2010.319 $\to$ 2012.627 & 54783.312 $\to$ 54854.966 \\
 & Solar & 0.11100 & 0.00241 & 132.197 & 438.390 $\to$ 440.180 & 9118.305 $\to$ 9129.273 \\
\bottomrule
\end{tabular*}
\endgroup
\end{table}

\FloatBarrier

\morarevised{At $H=720$, MORA adds an average of 0.848 ms and 0.780 ms
per query to TimeFilter and CGTFra, respectively, across eight datasets.
For TimesFM 2.5, average latency increases from 639.908 ms to 641.442 ms,
corresponding to an overhead of 0.240\%. These costs arise from history
feature extraction, linear residual mappings, and correction
reconstruction, without an additional backbone forecasting call.}

\end{document}